# AI in Oncology: Transforming Cancer Detection through Machine Learning and Deep Learning Applications

Muhammad Aftab[1.3.6], Faisal Mehmood[2], Chengjuan Zhang[5], Alishba Nadeem[1], Zigang Dong[1.3.4,6,7], Yanan Jiang[1.3.4,6,7*], Kangdongs Liu[1.3.4,6,7, *]

[1]Pathophysiology Department, School of Basic Medical Sciences, Zhengzhou University, Zhengzhou, 450001, China.
[2]School of Electrical and Information Engineering, Zhengzhou University, Zhengzhou, Henan, 450001, China.
[3]Tianjian Laboratory of Advanced Biomedical Sciences, Academy of Medical Sciences, Zhengzhou University, Zhengzhou, Henan 450001, China
[4]China-US (Henan) Hormel Cancer Institute, Zhengzhou, Henan 450000, China
[5]Center of Bio-Repository, The Affiliated Cancer Hospital of Zhengzhou University, Henan Cancer Hospital Zhengzhou, Henan, P. R. China.
[6]State Key Laboratory of Esophageal Cancer Prevention and Treatment, Zhengzhou, Henan 450000, China
[7]The Collaborative Innovation Center of Henan Province for Cancer Chemoprevention, Zhengzhou, Henan 450000, China
Contributing Author: maftab@gs.zzu.edu.cn, yananjiang@zzu.edu.cn, faisalmehmood685@uaf.edu.pk, Alishbah44@gmail.com, dongzg@zzu.edu.cn, zcj2016@126.com
*Corresponding author: kdliu@zzu.edu.cn

## Abstract

Artificial intelligence (AI) has potential to revolutionize the field of oncology by enhancing the precision of cancer diagnosis, optimizing treatment strategies, and personalizing therapies for a variety of cancers. This review examines the limitations of conventional diagnostic techniques and explores the transformative role of AI in diagnosing and treating cancers such as lung, breast, colorectal, liver, stomach, esophageal, cervical, thyroid, prostate, and skin cancers. The primary objective of this paper is to highlight the significant advancements that AI algorithms have brought to oncology within the medical industry. By enabling early cancer detection, improving diagnostic accuracy, and facilitating targeted treatment delivery, AI contributes to substantial improvements in patient outcomes. The integration of AI in medical imaging, genomic analysis, and pathology enhances diagnostic precision and introduces a novel, less invasive approach to cancer screening. This not only boosts the effectiveness of medical facilities but also reduces operational costs. The study delves into the application of AI in radiomics for detailed cancer characterization, predictive analytics for identifying associated risks, and the development of algorithm-driven robots for immediate diagnosis. Furthermore, it investigates the impact of AI on addressing healthcare challenges, particularly in underserved and remote regions. The overarching goal of this platform is to support the development of expert recommendations and to provide universal, efficient diagnostic procedures. By reviewing existing research and clinical studies, this paper underscores the pivotal role of AI in improving the overall cancer care system. It emphasizes how AI-enabled systems can enhance clinical decision-making and expand treatment options, thereby underscoring the importance of AI in advancing precision oncology

# 1. Introduction

Recent updates from International Agency for Research on Cancer reveal that cancer is a major global health concern, affecting nations worldwide and posing a huge burden on healthcare systems. Global cancer-related mortality exceeded 10.0 million in 2020, while the number of newly diagnosed cancer cases reached approximately 19.3 million[1]. AI has transformed cancer research and treatment by leveraging its capability to detect complex patterns in medical data and provide accurate quantitative assessments of clinical conditions[2]. The influence of AI is evident in various fields today, such as banking, healthcare, and transportation, and it is anticipated to continue expanding.

AI has been employed in academia to create intelligent tutoring systems, computer programs designed to address the unique needs of individual students. These methods have improved learning outcomes in various subjects, including science and math. However, AI's outstanding analytical abilities have potential to profoundly transform medical research, diagnosis, and treatment. The use of big data in healthcare research offers a unique opportunity to integrate data with complex study findings, requiring significant computational expertise [3, 4].

Healthcare institutions face a variety of complex challenges, particularly those related to biological abnormalities like cancer. AI employs computational mathematical models that mimic human brain functions. Progress in combinatorial chemistry, genomics, and proteomics has made numerous databases of biological and chemical data available. These advancements could greatly deepen our understanding of molecular biology of cancer. A better grasp of cancer biology can substantially impact the evaluation and treatment of cancer patients in clinical settings. However, clinical oncologists face a significant challenge in choosing therapeutically relevant information from extensive raw genomic data now available [5]. Researchers have utilized AI to identify subgroups in various cancers based on gene, mRNA, and miRNAs clusters. The mRNA expression, miRNAs expression, and DNA methylation data were integrated using stacked autoencoders and deep flexible neural forest network models to categories tumors in breast cancer, glioblastoma, and ovarian cancer into subcategories [6].

By utilizing RNA transcriptomics, miRNA transcriptomics, and DNA methylation data, the analysis of hepatocellular carcinoma (HCC) employed both supervised and unsupervised learning techniques. This approach identified consensus driver genes linked to patient survival and delineated two distinct patient subgroups with significantly different survival rates [7]. The integration of proteomics and metabolomics data analysis effectively classified a healthy group into low-risk and high-risk subcategories in breast cancer, as revealed by multi-omics data integration [8]. Two techniques, multiple-kernel frameworks and autoencoders, have been employed to integrate multi-omics datasets for the purpose of breast cancer sub typing [9]. AI plays a crucial role in cancer assessment because it can recognize smaller groups of patients within larger groups based on data related to survival and prognosis. This enables the early detection of cancer and the prediction of its future course. AI was employed to categories patients into subgroups in neuroblastoma by identifying significant characteristics through the analysis of gene expression and copy number alteration data [10].

Biomarkers for relapse in colorectal cancer were identified through a combination of copy number variation, metabolomics, miRNA expression, and gene expression. Similarly, researchers used the MRMR approach to determine significant features for predicting patient survival in ovarian cancer cases [11]. Neural networks trained with DL have also been applied to predict breast cancer survival. The SALMON survival analysis technique identifies relevant feature genes and cytobands by integrating multi-omics data with traditional cancer biomarkers using eigengene matrices of co-expression network modules [12]. Additionally, the predictive value of genomic, epigenomic, and transcriptomic data for various cancers was evaluated using a kernel-based ML method [13]. This approach has shown significant improvements when incorporating clinical parameters, with its efficacy varying by cancer type [14].

Like other cell types, cancer cells respond to environmental signals that influence DNA transcription, cell behavior, and overall function. However, tumor cells are just one component of the cancer development process. [15]. Owing to technological advancements, health researchers and software engineers can collaborate closely to employ multi-factor analysis to improve prediction [16]. According to reports, these analyses are significantly more precise than the actual estimates. Researchers are increasingly prioritizing the development of models that employ AI algorithms for the purpose of cancer detection and prognosis [17]. Currently, these strategies are being utilized to enhance the precision of survival, recurrence, and diverse cancer prognosis [18].

The primary emphasis of research in clinical oncology is to comprehensively comprehend the mechanisms underlying the growth of cancer cells, with the aim of gaining valuable knowledge about the molecular foundation of the disease. Furthermore, its objective was to address the rising global mortality rate attributed to cancer by employing computational biology to handle vast amounts of data from millions of pertinent cases [19]. Moreover, the integration of AI into clinical decision-making is expected to enhance the use of high-resolution imaging and Next Generation Sequencing (NGS) for early detection and prediction of diseases. By leveraging these advanced technologies, AI can significantly improve the accuracy and timeliness of disease diagnosis and prognosis

AI can be utilized to generate novel biomarkers for the purpose of diagnosing cancer[20]. The objective of AI functionality is to construct a system that is adequately trained to accurately determine if a patient will require immunotherapy. AI has  capability to identify patients who require further testing, such as Whole Genome Spectroscopy, and determine the immunotherapy medicine that will have the most influence on a patient's recovery [21]. The increasing utilization of AI in the medical domain, supported by verified real-world case studies, aims to effectively overcome the limitations and constraints in accurately identifying different types of cancer. Included is a thorough explanation of the framework required for AI to function as intended.

The main objective of the present study was to examine the potential application of AI in enhancing cancer prognoses, with a specific focus on top 10 most prevalent malignancies globally. A range of imaging techniques is concisely examined, alongside a diverse array of DL algorithms and models they generate. Researchers are currently examining the utilization of AI in the field of digital pathology and diagnostics to gain a better understanding of how AI could potentially influence cancer treatment in future.

## 1.1 Role of AI in Cancer Diagnosis by using medical imaging

Imaging tests such as MRIs, CT scans, and PET scans formed the basis of the previous cancer diagnosis procedure [22]. These traditional techniques may result in invasive or uncomfortable diagnostic procedures, and various healthcare professionals may have differing interpretations of the imaging data [23]. AI systems, especially those utilizing DL techniques, demonstrate exceptional proficiency in interpreting medical images and identifying abnormalities that are frequently overlooked by human observers.

By leveraging extensive cancer databases, it is possible to identify individuals at high risk of developing cancer by analyzing their health characteristics, thereby enabling early screenings. AI has potential to revolutionize cancer detection. For instance, Google's DeepMind project, in collaboration with the UK's National Health Service, developed an AI system capable of outperforming human doctors in accurately detecting over 50 eye diseases using 3D scans [24]. Furthermore, PathAI incorporates AI into the field of pathology in order to assist in the detection of various diseases, such as cancer. It ensures comprehensive detection of all possible elements, aiding pathologists in quickly recognizing patterns [25, 26].

AI is also revolutionizing the field of medicine by offering exceptional precision the detection of many forms of cancers [27]. AI employs sophisticated algorithms and ML techniques to effectively and swiftly identify malignancies, a skill that was previously unattainable.

AI algorithms provide a chance to enhance the precision of medical imaging, hence decreasing the amount of time physicians need to spend on patient care. A recent study published in Academic Radiology reveals that radiologists must assess an average of one image every three to four seconds in order to manage their daily workload. The integration of AI capabilities in image processing and radiography has the potential to enhance productivity in both domains. Mammography is the predominant technique used to identify breast cancer in its early stages, with more than 200 million women undergoing this procedure annually on a global scale.

The scientists proved their statement by examining 500 randomly selected cases and comparing them to previous assessments made by six professional radiologists, concluding that the AI system outperformed the radiologists' conclusions [28]. Overall, the US experienced a 5.7% reduction in false-positive results, while UK as a whole observed a 1.2% decline. Additionally, there was a 9.4% decrease in false negatives in US and a 2.7% decrease in UK. The findings of this first study open the path for larger clinical studies with the potential to improve the accuracy and efficacy of cancer diagnosis utilizing digital technologies [29].

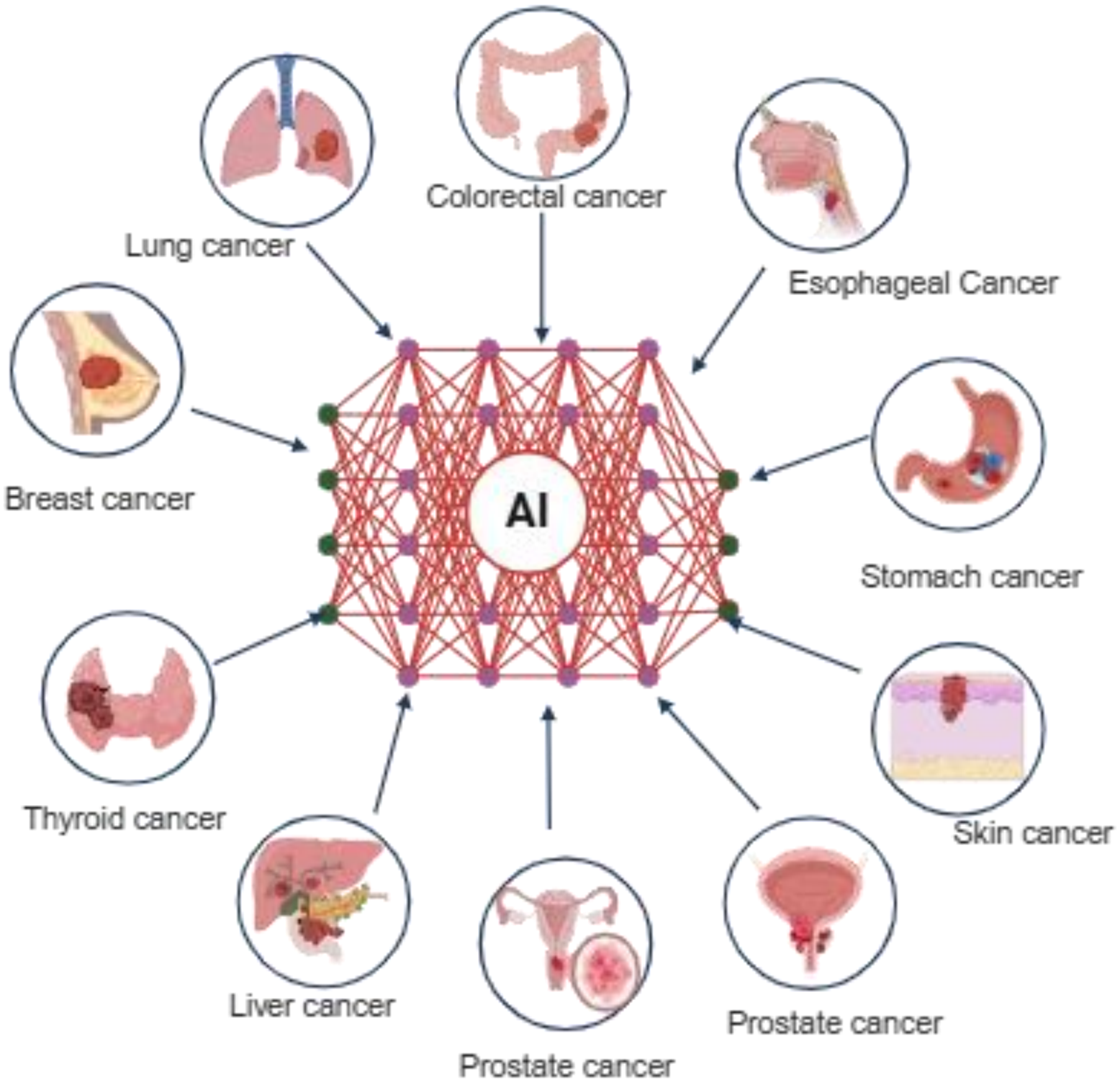

**Figure 1:** This image highlights AI's transformative role in detecting and diagnosing ten types of cancer, enabling precise and rapid analysis across various cancers.

An examination was conducted on MammoScreen system to demonstrate the benefits of integrating novel AI technology with radiologists. This approach facilitates the detection of suspicious areas in 2D mammograms indicative of potential breast cancer presence and enables cancer risk assessment. Various parameters were evaluated by the researchers, encompassing reading time, sensitivity, specificity, and the area under the ROC curve (AUC). Results indicated that individuals not utilizing AI achieved an average AUC of 0.769 (95% CI, 0.724-0.814). Conversely, when AI was employed, the AUC recorded was 0.7494 (95% CI: 0.754-0.840) [30]. The enhancement of screening consistency was evident among eight radiologists, leading to a notable 25% decrease in false-positive results. In March 2022, MammoScreen received marketing approval from the US Food and Drug Administration (FDA). A noteworthy instance of this progress is exemplified by a computer vision DL algorithm, which autonomously acquires the capability to detect anomalies in lung tissue through the utilization of a dataset comprising 1,000 AI-generated CT scans. Remarkable strides have been achieved in the identification of diverse cancer types, encompassing thyroid, skin, esophageal, lung, and breast tumors.

Nonetheless, certain challenges persist within our current methodologies. While fine needle aspiration is commonly utilized in thyroid cancer diagnosis, it encounters obstacles in accurately distinguishing between benign and malignant nodules. The integration of AI into the analysis process holds promise for enhancing the categorization of medical imaging data, particularly ultrasound images.[31]. The system determined that AI had a 30% higher accuracy rate in detecting lung cancer compared to human detection [32]. These findings demonstrate that the integration of AI technology with human radiologists' assessments can be utilized to detect cancer screening tests, hence improving their precision.

People are still more capable of reasoning than AI, even if AI systems may be able to detect subtle changes in tissues that are hidden from the human eye. The ultimate goal is to determine the best strategy for combining the two to transform the radiology industry [33].

## 1.2 Clinical Applications of AI and Machine Learning

Clinical applications of cancer employ a diverse array of tactics and processes to effectively diagnose, treat, and manage the disease. These applications utilize advanced diagnostic technologies such as molecular profiling and genetic testing to detect specific cancer subtypes and develop personalized treatment recommendations.

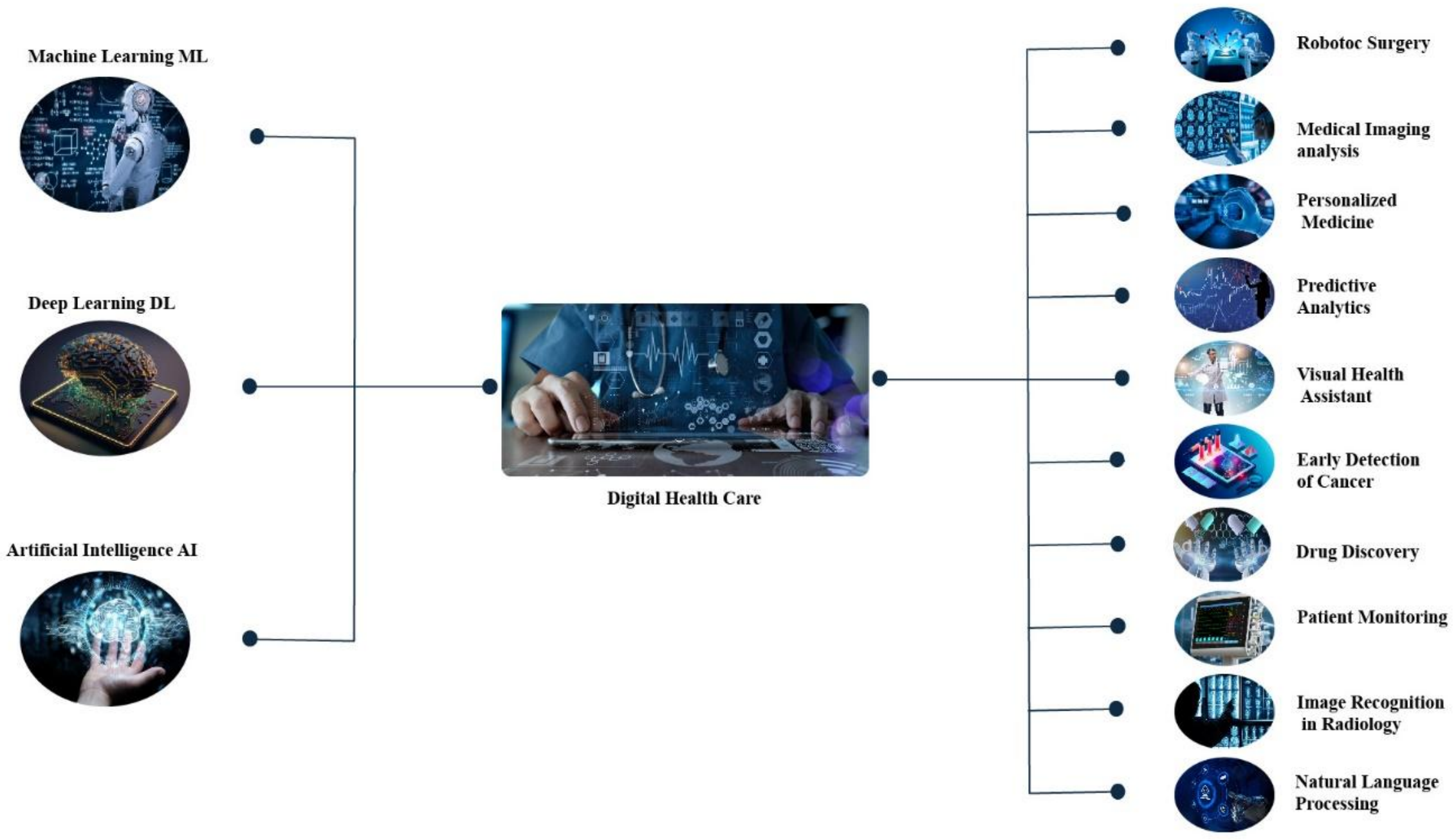

**Figure 2:** Applications of AI, ML, and DL in oncology and digital health care can predict the best line of treatment and cure health issues.

The application of AI, ML and DL are shown in Figure 2. In addition, new therapeutic options such as targeted therapies and immunotherapies have arisen, which precisely aim at cancer cells or enhance the body's immune system to combat the disease. Clinical studies are crucial for assessing innovative cancer therapies and drugs as they have the potential to improve outcomes and increase survival rates. Alongside treating physical symptoms, palliative care and supporting services assist cancer patients live better lives by taking care of their psychological and emotional needs. Overall, new clinical applications in oncology are constantly being developed, which facilitates more effective cancer management and enhances patient care.

### 1.2.1 Clinical Advantage of AI and Precision Oncology

Prior studies have unequivocally demonstrated the substantial benefits of AI in healthcare, with advancements occurring at an exceptionally rapid pace. For AI systems to be successfully integrated into clinical practice, they must perform at least as well as, if not better than, human intervention. In clinical settings, next-generation sequencing (NGS) offers numerous benefits for identifying prognostic or predictive biomarkers [34]. Over the past decade, NGS has significantly advanced in terms of speed, accuracy, affordability, and efficiency. Additionally, NGS has incorporated both short-read and long-read platforms. Short reads can be employed in precision medicine to detect polymorphisms that provide clinical advantages and for screening large populations, while long reads facilitate full-length isoform sequencing. The utilization of sophisticated algorithms to analyze complex datasets will unveil new opportunities for targeted therapies in precision oncology [35-37]. AI-driven integration of genomics, transcriptomics, and proteomics data, as shown in Figure 3, distinguishes cancer subtypes, predicts disease progression, and identifies therapeutic targets, enhancing personalized cancer diagnosis and treatment

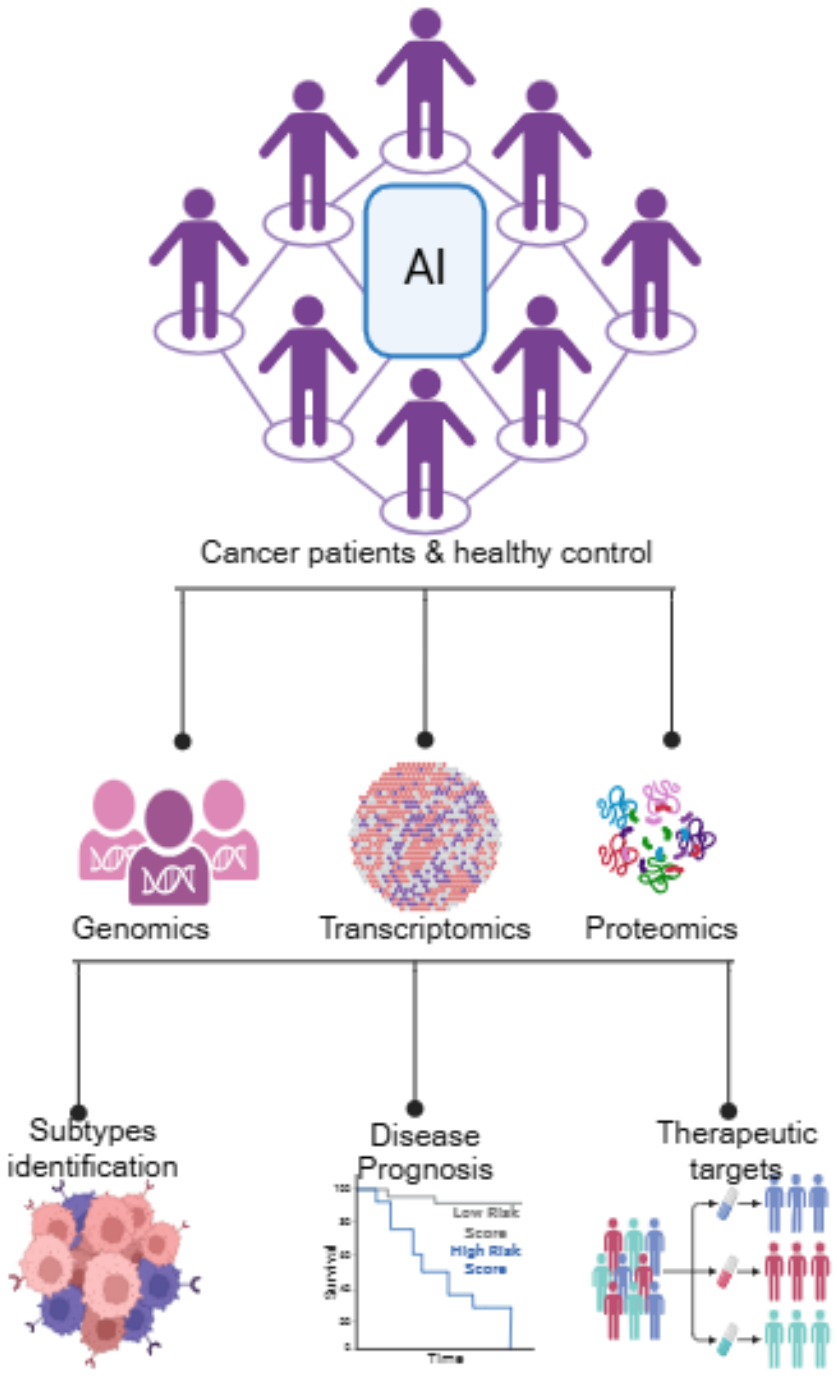

**Figure 3:** The image shows AI-based machine for analyzing multi-omics cancer patients' and healthy donors' data to improve cancer diagnosis and treatment. By integrating genomics, transcriptomics, and proteomics data, the AI algorithm achieves several key outcomes: meeting new challenges related to identification of cancers subtypes, predicting diseases progression and survival probabilities and discovering new drug targets. This multifaceted approach integrates these advanced mathematical methods to improve the comprehension of the cancer exhaustion as well as to create the individually-targeted therapies.

Automation of the healthcare system is particularly vital in regions with little resources. An important issue in developing nations is the need for more adequately trained healthcare personnel and specialists. This problem can be addressed by introducing AI systems that can identify diseases at a faster rate. AI provides an extra advantage by reducing the costs associated with managing health records and eliminating monotonous administrative tasks.
AI systems enhance the organization and examination of patient health information through automated processes, assisting physicians in making clinical decisions and reducing their documentation workload. Furthermore, patient health records can be used to precisely forecast future disease risk [38].

### 1.2.2 From Lab to Clinic: AI's Rapid Advancements in Cancer Care

For almost 20 years, AI has been used in cancer research. The area of cancer research has shown encouraging advancements, and noteworthy accomplishments at the level of expertise [39]. AI is also being used by a growing number of businesses and industry research groups to identify, diagnose, and treat cancer. IBM is leading the effort to incorporate AI into medical facilities. IBM introduced Watson, an AI system, in 2014 to provide cutting-edge medical support in the area of cancer research [40]. Corresponding to this, a group of researchers and computer scientists at Microsoft are working to use ML and NLP to create a programming framework that will allow them to manipulate biological processes in order to treat cancer. However, there are concerns over the regulation of AI tools. In 2022, the FDA authorized more than sixty medical devices or algorithms powered by AI, as determined by Jaber's research [41].

AI is positioned to transform the medical industry by facilitating suggestions and recommendations for treatment options, improving the accuracy and speed of diagnosis, and advancing prognostic outcomes [42, 43].

### 1.2.3 Consensus regarding the AI procedure in clinical cancer research

AI is presently employed exclusively in an adjunctive capacity within medicine. However, there remains uncertainty regarding the potential for technology to eventually supplant human expertise in diagnostic imaging, treatment decision-making, and other areas. Currently, AI encounters numerous challenges in cancer research, particularly in the development and evaluation of solutions. To verify the capabilities of AI effectively, meticulously designed testing protocols are paramount. Consequently, an ideal framework is essential to facilitate the advancement of AI methodologies. Prior to the implementation of an AI solution, it is crucial to define and characterize the specific issues to be addressed and to evaluate whether addressing these issues is both feasible and beneficial [44]. Multiple clinical data sets reveal an inherent disparity in the allocation of samples among different classes. When dealing with distinguishing features like asymmetry, classification algorithms which are typically built to handle balanced classes may encounter difficulties [45]. Three separate categories training, validation, and testing sets are used to group the available data. This category facilitates an impartial assessment of the AI model. The typical ranges are 60–70%, 15–20%, and 15–20%. AI employs three separate datasets: training, validation, and testing. The former is used for model training, while the later is used for feature selection and parameter modification.

The diagnostic workflow of AI is shown in Figure 4. Some characteristics may have a detrimental impact on the classifier's performance as a result of noise or irrelevance, and not all of them are valuable for generating AI models. Feature selection can be used to highlight the most crucial features for discovering suitable bioactivity [46].

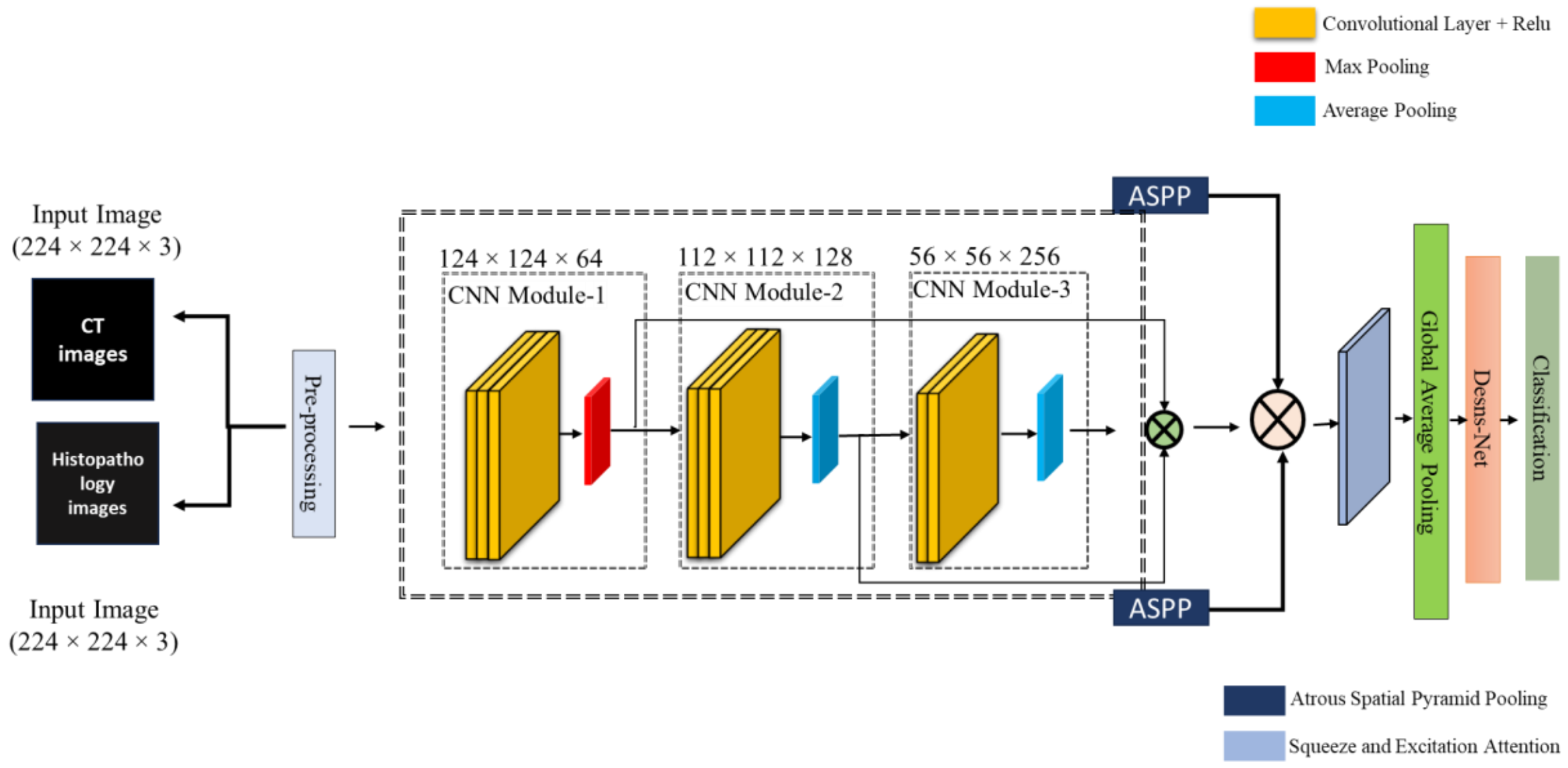

**Figure 4:** illustrates an advanced CNN system integrating CT and histology images, employing three CNN modules and ASPP for multi-scale contextual feature extraction. After global average pooling, the unified vector is fed into a pre-trained Dense Net and classification block to detect cancer, exemplifying how state-of-the-art ML techniques enhance diagnostic accuracy

AI, especially DL, has made significant strides in cancer clinical research thanks to the increasing amount of contemporary biological data. AI-based technologies are increasingly being employed to improve accuracy and efficiency in a variety of cancer clinical research applications. These include the detection of cancer by imaging, the analysis of genetic data, the extraction of important information from medical records, the development of new drugs, and the utilization of biological literature. Furthermore, we will analyze the limitations of traditional methods and the importance of AI in the context of the ten most prevalent types of cancer worldwide.

# 2. Role of AI in various cancers

## 2.1 Lung Cancer Detection and AI

Lung cancer is primary cause of cancer-related fatalities on a global scale and is among the most commonly diagnosed malignancies. About 2.20 million new patients receive a lung cancer diagnosis each year [47]. Additionally, 75% of people with a diagnosis die within five years [48]. The challenge in treating cancer has arisen due to intricate nature of cancer cells and the significant intra-tumor heterogeneity (ITH), which results in the development of medication resistance [49].

The progressive development of technologies in cancer research in recent decades has enabled the establishment of multiple clinical and medical imaging methods [50], and genomic databases through various large-scale collaborative cancer projects in recent decades[51]. These databases assist researchers in analyzing extensive patterns of lung cancer, encompassing its diagnosis, therapy, responses, and clinical outcomes [52]. Recent studies in -omics analysis, encompassing genomics, transcriptomics, proteomics, and metabolomics, have significantly enhanced our research methodologies and capabilities [53]. Currently, there is a prevailing tendency in cancer research to integrate vast quantities of data from several categories [54, 55].

Although dimension reduction techniques such as tensor and matrix factorizations can be helpful, the use of various and high-dimensional data types for clinical applications requires significant time and expertise [56]. Analyzing the datasets linked to cancer, which are growing exponentially, presents a major challenge for researchers [57]. Therefore, it is now more important than ever to utilize ML models that independently identify the inherent characteristics of different forms of data to assist doctors in their decision-making process [58].

This article presents a comprehensive examination of primary ML methods that have been employed to integrate intricate biomedical data, such as sequencing or imaging data, pertaining to various facets of lung cancer. Additionally, it explores significant obstacles and prospects for the future utilization of ML in the realm of clinical research and the treatment of lung cancer. The purpose of this review is to improve understanding of the capabilities and potential applications of ML in this specific field.

## 2.2 Application of Machine learning (ML) for lung cancer early detection and supplementary diagnosis

Applying ML techniques facilitates early diagnosis and detection by analyzing medical imaging datasets. Timely identification is crucial in reducing lung cancer mortality rates [59]. The primary method for monitoring individuals at high risk of developing lung cancer is low-dose computed tomography (CT) for chest screening [60]. The main objective of developing computer-aided diagnosis (CAD) systems is to enhance efficiency of medical diagnostics. These systems assist clinicians in analyzing medical imaging data, providing a valuable additional perspective [61]. Traditional feature-based CAD tasks consist of three phases: nodular segmentation, feature extraction and selection, and clinical judgment inference (classification). Various techniques are available for predicting the probability of cancer using ML classifiers, including logistic regression (LR) [62] or linear discriminant analysis (LDA) [63], which consider the observed textural properties of specific nodules in CT images along with the patient's clinical data.

CT images commonly provide factors such as nodule size, type, location, count, border, and emphysema information. Clinical variables include the patient's age, gender, time of specimen collection, family history of lung cancer, and smoking history, among others. However, these traits are typically based on subjective assessments and lack well-defined, comprehensive, and quantifiable descriptions of the look of the malignant nodule.

The implementation of DL based models within CAD systems has been the focus of increasing research efforts due to advancements in DL techniques, particularly CNNs. The goal is to improve system precision, reduce false positives, and decrease detection time for lung cancers [64]. The application of ML techniques for predicting lung cancer prognosis is demonstrated in Tables 4 and 5. Similar to feature-based CAD systems, these models typically follow three stages: identifying and separating nodules, extracting their features, and making conclusions based on clinical judgment [65].

Unlike traditional feature-based systems, DL-based CAD systems can accurately represent the three-dimensional structure of a nodule and independently identify and extract inherent characteristics of potentially abnormal nodules [66, 67]. For instance Tyagi et al. [68] developed a model by extracting three 3D-view feature vectors (axial, coronal, and sagittal) of the nodule from CT scans. These feature vectors were derived from OverFeat [69].

Recently, a comprehensive global evaluation of nodules to define their properties using CT images has been made possible by incorporation of CNN models. Shariaty et al. [70] developed a complementary CNN model in which shape descriptions (the "shape" feature) of the segmented nodule were obtained using a spherical harmonic model [71] for nodule segmentation and the texture and intensity features (the "appearance" feature) of the nodule were extracted using a DCNN based model [72]. Based on the addition of "shape" and "appearance" attributes, the following classification was made. Venkadesh et al. [73] employed an ensemble model comprised of two distinct models 2D-ResNet50-based [74] and 3D-Inception-V1 [75] to extract two different aspects of a lung nodule. Subsequently, these two characteristics were combined and employed as input characteristics for the classification procedure.

Applying the ensemble CNN model to raw CT scans offers several benefits, including the ability to distinguish between cancerous and non-cancerous nodules. By leveraging features from state-of-the-art CNN models, common ML techniques such as support vector machines (SVM), neural networks (NNs), and LR can be applied to clinical decision inference. Notably, some studies also utilized CNN models for conclusive clinical assessments. Mikhael et al. [76] presented a thorough approach for systematically simulating the classification of lung cancer risk and localization tasks using only the input CT data.

Their approach to forecasting the likelihood of malignancy relied on a fusion of three CNN models: a Mask-RCNN [77] model for segmenting lung tissue, a modified U-Net [78] model for detecting cancer regions of interest (ROI), and a full-volume model derived from 3Dinflated Inception-V1 [79] for estimating the likelihood of cancer. Beyond their use in CT images, CNN-based models are widely used in histology imaging to assist in the diagnosis of lung cancer. Histological imaging offers a greater depth of cellular-level biological data on malignancy compared to CT imaging.

Abdul-Jabbar et al. [80] achieved this by segmenting individual cells from hematoxylin and eosin (H&E)-stained and immunohistochemistry (IHC) images using a Micro-Net [81] model for identifying tissue boundaries and a SC-CNN [82] model for the same purpose. Although ML algorithms are widely used in CAD, the challenge lies in the fact that only a limited number of images are actually used.

### 2.3 AI in Lung Cancer detection and subtype Differentiation

It is challenging to implement periodic medical imaging scans, which are advised for high-risk groups, due to a high probability of false detection [83, 84]. That is why it is essential to develop more effective methods for early detection of lung tumours. The numerous applications of AI and ML are illustrated in Figure 5. Advancements in sequencing technologies have enabled the creation of an abundance of instruments for the early detection of lung cancer. [85]. However, it is crucial to precisely classify different subtypes of lung cancer in order to guide the selection of the most efficient treatment strategies. Except for targeted therapy, the management of the two most common subtypes of lung cancer, LUAD (45%) and LUSC (25%), is often done in a similar manner [86]. Nevertheless, studies have shown that LUAD and LUSC have distinct biological characteristics, indicating that these two forms of cancer should be classified and treated as separate entities [87]. Early detection and subtype identification using a computer approach are both included in the classification job.

To ascertain the relative abundances of each cell type in the images, the segmented cells were then employed in cell type classification. This model allows for the precise differentiation of immune evasion mechanisms and differential evolution between lung LUAD and LUSC with exceptional precision.

A subsequent analysis[88] categorized tissue as LUAD, LUSC, or normal based on H&E-stained Histopathological whole-slide pictures using the Inception-V3 network [89]. An important feature of this research is that the model shows the ability to predict the occurrence of somatic mutations in different driver genes of lung cancer, including STK11, EGFR, FAT1, SETBP1, KRAS, and TP53, in a specific tissue. Several studies utilized transfer learning to improve the effectiveness and resilience of their novel model training procedures [90], considering the significant resources and complex nature of the datasets.

Previous research in ML has shown that massive pan-cancer sequencing datasets can be effectively used to classify different types of cancer and detect them at an early stage [91, 92]. These results may provide evidence that supports the validity of the lung cancer diagnosis. Cancer cells are characterized by a multitude of genetic changes, which can be used as distinctive markers to document the specific mutational patterns associated with different forms of cancer [93, 94]. Current research has mostly concentrated on improving the accuracy of their ML models by extracting superior genetic signatures as input features.

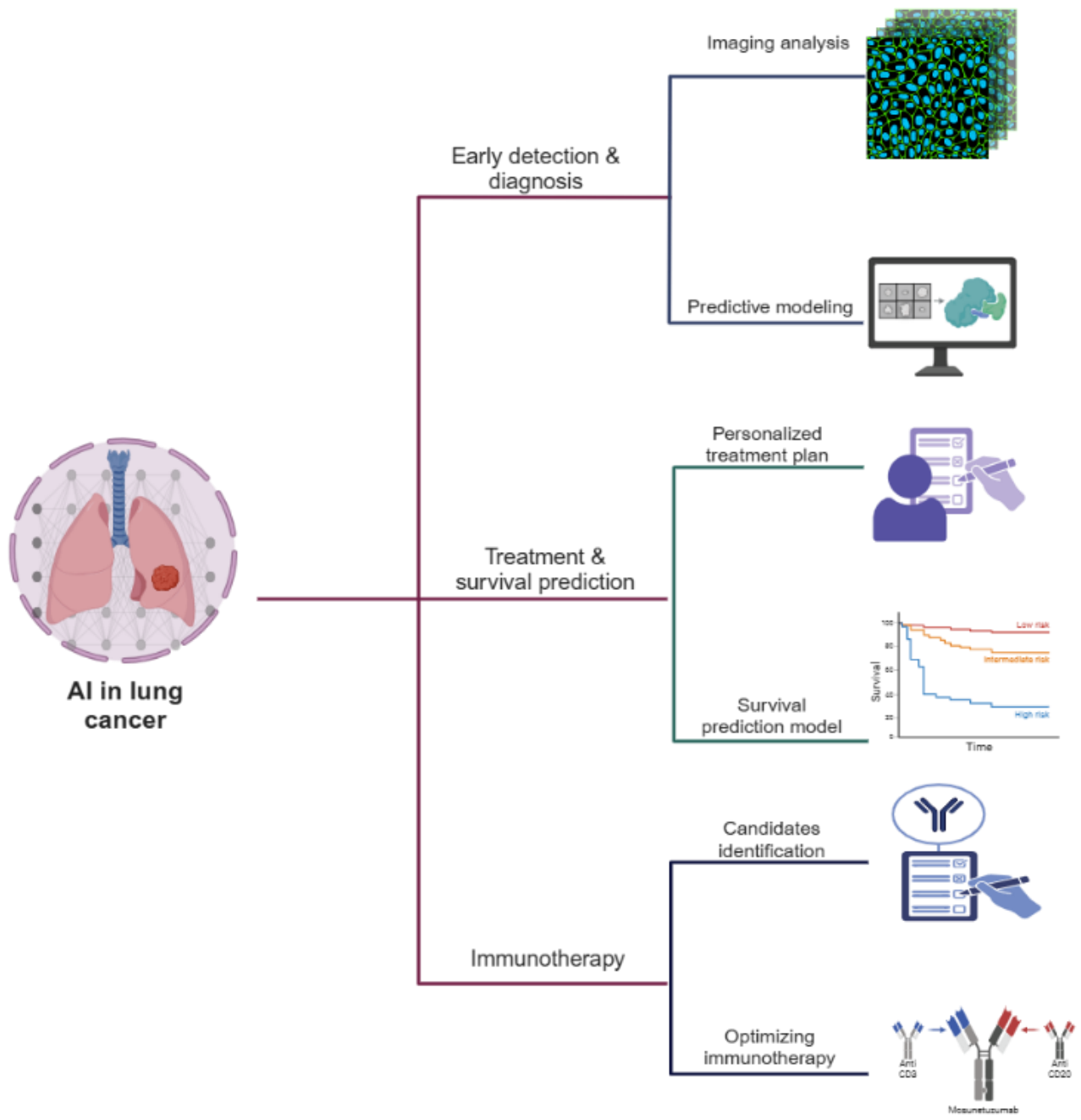

**Figure 5:** AI significantly enhances lung cancer management by improving early detection, diagnosis, and personalized treatment plans. It aids in survival prediction and optimizes immunotherapy by analyzing patient-specific data and immune responses.

The graphic demonstrates a full AI method for tailored lung cancer care, encompassing early identification, prediction of treatment effectiveness, and adjustment of immunotherapy protocols. Table 1 and Table 2 provided a concise summary of how Al integration contributes to detection, utilizing Image and sequencing data, respectively.

**Table 1:** ML for imaging-based early diagnosis and examination.

| Input | Sample Type/Images | Performance | Advantages | References |
|---|---|---|---|---|
| Clinical risk factors + nodule | CT images/ 2961 | AUC (0.907–0.960) | High AUC in small nodules (< 10 mm) | [95] |
| Clinical factors + nodule | CT images/300 | AUC (0.706–0.932) | Comparable performance to human observers | [62] |
| Mass spectra from ROIs of MALDI image | MALDI/326 | Accuracy (0.991) | High accuracy on FFPE biopsies | [63] |
| CT imaging patches + radiologists' binary nodule segmentations | CT images/1018 | Accuracy (0.793–0.824) | Higher predictive accuracy by integrating shape and appearance features | [96] |
| 3D CT volume feature | CT images/1018 | Accuracy (0.9126) | Higher accuracy than benchmarked models | [97] |
| 3D CT volume feature | CT images/6960 | DSC (0.91); classification accuracy (0.97) | Integration of clustering and scarification algorithms for accurate extraction | [98] |
| 3D CT volume feature, nodule position coordinate, and maximum diameter | CT images/1729 | AUC (0.868) | First study attempting to classify benign or malignant nodules | [99] |
| 3D CT volume feature and nodule coordinates | CT images/16429 | AUC (0.86–0.96) | Higher AUC than benchmarked models | [100] |
| Patient's current and prior 3D CT volume features | CT images/14851 | AUC (0.944) | Higher AUC than radiologists when prior CT images are not available | [101] |
| Image features of H&E-stained tumor section histological slides | Histological images/100 | Accuracy (0.913) | Annotates cell types at single-cell level using histological images | [80] |
| Transformed 512x512-pixel tiles from nonoverlapping 'patches' whole-slide images | Histological images/1634 | AUC (0.733–0.856) | Predicts somatic mutations in specific genes | [88] |
| Initial + synthetic CT images | CT images/22489 | Accuracy (0.9986) | Uses GAN to generate synthetic lung cancer images to reduce over fitting | [102] |
| Initial + synthetic Histopathological images | Histopathological images/15000 | Accuracy (0.9984); F1-score (99.84%) | Uses GAN to generate synthetic lung cancer images and a regularization-enhanced | [103] |

**Table 2:** Machine learning applied for the analysis of sequencing data to enable early detection and classification.

| Model Type | Sample type/Image | Performance | Feature Selection | Advantages | Reference |
|---|---|---|---|---|---|
| LR with LASSO penalty | cfDNA fragments/799 | AUC (0.98) | cfDNA features + clinical + CT imaging | Combines cfDNA with other markers for lung cancer detection | [104] |
| 5-NN, 3-NN, NB, LR, DT | cfDNA/160 | AUC (0.69-0.98) | SNV + CNV features | Early lung cancer detection using cfDNA | [105] |
| LR | ctDNA/296 | AUC (0.816) | 9 DNA methylation markers | Early lung cancer detection using DNA methylation | [106] |
| RF, SVM, LR with ridge/elastic net/LASSO | cfDNA/843 | mAUC (0.896-0.936) | Copy number profiles | Biomarker framework for lung cancer detection using cfDNA copy number profiling | [107] |
| Diet Networks with EIS | Somatic mutations/954 | Accuracy (0.8) | SNVs, insertions, deletions across 1796 genes | EIS stabilizes Diet Network training | [108] |
| LR | RNA-seq of BECs/299 | AUC (0.81) | Lung cancer-associated & clinical covariate RNA markers | Maintains sensitivity for small/peripheral lesions | [109] |
| KNN, NB, SVM, C4.5 DT | RNA-seq/529 | AUC (0.91) | Hold-out RNA-seq | Systematically compares models for lung cancer subtype classification | [110] |
| Ensemble model with | RNA-seq of bronchial | AUC (0.74) | RNA-seq of 1232 genes + clinical covariates | Integrates RNA-seq and clinical info for | [111] |
| elastic net LR, SVM, | brushing samples/2285 | | | improved risk prediction | |
| Linear SVM, polynomial-kernel SVM | RNA-seq/203 | AUC (0.8783-0.9980) | RFE & UAF selected genes | Improves classification accuracy with different gene selection algorithms | [112] |
| DT, KNN, linear/polynomial/RBF-kernel SVM, NN | CNV measured by CGH/37 | Accuracy (0.892) | Copy number of 80 genes selected by linear SVM | Systematically compares models for lung cancer subtype classification | [113] |
| HMM, weighted LS-SVM | CNV measured by CGH/89 | Accuracy (0.880-0.955) | CNV measured by CGH | High accuracy for cancer classification with recurrent HMMs for CNV detection | [114] |
| NN, SVM, RF | DNA methylation/972 | Accuracy (0.878-0.964) | Top 2000 variable CpG sites | Predicts tumor metastases using DNA methylation data | [115] |

### 2.4 AI in Cancer Immunotherapy: Enhancing Treatment Predictions and Personalized Approaches in lung cancer

Immunotherapy is currently a crucial component of cancer treatment. Immunocheckpoint inhibitors (ICIs), which target pmd1 (PD-L1) and programmed cell death protein 1 (PD-1) are used in blockade treatment. [116], have demonstrated efficacy in treating non-small cell lung cancer (NSCLC) [117, 118]. Immunotherapy has shown significant effectiveness and is now more widely used than traditional therapies including radiation therapy, surgery, and chemotherapy. The various responses observed across people can be attributed to the distinct characteristics of the tumor immune microenvironment (TIME) of each patient. Therefore, it is fundamental to accurately forecast whether a patient will exhibit a positive response to immunotherapy to efficiently manage cancer. The development of AI technology has made it possible to create systems that can predict how a patient will respond to immunotherapy.

AI-based methods utilize signatures obtained from medical imaging and immunological sequencing [119]. Wiesweg et al. [120] utilized gene expression patterns and cell type-specific genes to train an auxiliary support vector machine (SVM) classifier for RECIST categorization in response to PD-1/PD-L1 blocking therapy. Moreover, studies have investigated the utilization of contrast-enhanced computed tomography (CE-CT) scans and radiomics biomarkers and imaging features to educate classifiers for RECIST classification [121, 122]. Some well-known classifiers in this category include logistic regression (LR) and random forest (RF). Evaluating tumor-infiltrating lymphocytes (TILs) is an essential part of assessing the response to immunotherapy.

The CIBERSORT model was refined by DeepTIL [123] to determine the amount of leukocyte subsets in a tumor sample. In contrast, alternative studies utilized radiomics characteristics derived from CE-CT images and RNA sequencing data to predict the prevalence of specific T cell subsets [124]. In addition, a DL model was developed to identify tumor-infiltrating lymphocytes (TILs) in digitized pictures by including CNNs. This model employs distinct CNN modules to categorize lymphocyte infiltration and delineate necrotic segmentation [125]. ML algorithms have made significant advances to the prediction of immunotherapy response and the prognosis of neoantigens for immunotherapy. Tumor-specific peptides called neoantigens are altered and crucial for inducing antitumor-immune responses.

ML models, including NetMHC [126, 127] and NetMHCpan [128, 129], have been utilized to assist in the identification of immunogenic neoantigens [130, 131] by estimating the binding ability of mutant peptides to human leukocyte antigen (HLA) alleles. The aforementioned technologies have been useful in studying the neoantigens landscape in lung cancers, making substantial contributions to the progress and improvement of immune treatments targeting neoantigens [132, 133].

### 2.5 Limitation for Lung Cancer detection

Even with notable progress in the use of ML in clinical lung cancer research, there are still issues that will likely influence the direction of further studies in this area.

Accurately obtaining sensitive information from imaging data is essential for therapeutic applications. Unlike its predecessors, which used arbitrary feature extractions [134], DL-based CAD systems employ CNN models [135]. Significantly, the advent of Vision Transformer (ViT) [136] demonstrates potential in the analysis of imaging data, surpassing the efficiency and precision of conventional CNNs. On the other hand, DL models have shown effectiveness in other types of omics analysis. Genomics data analysis utilizes approaches such RNNs and CNNs.

In addition, to dealing with the complexity of omics data with many dimensions, autoencoders, deep generative models, and self-supervised learning models improve the process of extracting important features and reducing the number of dimensions.
While it is typical to gather many kinds of information from one patient, the process of combining data from different platforms is still difficult because of the interference caused by platform noise. Various medical classification networks aid in the analysis of a diverse array of medical pictures, while the incorporation of data integration and the removal of batch effects help reduce noise in sequencing data. In spite of continuous efforts to create integrative tools, batch effects continue to be a major source of worry and full answers to the problems unique to each platform in clinical investigations are still elusive.

The implementation of medical studies is hampered by the lack of robustness and generalizability in ML models. The differences observed in various researches regarding the choice of marker genes for categorization highlight the need for markers that are both biologically meaningful and applicable across different contexts. The efficiency and strength of models may be enhanced by employing transfer learning with pre-existing models. This approach provides a practical solution to create unified and standardized DL frameworks for analyzing lung cancer data. To ensure the ongoing advancement of ML in lung cancer research, it is crucial to address these challenges. This will ultimately result in better clinical results and individualized treatments for cancer [137-140].

## 3. AI In the prognosis of Breast Cancer

One of the most common and well-researched cancers is breast cancer, which has several subtypes identified by their molecular profiles [141]. The mortality rate for breast cancer among diagnosed women in 2022 was roughly 29.1%, resulting in 670,000 fatalities out of 2.3 million recorded cases [142]. Although there have been improvements in identifying and treating breast cancer, it continues to be a major issue in world health [143].

The primary focus of efforts to prevent and manage breast cancer is on secondary prevention, which entails improving screening for those who are at a high risk. Timely identification and medical intervention for breast cancer are also crucial in implementing these measures [144, 145]. The most significant benefit of AI in breast cancer screening is its ability to alleviate the workload of imaging specialists by efficiently identifying tumor lesions within the extensive image dataset of healthy individuals [146]. CADe/CADx systems serve as the basis for the advancement of AI-assisted breast imaging diagnostics. Radiologists depend on CAD, which is a prominent form of ML, to aid them in identifying small tumor lesions that may otherwise be overlooked. This is accomplished through the incorporation of mathematics, statistics, image processing, and computer analysis.

However, the use of CAD identification in clinical practice is limited due to the high occurrence of false positive results and the need for biopsies [147, 148]. The input picture data is used as a training set to build a model in order to improve the functioning of CAD. The system is capable of obtaining clinical data sets, standardizing the data sets using neural networks, choosing ML classification algorithms, and assessing its overall performance [149]. A substantial quantity of high-resolution breast examination photos is required as the training dataset for a deep learning-based AI application tool for the diagnosis of breast cancer [150]. Furthermore, it is imperative to create a DL system that maintains uniformity across different users, devices, and modes of operation. Tumor images can be annotated either manually or using DL techniques to label the characteristics and boundaries of the lesions. Annotating images manually necessitates the proficiency and understanding of an imaging specialist. The manually designated lesions act as reference points for the automatic segmentation process.

However, it can be difficult to distinguish between normal breast tissue and tumors that have small sizes or unclear characteristics. Furthermore, endo breast is more prevalent in Asian women, namely those who are premenopausal or have undergone estrogen replacement therapy [151] which further complicates tumor segmentation. Consequently, the lesions identified by experts do not accurately and thoroughly represent the true size and extent of the lesion, and the level of consistency in terms of being able to repeat the results is insufficient [152, 153]. Clustering-based, thresholding-based, region-based, edge-based, and other segmentation approaches are commonly used in automatic image segmentation methods. Currently, there is no widely recognized benchmark for image segmentation algorithms. DL algorithms are commonly linked with neural network algorithms.

CNN is distinguished among the most sophisticated DL algorithms because of its convolutional, nonlinear, pooling, and fully connected layers along the computation path (Figure 6). Applying AI to breast cancer screening has the ability to reduce or prevent the occurrence of errors or misinterpretations in identifying specific visible abnormalities [154]. Conversely, picture interpretation systems using DL are primarily used for the secondary assessment of negative cases that are evaluated by humans [155, 156]. Furthermore, there is a lack of randomized controlled studies that directly evaluate the accuracy of AI as an independent system for assessing breast cancer screenings with the interpretations made by radiologists. Subsequent examinations have revealed that AI systems exhibit worse accuracy compared to radiologists when it comes to interpreting data [157]. However, with the utilization of DL and training models on image data, AI still holds the potential to be employed in the future for the purpose of doing early breast cancer screening [150].Figure 6 explain the role of Al in the detection of breast cancer.

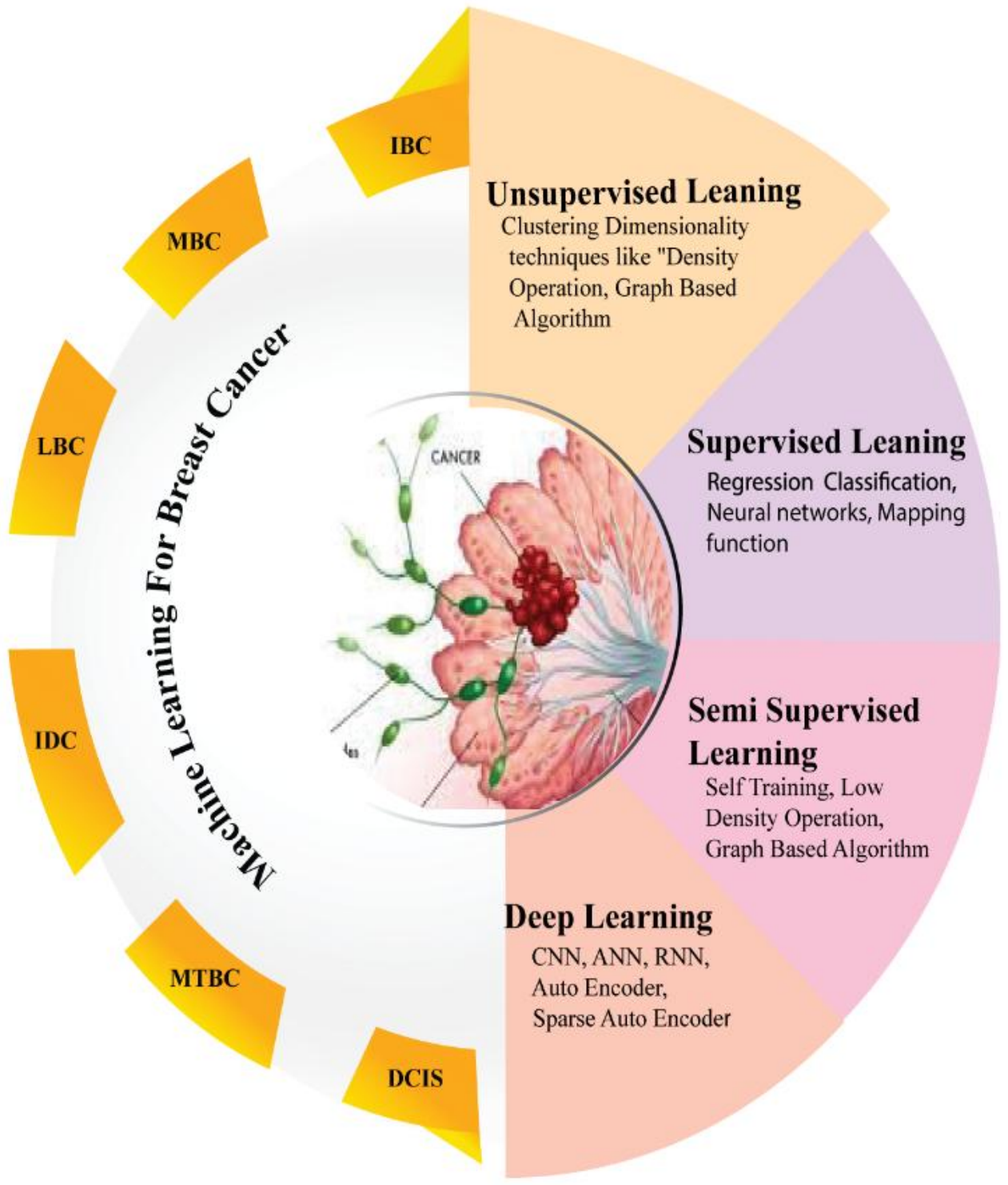

**Figure 6:** The diagram shows how AI and machine learning methods spanning supervised, unsupervised, semi-supervised, and deep learning enhance breast cancer management through early detection, precise diagnosis, personalized treatment, and improved survival outcomes.

### 3.1 Challenges in the current detection methods and Role of AI in Breast Cancer

Medical imaging is the most effective method for identifying breast cancer, utilizing various modalities such as MRI, CT, PET, mammography, radiographic ultrasonography, and duplex ultrasound. Medical images are essential for diagnosing diseases, identifying abnormal areas, providing patient treatment, and suggesting various disorders.

AI and ML may be responsible for the most recent advancements in medical image analysis. Breast cancer diagnoses are now more precise than ever before as a result of these advancements in early detection and evaluation of malignancies. Breast cancer detection methods include computed tomography, positron emission tomography, magnetic resonance imaging (MRI), ultrasonography, X-ray mammography, and breast temperature measurement [158]. Typically, pathology diagnosis is considered the most reliable method for detecting breast cancer. In this process, extracted tissue is stained in the laboratory to enhance visibility in the resulting images, with hematoxylin and eosin being commonly used stains [159]. There are two primary approaches for diagnosing breast cancer: genomics and histological image analysis. Histopathological images, which are microscopic photographs of breast tissue, are particularly valuable for early-stage cancer treatment. The limitations of current diagnostic methods are summarized in Table 3.

Table 3: Issues and Challenges

| Technique | Issues | Challenges |
|---|---|---|
| Mammography | • Not effective for women under 40.<br>• Difficulty detecting lesions in dense breasts and in those with surgical implants.<br>• Radiation exposure | • Limited dynamic range, low contrast, and grainy images<br>• Breast density affects sensitivity and specificity<br>• Age factor, rupture risk, and BRCA1/2 mutations |
| Breast MRI | • Not standardized<br>• Expensive<br>• Non-portable device<br>• Further biopsies needed | • Low specificity<br>• Slow imaging time<br>• Not recommended for patients with metallic devices<br>• Recommended only for high-risk women |
| Breast Ultrasound | • Difficulty covering the entire breast surface due to probe size<br>• Depends on the quality of the transducer and the physician's expertise in scanning and interpretation | • Poor resolution<br>• Low-contrast images |

Improved diagnosis and treatment sensitivity and accuracy have resulted from the use of AI in breast cancer diagnostics. A number of machine learning models have proven themselves to be useful in thermography data classification. These methods include meta classifiers that incorporate ANN, DT, Bayesian techniques, ELM, and MLP [160]. High accuracy in distinguishing between benign and malignant breast tumors has been proven by advanced DL models, including ResNet and VGG architectures [161, 162].

Furthermore, the use of CNN designs like Inception V3, ResNet50, and VGG-19, as well as support vector machines (SVM) and fuzzy C-means (FCM) algorithms, highlights the revolutionary role that AI is playing in the detection of breast cancer [163, 164] utilized a variety of models (DT, SVM, RF, K-NN) to reliably differentiate benign lesions from unsettling ones. Byra*et al.*[163] employed three convolutional layers and two connected layers to categories 166 individual tumors, of which 292 were non-cancerous lumps, as breast cancer. The accuracy rate attained was 83%. Xu *et al.*[165] proposed CNN to separate adipose tissue, epidermis, fibro glandular tissue, and 3D masses from breast ultrasound pictures. The quantitative metrics used to evaluate the effect of segmentation, such as precision, recall, and F1 scores, all exceed 80%. This demonstrates that the suggested method can correctly identify functioning tissue in a breast ultrasound image. To determine the most accurate method for classifying mammography images related to breast cancer, most researchers have utilized multiple strategies to categories ML for mammography photographs. Vijayarajeswari *et al.*[166] employed a Support Vector Machine (SVM) classifier to categories mammography images using features extracted from the Hough transformation, in order to determine the attributes of the mammograms.

The SVM algorithm was employed on a dataset of 95 clinical images. The findings demonstrate that the proposed approach successfully categorizes the challenging classes of mammograms; however, the accuracy is not precisely mentioned.

Hamoud *et al.*[167] applied the Fuzzy C-Means (FCM) method to the DDSM and MIAS datasets in order to classify tumors as either benign or malignant. The accuracy rate has been confirmed to be 87%, with a sensitivity value of 90.47% and a specificity value of 84.84%. The MIAS dataset was analyzed by [168] using CNN architectures including Inception V3, ResNet50, VGG-19, VGG-16, and Inception-V2 ResNet. The results of the classification technique for mammography breast pictures are as follows: The 10-fold cross-validation method and the 80-20 method achieved high precision, sensitivity, accuracy, and specificity. Specifically, the precision, sensitivity, accuracy, and specificity for the 10-fold cross-validation method were 0.995, 97.83%, 99.13%, and 97.35% respectively. Similarly, the 80-20 method achieved a precision, sensitivity, accuracy, and specificity of 0.995, 97.66%, 99.13%, and 97.35% respectively. The author obtained an accuracy of 92.84% by utilizing approaches on the DDSM dataset and incorporating transfer learning into CNN architecture. Table 4 provides a summary of the research that employ many machine learning models, such as SVM classifier, Artificial Neural Network (ANN), K-Nearest Neighbours (K-NN), Fuzzy C-Means, and CNN.

**Table 4:** Some latest Breast cancer classification, detection and segmentation models.

| Model | Diagnostic method | Dataset | Performance | Ref. |
|---|---|---|---|---|
| ROI, ANN | thermography images | Mastology Research | Accuracy is 90.2% Sn value is 89.34% Sp value is 91% | [169] |
| CNN | thermography images | Mastology Research | Accuracy is 98.95% | [161] |
| DWAN | thermography images | Mastology Research | Sn value: 0.95 | [170] |
| CNN based on VGG19 | Ultrasound | Private images 882 | Acc value is 88.7% TP value is 84.8% TN value is 89.7% AUC value is 93.6% | [163] |
| DT, KNN, RF SVM | Ultrasound | 283 owned cases | SVM accuracy is 77.7% AUC Value is 0.84 RF accuracy is 78.5% AUC value is 0.83 | [164] |
| Fuzzy C-Means (FCM) | mammogram images | DDSM | Accuracy is 87% Sn value is 90 to 47% Sp value is 84 to 84% | [167] |
| CNN | mammogram images | MIAS | Acc 92.1<br>AUC 95.0% | [171] |

# 4. AI in Colorectal Cancer (CRC)

Colorectal cancer (CRC) is the third most prevalent type of cancer in 2020, behind lung and breast cancers. It occupies the top position for males and the second position for girls. Annually, approximately 1,931,590 new cases of CRC are detected, accounting for 10.01% of all cancer cases.

CRC accounts for 17.6% of all cases in Asia. Contributes to 8.7% of all cancer-related fatalities. The rising incidence of CRC is a significant health issue in developing nations. Environmental factors contribute to the development of CRC, as shown by the variations in global incidence rates and the increased risk among immigrants who move from regions with low prevalence to regions with high prevalence within the same generation [172].

CRC manifests as an abnormal proliferation of epithelial cells, forming a tumor in either the colon or rectum [173]. The global incidence of cancer, specifically CRC, is on the rise. Genetic variations and environmental factors contribute to the probability of getting CRC. Furthermore, those diagnosed with colitis and Crohn's disease are inclined to get cancer, specifically CRC, as they age. Multiple academic research have revealed that diet and lifestyle choices, family history, and persistent inflammation are factors that add to the chance of developing cancer [174]. CRC is more prevalent in modern cultures, primarily due to the adoption of Western diets and lifestyles characterized by high consumption of meats, fats, and overall calorie intake. Changes in population composition and increased longevity are contributing to a wider range of factors affecting the situation [175].

### 4.1 Traditional Methods for the Detection of Colorectal Cancer

People over 50 are advised to have a digital rectal examination (DRE) at least once by the Indonesian Ministry of Health. If symptoms manifest, it may be necessary to do another examination [176]. The goal of DRE is to detect any abnormalities in the perianal region, the lower portion of the rectum, the anal canal, and the area between the rectum and vagina in females. Additionally, it assesses the functionality of the anal sphincter and examines the dimensions, connection, and mobility of any tumors located in the lower two-thirds of the rectum. This examination primarily focuses on the specific characteristics of colorectal cancer, such as the extent of its spread on the rectal wall, its location in respect to vital bodily structures, its mobility, and its metastasis. This information is vital for the purpose of surgical planning [177].

When performing a DRE, health care providers start by inquiring about any rectal symptoms, such as tumors, bleeding, or pain. Next, they thoroughly examine the region surrounding the anus and feel the bottom portion of the rectum as well as the inside of the anal canal. This efficient and cost-effective approach often requires less than one minute and is particularly effective in detecting tumors near the distal end of the rectum [177].

FOBTs are essential for the early identification of colorectal cancer as they detect occult blood in faces. Presently, there exist three types of FOBTs, distinguished by their methods of blood detection: chemical tests, immuno-chromatography testing, and DNA-based tests [178]. The guaiac-based faecal occult blood test (gFOBT) is a diagnostic test that detects the presence of haemoglobin in stool by assessing its pseudo-peroxidase activity. Although the gFOBT is beneficial, its specificity is relatively low, often requiring further tests for a correct diagnosis of colorectal cancer [179]. The 3,3′,5,5′-tetramethylbenzidine (TMB) test using a test paper, offers a simpler approach for patients as it undergoes a color change upon reacting with certain components in the stool. While this method is designed to be easy for users, it can only identify blood that is visible on the surface of the faces. This limitation increases the risk of errors in interpretation since it relies on visual assessment [180].

Endoscopic procedures are essential for screening CRC as they enable direct visualisation of the inner surface of the colon and rectum. Out all these options, colonoscopy is the most comprehensive method, as it involves a direct inspection and is linked to significant reductions in the risk of CRC and mortality rates.

Studies conducted in Canada and Germany have revealed that following a colonoscopy, there is a 67% reduction in death rates and a 77% decreased risk of acquiring CRC [181]. Colonoscopy negative results were associated with a 56% decreased overall risk of colorectal cancer in the United States, according to a long-term study that followed participants for over 22 years [182]. Although there are differences in the effectiveness of identifying proximal versus distal CRC, depending on things like the operator's expertise, the difficulty of visualizing the condition, and the thoroughness of the inspection. Notwithstanding its advantages, colonoscopy has several disadvantages as well, such as the discomfort of preparing the intestine, the dangers of anesthesia, and the possibility of procedural complications such as bleeding and perforation. The risks of bleeding after surgery and perforation are highlighted by meta-analyses, which emphasize the importance of careful patient selection and operational preparation [183].[184].

A long-term study conducted in the United States that tracked participants for over 22 years discovered that those who had negative colonoscopy results had a 56% decreased overall risk of colorectal cancer [185]. For CRC screening, the FDA is in favor of stool-based testing, especially when it comes to the combination of faecal immunochemical test (FIT) and multitarget stool DNA (Fit-DNA). Fit-DNA is more sensitive than FIT alone in detecting colorectal cancer (CRC), but it is less specific, which could lead to more false positives and require additional colonoscopy research [186]. The optimal screening frequency for FIT-DNA is yet unknown, but it is recommended to be performed every three years in the United States. Its greater price in comparison to FIT may also prevent it from being used in less developed areas [181]. The development of CT colonography stands out as a noteworthy advancement in radiology-focused colorectal cancer (CRC) screening.

Its efficacy exceeds that of conventional barium enemas; with a sensitivity rate of 67–94% and a specificity of 86–98%, it demonstrates greater sensitivity in detecting adenomas of at least 1 cm in size [187]. The excellent sensitivity of CT colonography, which is on par with colonoscopy for identifying polyps larger than 5 mm, is highlighted by research conducted in Europe. It also highlights a noticeably reduced perforation risk related to this technique. The capacity of CT colonography to identify tiny polyps is inferior to that of colonoscopy. The need for intestinal preparation and radiation exposure are two major disadvantages. Despite these problems, CT colonography is recommended, especially for individuals who are more likely to experience colonoscopy-related difficulties [179].

A new method for CRC screening called liquid biopsy examines blood samples for protein markers, circulating tumor cells (CTCs), and cell-free tumor DNA (ctDNA) [188]. The FDA has granted approval for the methylation DNA septin9 test, which is the first serum test for CRC screening. This test has a sensitivity of 48.2% and a specificity of 91.5%. The sensitivity of CRC stages I-IV ranges, with percentages of 35.0%, 63.0%, 46.0%, and 77.4%, respectively. However, for advanced adenomas, the sensitivity is significantly low at only 11.2% [189]. The Septin9 test offers the benefit of being both uncomplicated and non-invasive, hence increasing the likelihood of more participation among individuals [190]. The following are the conventional techniques employed for the identification of CRC.

### 4.2 Limitations of Current Detection Methods for CRC:

Early detection of CRC is crucial for enhancing patient outcomes and increasing survival rates. Healthcare professionals must have a comprehensive understanding of the various traditional methods that have been developed in recent years to detect CRC at its earliest stages. This knowledge is crucial for determining the most suitable screening strategy for each patient, considering their individual needs and circumstances. Additionally, it is important for healthcare professionals to stay informed about the current and future advancements in CRC screening. Table 5 provides a comprehensive summary of the various methods used to CRC globally. These methods include endoscopic, radiologic, and stool-based tests, as well as physical examination and emerging blood-based screening tests. The table also includes information on the applications and limitations of each method, along with primary references. This overview aims to offer a clear understanding of the current landscape of CRC screening strategies.

**Table 5:** Traditional Methods for the Detection of CRC.

| Traditional Method | Application | Limitations | Ref. |
|---|---|---|---|
| Digital Rectal Examination (DRE) | Recommended for individuals over 50 years old or with clinical symptoms. Assesses abnormalities in the perianal area, distal rectum, and evaluates tumor size, mobility, and extension for surgical planning. | Limited reach, depends on examiner's expertise, more valuable for locally advanced tumors than early-stage tumors. | [177] |
| Fecal Occult Blood Tests (FOBTs) | Early detection tool identifying unseen blood in feces. Includes chemical, immunochromatography, and DNA-based tests. | gFOBT prone to false positives and requires dietary restrictions. TMB test limited by interpretative errors and only detects blood in stool's outer layer. | [180] |
| Endoscopic Methods | Direct visualization of the colon and rectum's mucosal lining. Colonoscopy is linked to significant reductions in CRC risk and mortality. | Discomfort from bowel preparation, risks of sedation, perforation, and bleeding. Sigmoidoscopy does not assess the entire colon. | [185] |
| Stool-based Testing | FDA endorses multitarget FIT-DNA combined with FIT for CRC screening, recommended at three-year intervals in the US. | FIT-DNA shows higher sensitivity but lower specificity than FIT alone, leading to more false positives. Higher cost could hinder adoption. | [186] |
| Radiologic Method: CT Colonography | Superior sensitivity in detecting adenomas of at least 1 cm compared to traditional barium enemas. Recommended every 5 years for individuals at elevated risk of colonoscopy-related complications. | Exposure to radiation, need for bowel preparation, and lower sensitivity for smaller polyps compared to colonoscopy. | [191] |
| Blood-based Screening Tests (Liquid Biopsy) | Analyzes circulating tumor cells, cell-free tumor DNA, and protein markers in the blood. The FDA approved the methylated DNA septin9 test for CRC screening. | Low sensitivity for detecting CRC and advanced adenomas, hindering early detection efforts. | [190] |

## 4.3 The Role of AI in Enhancing CRC Detection

AI, particularly utilizing ML and DL techniques, has substantial potential in overcoming the limitations of conventional approaches for identifying CRC, as depicted in Figure 7. By examining large datasets obtained from colonoscopy images, radiological scans, and genetic tests, AI systems may accurately detect signs of CRC.

This system decreases reliance on human operations in endoscopic procedures by providing standardized assessment tools, while also improving early detection capabilities by combining data from various sources.

Furthermore, AI has the capacity to improve the accuracy and precision of non-invasive testing and can offer personalized screening schedules based on individual risk factors. This can lead to better utilization of resources and improved outcomes for patients [192].

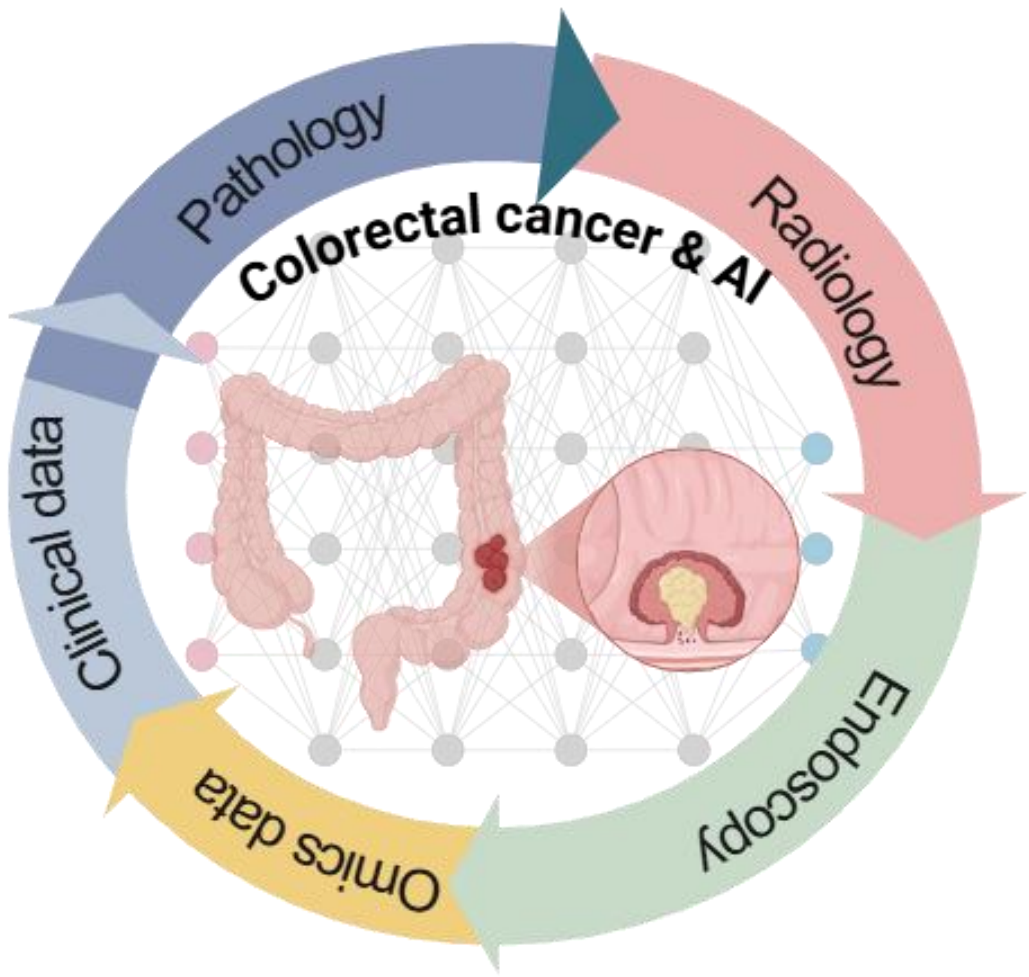

**Figure 7:** This diagram illustrates the integration of AI into colorectal cancer analysis. It synthesizes data from six sources: pathology, radiography, endoscopy, clinical data, omics data, and AI algorithms. Together, these components facilitate a comprehensive approach to the diagnosis and management of colorectal cancer.

The application of AI in the clinical field of CRC is facilitated by its capacity to identify patients at high risk, choose accurate and personalized treatment strategies, and forecast prognoses. This demonstrates the present and future therapeutic relevance of AI technologies in the management of CRC [191]. Implementing these methods in clinical practice has the capacity to enhance colorectal cancer (CRC) screening, detection, and management. This can result in earlier detection, decreased burden from screening procedures, and improved outcomes for individuals at risk of CRC.

AI technology has lately made significant advancements in the diagnosis of CRC. Despite being a long-standing cornerstone of early colorectal cancer (CRC) screening, colonoscopy has faced considerable difficulties in detecting polyps and adenomas due to inconsistencies among operators and inadequate preparation. Computer-aided detection (CADe) systems, along with AI techniques such as DL, have been proven to overcome these challenges and improve the adenoma detection rate (ADR) [193]. The use of AI technology, specifically YOLOV3 AI, in colonoscopy has significant potential, particularly in enhancing real-time ADR.

The use of AI technology, specifically YOLOV3 AI, in colonoscopy has significant potential, particularly in enhancing real-time ADR.

This solution could also resolve the identified limitation of not being able to completely utilize high-definition endoscopy due to limited adoption of HD technology or integration with colonoscopies [191]. Furthermore, developments such as YOLOV3 might provide a substitute for current techniques that find it difficult to obtain significant ADR improvements using CADe and colonoscopies [193]. The end result might be an easier-to-use, more affordable system for improving ADR that can be implemented more widely and with limited resources, especially in situations with limited resources [194]. An expanded innovation pipeline in place of the current flat colonoscopy monitor may also greatly enhance the identification of tiny, easily missed polyps. Increased utilization of CADe technology has the potential to reduce the occurrence of colorectal cancer by up to 53 percent by ensuring that every polyp is recognized and not missed [195]. The recent implementation of the CNN algorithm has reduced the adenoma miss rate (AMR) to 7%. This indicates an improvement in the accuracy of AI-enabled detection of lesions that were previously overlooked by traditional methods [174].

AI has extensive promise that extends beyond its use in endoscopic applications. Non-invasive approaches for colorectal cancer (CRC) screening provide a promising approach to detect tumor indicators in bodily fluids. These methods offer a safer alternative to traditional colonoscopy, as they do not require the same strict preparation procedures [196]. ML-type technologies have been used to find new potential markers for early colorectal cancer diagnosis as well as to increase the sensitivity and specificity of well-established markers including faecal occult blood tests (FOBT) and carcinogenic embryonic antigen (CEA) [193]. AI has also provided benefits to the field of diagnostic histopathology. The combination of DL and pattern recognition in digital pathology has improved diagnostic accuracies in this sector. This has resulted in a decrease in wrong diagnoses and an enhancement in the efficiency of pathologists' workflow [197]. AI-enhanced radiomics have revolutionized medical imaging for colorectal cancer by enabling the accurate identification and measurement of tumors, leading to greater diagnostic and prognostic capacities [198].

The advancements mentioned highlight the significant impact that AI can have on the diagnosis and treatment of CRC. This includes enhancing the accuracy of endoscopic procedures, enabling non-invasive screening, and improving Histopathological and radiological assessments. It is reasonable to anticipate that AI will continue to advance CRC management as these skills develop quickly, democratizing and optimizing the benefits of early identification and treatment. The development of AI has brought a fresh perspective to the treatment of CRC, adding a new dimension to established techniques including chemotherapy, targeted therapy, surgery, and their combination.

#### 4.3.1 Surgical Innovations of CRC by AI

In recent years, there have been significant breakthroughs in minimally invasive surgical techniques. However, the integration of AI has emerged as a revolutionary development in lung cancer surgeries and is now gaining recognition in the complicated field of CRC surgery [175].

The latest advancement in surgical practices in Japan involves groundbreaking AI research; specifically in the field of computer vision (CV) applied to laparoscopic surgery videos. Researchers in Japan have analyzed extensive video data from laparoscopic colorectal surgeries to enhance the performance of CNN in computer vision [179]. Simultaneously, South Korean research has shown nearly flawless precision in the study of microcirculation. This was achieved by employing AI to develop versatile and very advanced learning models for the examination of indocyanine green (ICG) angiography in laparoscopic surgeries [199].

AI has had significant growth in the field of robotic surgery, namely in the domain of rectal surgery. The advantages of using a robotic system compared to traditional approaches include a decrease in perioperative problems and an enhancement in postoperative quality of life. Robotic systems are currently advancing, specifically in relation to the da Vinci platform. Subsequent generations have enhanced the accuracy and adaptability options for patients. Nevertheless, further efforts are required to address the issues of operation time and range of motion [200].

### 4.3.2 Chemo-radiotherapy Advances of CRC by AI

AI has significant effects on the field of chemoradiotherapy. Significantly, a significant amount of the decision-making and assessment of effectiveness in the treatment of neoadjuvant chemoradiotherapy (NCRT) patients is impacted by AI. The utilisation of AI in the construction of Clinical Decision Support Systems (CDSS) is a significant achievement that improves clinical decision-making. An example of this is seen in South Korea's utilisation of a CRC chemotherapy prescription system, which displayed a remarkable level of precision (AUC > 0.95).

Moreover, AI-powered systems that aid in risk assessment following colectomy are enhancing the capacity to detect individuals who can safely skip NCRT, a characteristic that is expected to have a substantial impact on improving survival rates [201].

### 4.3.3 Personalized Medicine through AI in CRC

AI has greatly influenced personalized medicine, specifically in identifying individuals who could potentially benefit from targeted medicines aimed at EGFR sensitivity for KRAS mutations. The use of DL to non-invasively determine the status of KRAS mutations demonstrated good prediction abilities (AUC = 0.90). Further obstacles in treating CRC include the emergence of drug resistance and the need to create targeted drug delivery systems and predictive models for treatment resistance. These challenges pave the way for precision medicine, which aims to provide the most suitable treatment to each patient at the optimal time [202]. Table 6 summarizes the CAD method.

**Table 6:** Comparative Analysis of Computer-Aided Detection Systems in Colonoscopy.

| CADe System | Control | Patients (n) | ADR (AI vs. Control) | Advanced ADR | Ref. |
|---|---|---|---|---|---|
| CAD EYE | HD-WLI | 415 | 59.4% vs. 47.6% (p = 0.018) | 7.2% vs. 7.7% (p = 1) | [203] |
| Eagle-Eye | HD-WLI | 3059 | 39.4% vs. 32.4% (p < 0.001) | 6.6% vs. 4.9% (p = 0.041) | [204] |
| Endo Screener | HD-WLI | 1261 | 25.8% vs. 24.0% (p = 0.464) | 0.314% vs. 0.39% (p = 0.562) | [205] |
| Endo Vigilant | HD-WLI | 769 | 37.2% vs. 35.9% (p = 0.774) | N/A | [206] |
| GI Genius | HD-WLI | 658 | 71.4% vs. 65.4% (p = 0.09) | N/A | [207] |
| ENDO-AID | HD-WLI | 370 | 55.1% vs. 43.8% (p = 0.029) | 11.6% vs. 12.1% (p = 0.89) | [208] |
| GI Genius | HD-WLI | 660 | 53.3% vs. 44.5% (p < 0.02) | 12.7% vs. 12.7% (p = 0.956) | [209] |
| CAD EYE | HD-WLI | 800 | 53.6% vs. 45.3% (RR 1.18, 95% CI 1.03–1.36) | 15.9% vs. 15.8% (RR 1.03, 95% CI 0.96–1.09) | [210] |
| SKOUT | HD-WLI | 1359 | 47.8% vs. 43.9% (p = 0.065) | N/A | [211] |
| Xiamen Innovation | HD-WLI | 150 | PDR 38.7% vs. 34.0% (p < 0.001) | N/A | [212] |
| N/A | HD-WLI | 2352 | PDR 38.8% vs. 36.2% (p = 0.183) | N/A | [213] |
| Endo Screener | HD-WLI | 790 | 29.0% vs. 20.91% (p = 0.009) | 1.43% vs. 3.92% (p = 0.607) | [214] |

CAD systems have been developed to improve colorectal cancer diagnostics. They represent a significant advancement in colorectal cancer diagnosis and treatment, providing a combination of imaging techniques and computational algorithms to improve polyp detection, classification, and histological analysis. Table 7 presents a summary of the most innovative works in the past decade, encompassing personalized and multimodal CAD programmers for analyzing endoscopic pictures and narrow-band imaging samples. This compilation aims to advance the area.

**Table 7:** CAD System Development Techniques for Endoscopic Image Analysis

| Objective | Methods Used | Image Type | Maximum Results | Ref. |
|---|---|---|---|---|
| To develop CAD systems for the classification of colorectal polyps. | Texture analysis, k-NN | Endoscopic images | Sensitivity = 98.2%, Specificity = 87.5% | [210] |
| Automated and fully developed system for the classification of colorectal polyps and mucosal abnormalities. | SVM, texture analysis, shape classification | Endoscopic images | Sensitivity = 98.2%, Specificity = 87.5% | [211] |
| To develop an objective classifier of colorectal polyps to | CNN extraction, gray-scale conversion, feature extraction | Endoscopic images | Accuracy = 93.5% | [215] |
| reduce the workload of endoscopists. | | | | |
| To evaluate the effectiveness of computer-aided diagnosis for endoscopic images using NIR imaging of colorectal adenomas. | ROI extraction, grayscale conversion, energy feature extraction, Bayesian methods | Near-infrared (NIR) images | Sensitivity = 88.3%, Specificity = 82.6% | [213] |
| To classify NIR imaging of colorectal adenomas based on CNNs and their effectiveness. | Feature extraction using texture analysis | Near-infrared (NIR) images | Sensitivity = 94.3%, Specificity = 89.8% | [214] |
| To provide histological diagnosis of colorectal lesions detected on narrow-band imaging samples. | Feature extraction using random forest classifier from a two-stream CNN | Narrow-band imaging samples | PPV (Positive prediction value) = 95.4%, NPV (Negative prediction value) = 93.8% | [216] |
| To investigate the feasibility of computer algorithms to analyze early Barrett's esophagus (BE) volume laser endomicroscopy (VLE) images. | Processing of input images, feature extraction with gray-level co-occurrence matrix, local binary pattern, gradient orientation, watershed segmentation, SVM detection, linear SVM, KNN, decision trees recognition, and logistic regression. | Volumetric laser endoscopy images | Sensitivity = 96%, Specificity = 95%, AUC = 0.95 | [217] |
| To develop a CAD system based on deep learning imaging (DLI) methods to assist in the diagnosis of early-stage polyps by analyzing cell patterns. | Processing of input images, feature extraction from CNN to HOG and bag-of-visual words, k-means clustering, SVM classification using a heat map orientation, and identification using a CNN. | High-definition white-light imaging samples | Sensitivity = 83.5%, Specificity = 81%, PPV = 79.3%, NPV = 72.1% | [218] |

AI significantly enhances the diagnostic armamentarium, although non-invasive screening remains limited by marker constraints. Ethical and methodological challenges, such as data privacy, algorithmic bias, and the need for thorough clinical validation, highlight the complexities of AI integration into colorectal cancer (CRC) management. Continued interdisciplinary collaboration and ethical oversight are crucial for responsibly harnessing AI's full potential.

The incorporation of AI into CRC screening marks a significant turning point with far-reaching implications. AI effectively resolves the issues associated with traditional screening approaches. Machine-learning and deep-learning models enhance the precision of diagnoses in endoscopy, increase the rates at which adenomas are detected (ADR), and decrease the occurrence of missed adenomas (AMR). The use of AI improves the accuracy of non-invasive screening in detecting tumor markers. Upon the detection of CRC, AI enables the development of customized treatment strategies, resulting in enhanced patient outcomes.

AI enhanced robotic and laparoscopic surgeries are improving accuracy and lowering risks. AI streamlines the process of making decisions about chemoradiotherapy in medication development by using a variety of data sources to enhance the assessment of effectiveness. In addition, the predictive analytics of AI are transforming targeted therapy, overcoming the longstanding problem of drug resistance that has posed a significant barrier to oncologists. Although AI has had a significant influence on the detection and management of CRC, its integration into clinical practice must be approached with careful consideration of scientific and ethical issues.

Important priorities include of establishing strong data protection regulations, tackling potential biases in algorithms, and guaranteeing thorough clinical validation. The increasing utilizations of AI in the detection and treatment of colorectal cancer highlights the importance of interdisciplinary collaboration and ethical evaluation in order to fully achieve its clinical potential.

## 5. Liver Cancer and AI

The most prevalent form of primary liver cancer is hepatocellular carcinoma (HCC), which is one of the leading causes of cancer-related fatalities worldwide.  Liver fibrosis can result from chronic inflammation, which in turn creates an environment where profibrogenic extracellular matrix (ECM) proteins accumulate excessively. This leads to a disruption of the liver's structure and function, which is then replaced by immunologically inert matrix. This process can progress to cirrhosis and ultimately liver failure, with the associated morbidity and mortality that make liver transplantation the most common reason for liver replacement [219]. The importance of early identification of HCC is evident due to its direct correlation with enhanced patient outcomes, particularly in persons at high risk, such as those with a previous record of chronic hepatitis B or C virus infection.

Utilizing non-invasive biomarkers such as AFP, DCP, and GPC3 in conjunction with state-of-the-art imaging techniques such as MRI and ultrasound [220], have revolutionized the early detection landscape, increasing sensitivity and specificity for diagnosis of HCC and leading to multi-modality algorithms which are optimum for predicting early tumor recurrence Imaging technology has made considerable advancements in the past decade, and one promising option is contrast-enhanced ultrasound (CEUS). The technology offers a clear and detailed view of the liver and its blood vessels in real-time, with high resolution. It also gives enhanced accuracy in detecting HCC and allows for precise measurement of tumor margins during surgery, surpassing the capabilities of standard ultrasound [221].

Furthermore, in addition to conventional imaging techniques, radiomics features can serve as a supplementary tool to enhance the precision of HCC diagnoses by assisting AI-based algorithms [222]. The von Willebrand factor (vWF) is now a critical prognostic indicator that influences the efficacy of systemic treatment, the progression of conditions after HCC removal, and the risk of HCC in cirrhotic patients. There is a possibility that vWF may affect acute liver failure and non-cirrhotic portal hypertension due to its involvement in the production of portal microthrombi. Non-selective beta-blockers, statins, anticoagulants, and antibiotics that are not ingested by the body are among the medications that have the capacity to regulate vWF levels. [223]. Recent advancements in radiomics models, particularly those utilizing multi-sequence MRI, have demonstrated potential in predicting the expression of PD-1/PD-L1 in HCC). This predictive capability could be valuable in informing decisions on immunotherapy treatment [224]. These are a few examples of the changing ways in which AI could have a significant impact, not just in identifying HCC, but also in using biomarkers to diagnose it.

### 5.1 Current Diagnostic Approaches for the Detection of Liver Cancer

Appropriate screening methods are essential for detecting HCC early on, as this improves patient outcomes. While there is no universally acknowledged screening strategy for HCC, ultrasound has proven to be the most cost-effective and commonly used tool for early diagnosis of HCC, particularly in high-risk categories such patients with cirrhosis [225]. As a result, efforts to find more accurate and fast early HCC detection and diagnosis tools, including enhanced biomarkers and cutting-edge imaging methods, have persisted [226].

Various biomarkers have been extensively studied in the ongoing pursuit of enhancing diagnostic precision. Alpha-fetoprotein (AFP), des-gamma-carboxy prothrombin (DCP), and glypican-3 (GPC3) have demonstrated encouraging outcomes as diagnostic or prognostic indicators for hepatocellular carcinoma (HCC). More accurate and early identification of HCC may be possible with the emergence and development of additional new biomarkers like long non-coding RNAs and microRNAs.

Advancements in targeted medicine and bioengineering have revolutionized the treatment of hepatocellular carcinoma, allowing for personalized therapy. Utilizing biomarkers, genetics, nanotechnology, and drug delivery technologies in the creation of customized treatment plans yields improved patient outcomes. Two urgent problems that demand attention are the need for more accurate diagnostic techniques and the advancement of innovative targeted therapies for HCC [227].

#### 5.1.1 Diverse Screening Approaches for liver Cancers Detection

Several screening methods have been studied for the identification of HCC, such as ultrasonography, computed tomography (CT), magnetic resonance imaging (MRI), and serum biomarkers like AFP and DCP. The American Association for the Study of Liver Diseases (AASLD) state that ultrasound, with or without AFP, is the recommended bi-annual screening test for populations at high risk of HCC. Ultrasound (US) in conjunction with AFP using ML algorithms have both shown significant improvements in sensitivity and specificity for HCC detection [228].

#### 5.1.2 Molecular Biomarkers for Liver Cancer

It has been discovered that more novel serum indicators are helpful for HCC screening. For instance, plasma glypican-3 (GPC3) is a novel biomarker for HCC screening that shows tremendous promise in the identification of early-stage HCC [229]. In fact, compared to various serum AFP readings, plasma GPC3 has a diagnostic sensitivity that is noticeably higher for identifying early-stage HCC. Additionally, serum hepatocyte growth factor (HGF) levels are inversely linked with prothrombin time and albumin concentration in individuals with cirrhosis; as a result, serum HGF may be utilized as an extra helpful biomarker for HCC screening [230]. CtDNA analysis has become a highly sensitive and specific diagnostic technique for HCC diagnosis [231]. Moreover, a combined radiomics model that incorporates DL-informed multisequence MRI produces excellent discriminatory performance in anticipating early recurrence for early-stage HCC as well as PD-1/PD-L1 expression [232]. AI algorithms and ML models have the potential to significantly improve diagnostic efficacy and accuracy for even less experienced physician readers, in addition to their ability to detect HCC early [233].

### 5.1.3 Imaging Techniques for Liver Cancer Detection

Imaging modalities play a critical role in the early diagnosis and detection of (HCC), the second most common cause of cancer-related deaths globally, as well as in the prediction of treatment outcomes and evaluation of treatment response.
A review of the research published in the Journal of Hepatology in May 2021 states that MRI and CT scans are essential for the diagnosis of HCC.

Particularly, three-dimensional multifrequency magnetic resonance elastography and multi-sequence MRI are becoming useful instruments to enhance preoperative evaluations and forecast PD-1/PD-L1 expression in HCC [221, 234]. Therapeutic decision-making and treatment stratification depend on this interpretation. Long recognized for its capacity to create preoperative prediction models for macro trabecular-massive HCC, contrast-enhanced CT has also made a substantial contribution to diagnostic accuracy by acting as the gold standard for numerous complex and real-world scenarios, such as the United Network for Organ Sharing's imaging criteria for HCC when screening liver transplant candidates [235]. Furthermore, using baseline MRI data, deep machine learning models demonstrated high prediction accuracy for early-stage HCC tumor recurrence, demonstrating the possibility of combining AI with imaging data to improve diagnosis [234].

With the use of microbubble contrast agents, contrast-enhanced ultrasound (CEUS) has shown to be a valuable diagnostic tool for HCC. This technique is especially helpful for patients with cirrhosis, for whom other forms of imaging may not be as helpful [236]. It has shown to have good sensitivity and specificity in identifying tiny HCC nodules and distinguishing HCC from other liver lesions. It has been shown to be very helpful in determining the course of treatment after locoregional therapy [237]. CEUS surveillance has been proven to be more sensitive than CT and MRI in detecting tiny nodules in the early stages of HCC. It has also been demonstrated to detect early alterations in liver perfusion that take place throughout the development of HCC [238-240].

### 5.1.4 Liquid Biopsy in Liver Cancer Detection

A developing non-invasive method for diagnosing HCC is called liquid biopsy, and it entails looking for cancer-related biomarkers for HCC in extracellular vehicles (EVs), circulating tumor cells (CTCs), and cell-free DNA (cfDNA). Compared to traditional tissue biopsy, it has many benefits, such as a lower risk of complications and the ability to monitor disease development and therapeutic response more dynamically [226]. When it comes to detecting HCC, CTCs are quite accurate; their sensitivities and specificities are frequently greater than 80% [240],” Dr. Wong elaborated. “Likewise, it has recently been discovered that increased ctDNA levels are linked to the development of HCC as well as noticeably worse results [228], providing key proof of concept and a potential biomarker for early diagnosis. EVs have also demonstrated outstanding diagnostic accuracy for HCC [241].

### 5.1.5 Histopathological Analysis for Liver Cancer

Nothing is more accurate than using liver biopsy samples for histological investigation to diagnose hepatocellular carcinoma (HCC). This technique looks deeply into the liver's cellular structure to identify HCC rather than merely examining cells under a microscope. Its capacity to provide the full picture of cellular morphology—including the sizes, shapes, and configurations of cells that are indicative of HCC sets it apart.

This data is not only essential for verifying the existence of a tumor, but it is also critical for establishing the tumor's levels, which in turn directs the course of treatment. Histopathological examination is another method that can be used to assess the degree of fibrosis, which is crucial in determining the optimal course of treatment. Furthermore, this method also provides security in disclosing diagnoses of liver-related disorders other than HCC (such as necrosis, steatosis, or inflammation). Considering the high threshold for confirming HCC, liver biopsy carries inherent risks and potential consequences.

The process of confirming HCC may involve several problems, including substantial discomfort and the potential for bleeding, infection, and mortality, which occurs in less than 0.1% of patients. Additionally, 5% of the patients necessitate an extended hospitalization period for observation [242].The hunt for alternate, less invasive methods to diagnose HCC is gaining traction. Traditional liver biopsy has been replaced by more advanced imaging techniques and liquid biopsy, which are newer methods of detecting HCC. It is highly prospective that these secure, non-invasive methods can be employed to diagnose HCC [221, 226]. These changes could significantly reduce the hazards associated with the resolution of biopsy operations and increase the chances of early detection and timely intervention. Although non-invasive diagnostics have made significant advancements, histopathological analysis remains necessary for a comprehensive characterization of HCC. This is because it is the sole method for identifying prognostic biomarkers such as tumor grade and vascular invasion, which play a crucial role in assessing the severity of the disease and guiding treatment decisions [225].

There are multiple unusual variations of HCC, and the assessment of these variations relies heavily on histological examination. Cirrhotomimetic hepatocellular carcinoma is a rare type of HCC that imitates the characteristics of cirrhosis. This poses a diagnostic challenge, highlighting the need for thorough imaging and pathological testing to ensure accurate diagnosis and suitable treatment [243]. Due to its remarkable ability to replicate the pseudocirrhotic pattern of HCC, cirrhotomimetic HCC is very intriguing. The repercussions of failing to identify these unusual patterns in the differential diagnosis of liver masses are amplified in a patient with a documented history of chronic liver disease.

To accurately characterize and treat these rare cases, a comprehensive strategy that combines thorough cross-sectional imaging with histopathologic examination is necessary.

The limited and incomplete evidence we have on the origin of cirrhotomimetic HCC, whether it arises from hepatic adenoma or develops spontaneously, contributes to the overall occurrence of HCC. Amidst the era of swiftly expanding treatment alternatives, the timely identification and accuracy in diagnosis play an increasingly crucial part in enhancing patient outcomes [244].

## 5.2 Limitations of the Current Traditional Detection Method for Liver Cancer

The diagnostic approaches for HCC exhibit varying limits across a range of modalities. Ultrasonography is a cost-effective and non-invasive technique. However, it may be challenging to detect small lesions of early-stage HCC in patients with cirrhosis [225]. Advanced imaging methods such as MRI and CT provide higher levels of sensitivity and specificity. However, they need significant resources and are not easily accessible in places with limited resources. Additionally, there are possible dangers connected with the use of contrast agents [221, 235]. Gastrointestinal endoscopy has included contrast-enhanced ultrasonography (CEUS) to enhance the identification of HCC, although its efficacy is significantly influenced by the skill of the operator and the characteristics of the patient [239]. Histopathological examination is the established and authoritative method for diagnosing HCC. The invasiveness of liver biopsies is correlated with the likelihood of complications [243, 244]. The biological attributes of the tumor can influence the accuracy and precision of identifying biomarkers, presenting obstacles to the use of liquid biopsy as a non-invasive diagnostic alternative [243]. Collectively, these diagnostic modalities emphasize the importance of integrated approaches that prioritize effectiveness, safety, and availability in the management of HCC. Table 8 presents an overview of the existing detection techniques and their limitations.

**Table 8:** Limits of Current Detection Techniques for Hepatocellular Carcinoma (HCC).

| Detection Technique | Limitation | Ref. |
|---|---|---|
| Ultrasound (US) + serum α-fetoprotein (AFP) | Low sensitivity and specificity for early-stage HCC. | [245] |
| Multiphasic computed tomography (CT) or magnetic resonance imaging (MRI): | High cost, limited access, and requires contrast that may have contraindications | [246] |
| Hepatocellular carcinoma Early detection Screening (HES) algorithm | Slight improvement over AFP alone; further validation required | [247] |
| Circulating biomarkers (e.g., AFP, AFP-L3, DCP) | Single biomarkers often lack the needed sensitivity and specificity for early detection | [245] |
| Novel multi-target blood test incorporating methylation biomarkers and proteins | Requires additional, large-scale validation before widespread clinical adoption | [248] |
| Liquid biopsy technologies (e.g., circulating tumor DNA) | Still in the research phase with challenges concerning standardization and sensitivity for early HCC. | [249] |

## 5.3 Role AI in the Liver Cancer Detection

AI is becoming increasingly crucial in the identification of liver cancer, with a diverse range of applications that greatly enhance the field of medical imaging analysis, predictive modeling, diagnosis, treatment planning, and even the continuous management of patients' care. A recent study provides valuable insights into the application of AI in the diagnosis and treatment of liver cancer. The article emphasized the utilization of AI in the identification and control of liver cancer.

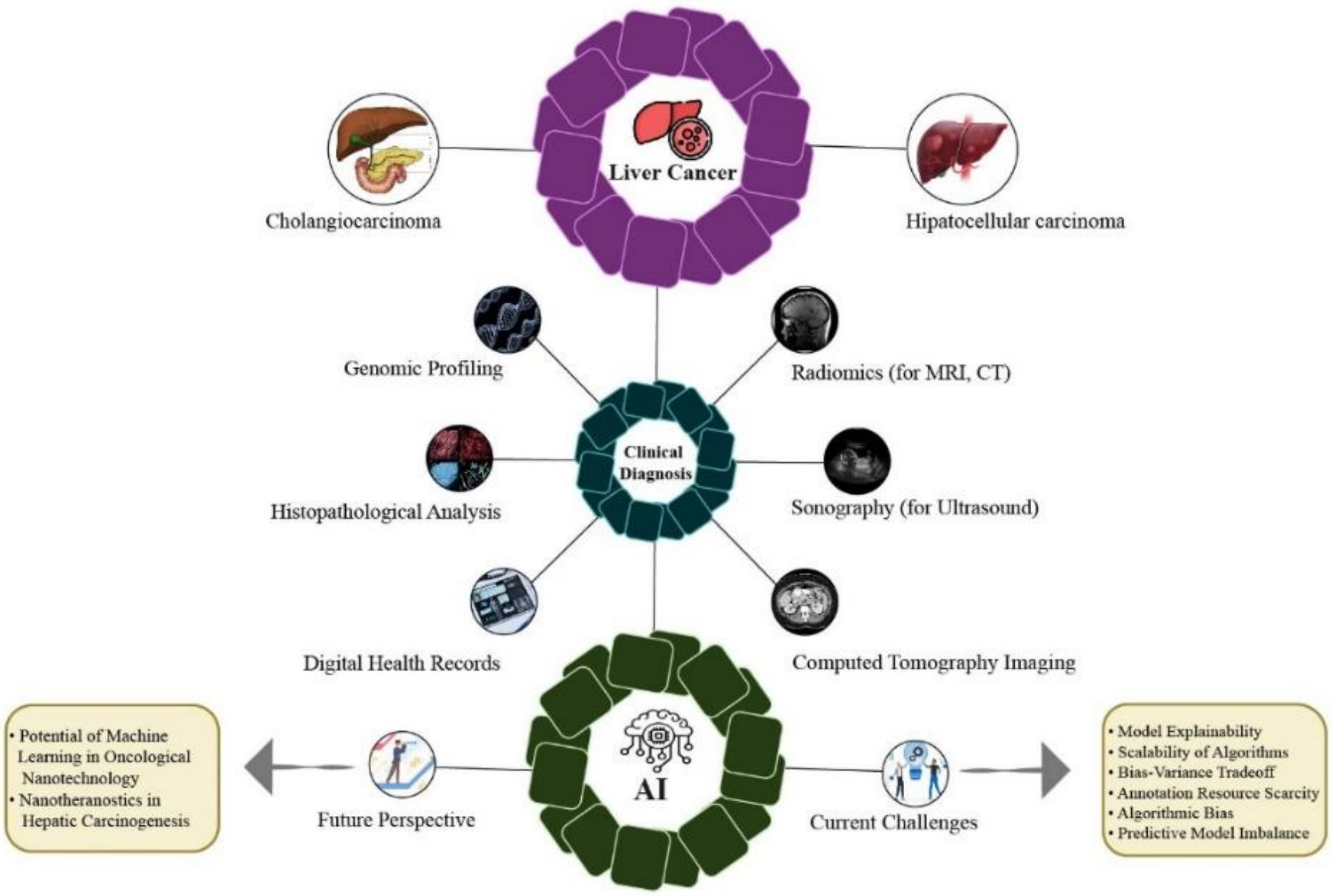

**Figure 8:** Multifaceted AI-Driven Approach in the Clinical Diagnosis of Liver Cancer.

### 5.3.1 Advancements in AI for Liver Cancer

AI has the potential to significantly enhance future healthcare, particularly in the realm of diagnosing and treating liver illnesses like HCC.

Due to technological innovations, clinicians can employ AI algorithms that not only forecast the likelihood of colon cancer spreading to the liver but also assess the effectiveness of treatment and identify any deviations. This intelligent assistant technology is revolutionizing the process of selecting the most appropriate treatment, ensuring that each patient receives optimal care. Table 9 displays many applications of artificial AI in the field of cancer management.

**Table 9:** Applications of AI in Liver Cancer Management and Diagnosis**.**

| AI Application | Function | Ref. |
|---|---|---|
| Predicting Liver Metastasis in Colorectal Cancer | Uses ML to predict the risk of liver metastasis in T1 colorectal cancer patients. | [250] |
| Evaluating Treatment Response | Applies AI for volumetric evaluation and assessment of HCC treatment response. | [251] |
| Precision Diagnosis and Treatment | Contributes to liver cancer management in six application scenarios including risk and treatment response prediction. | [252] |
| Hepatitis Evaluation | Predicts the incidence and progression of hepatitis, which is a key risk factor for HCC. | [253] |
| AI in Cancer Prognosis and Treatment Selection | Predicts cancer prognosis and assists in selecting appropriate treatments for liver cancer. | [254] |
| Risk Prediction Model for Liver Cancer | Develops and validates a risk prediction model for liver cancer based on available risk factors. | [255] |
| Diagnosis and Management of Hepatocellular Carcinoma | Aids in diagnosing and managing liver diseases, especially HCC, through ML. | [256] |
| Increased Cancer Risk in Autoimmune Hepatitis | Examines cancer risks in autoimmune hepatitis patients, focusing on the risk of HCC. | [257] |

AI-integrated medical imaging represents a major advancement in the diagnosis of liver cancer. AI algorithms utilize DL techniques to analyze pictures obtained from CT scans, MRIs, and ultrasounds.

They demonstrate a superior ability to identify biomarkers of liver cancer compared to even the most skilled radiologists. This capability for detection expedites and enhances the diagnostic process. Zhou et al. [258]. The use of AI has revolutionized medical imaging detection, especially when it comes to DL algorithms [259]. The application of AI in ultrasonography has demonstrated encouraging outcomes in the assessment of liver disorders. Studies have shown that AI is exceptionally precise in distinguishing between benign and malignant liver lesions, as well as in assessing the severity of liver fibrosis. The implementation of this technology holds the capacity to significantly enhance ultrasound diagnoses, a crucial factor in the development of effective and personalized treatment strategies for liver cancer. AI) is increasingly utilized in hepatic ultrasonography to differentiate severe liver lesions, evaluate diffuse liver conditions, and conduct screenings for liver illnesses.

This enhances the precision of diagnosis and contributes to the delivery of improved and efficient patient care [259]. Development of an AI system for the identification of HCC. A preliminary single-centre study has successfully demonstrated the construction of an AI system capable of diagnosing hepatocellular cancer using CT imaging data using CT to Ultrasonography Fusion Imaging. The newly developed system utilizes a fully convolutional network and can be trained by both specialized radiologists and general radiologists. It demonstrates accurate diagnostic performance that is equivalent to prior studies. Ultimately, we can expect the implementation of AI systems such as this in a healthcare environment [260].

AI is now making significant progress in the field of liver cancer pathology, particularly in automating the processing of tissue sections to identify abnormal biomarkers and pathways linked to cancer. These technologies enhance the effectiveness of liver biopsies in diagnosing various illnesses and improve the productivity of pathologists in examining thousands of slides each patient. Table 10 presents a comprehensive overview of the distinct tasks that AI carries out in the fight against liver cancer, with each task being more specialized than the one before it. The tasks encompassed are risk assessment, diagnosing liver tumors with a degree of proficiency similar to that of human experts, interpreting complex histopathology images, and providing valuable information about the liver microenvironment.

**Table 10:** Advancements in AI for the Detection and Stratification of Liver Cancers

| AI Application | Function | Ref. |
|---|---|---|
| AI's Role in Gastrointestinal and Hepatobiliary Cancers | Extracts complex information from histopathology images to aid pathologists. | [261] |
| Differentiating Common Liver Carcinomas | Diagnoses common primary liver tumors, comparing AI performance with pathologists. | [262] |
| Impact on Histopathologic Classification | Develops a deep learning assistant for differentiating between HCC and cholangiocarcinoma. | [263] |
| Increased Cancer Risk in Autoimmune Hepatitis | Examines cancer risks in autoimmune hepatitis for AI-assisted risk stratification. | [257] |
| Exosomal miRNAs in Liver Injury | Automates detection of liver cancer in whole-slide images for pathology advancements. | [264] |
| Bibliometric Analysis on AI in Liver Cancer | Investigates the role of exosomal miRNAs in liver injury and implications in liver cancer. | [265] |
| Immunological and Metabolic Landscape in Liver Cancer | Provides a bibliometric analysis on AI research in liver cancer for guiding future applications. | [266] |

Research in the field of oncology has demonstrated a growing demand for efficient biomarkers in the context of liver cancer. Table 11 provides a concise overview of the latest advancements in Biomarker detection. Identifying biomarkers for liver cancer is essential for enhancing early identification in those at high risk, achieving effective targeted therapies, and advancing personalized medicine. AI technologies are transforming the process of identifying and applying biomarkers in liver cancer research [224].

**Table 11:** Innovative Methods for Liver Cancer Biomarker Detection.

| Method Name | Function | Ref. |
|---|---|---|
| ML for Cancer Stem Cell Biomarker Detection | Identifies a set of genes as potential biomarkers for liver cancer stem cells using ML. | [267] |
| Multiplex Detection of MicroRNA and Protein Biomarkers | Developed a mass spectrometric method for the simultaneous detection of nucleic acid and protein biomarkers. | [268] |
| Multi-Omics Strategy for Biomarker Detection | Reviewed the integration of multi-omics technology in biomarker screening for hepatocellular carcinoma. | [269] |
| Quantum Dot Nanoprobes for Biomarker Detection | Suggests employing serological biomarker panels for liver cancer detection using fluorescent quantum dot nanoprobes. | [270] |
| High-Throughput Proteomics | Highlights the synergy between high-throughput proteomic technology and AI in developing next-generation biomarkers. | [271] |
| Sensitive Electrochemical Immunoassay for Biomarker Detection | Reports on the development of an effective electrochemical immunosensor for liver cancer biomarkers. | [272] |

The studies conducted by Shahini et al. [273] examine the clinical use of blood biomarkers and integration algorithms in hepatocellular carcinoma, a type of primary liver cancer that commonly occurs in individuals with chronic liver disease. These studies demonstrate the potential of AI in improving diagnostic and surveillance tools for this disease.

Several investigations currently concentrate on the methodical utilization of biomarkers in primary liver cancer, as well as the potential of AI to create reliable biomarkers for diagnosis, prognosis, and treatment [274]. AI is transforming the field of liver cancer detection and treatment by providing unmatched levels of accuracy, precision, and speed. Table 12 presents contemporary AI methodologies that illustrate the revolutionary capacity of AI in examining various data platforms, ranging from intricate medical scans to tiny cellular investigations.

**Table 12:** Performance Metrics of AI Applications in Various Liver Disease Diagnostics.

| AI Application | Data Type | Efficacy | Ref. |
|---|---|---|---|
| Various AI models | B-mode ultrasonography, medical imaging | Exceeds the diagnostic acumen of human experts | [275] |
| AI in medical imaging diagnosis | Medical imaging, virtual assistants | Enhances risk assessments and treatment response predictions | [252] |
| Six different AI models | Histopathology | Sets new benchmarks for sensitivity, specificity, and predictive values | [262] |
| Hybrid pre-trained models | CT scans, histology | Showcases exceptional levels of accuracy, precision, and recall | [276] |
| AI and machine ML | Surveillance Epidemiology and End Results database | Significant leap in sensitivity, specificity, and AUC | [250] |
| AI model based on CT data | CT data, clinical factors | Enhanced capability to predict liver metastasis with greater accuracy | [277] |
| Deep learning assistant | Whole-slide images (WSI) | Improves diagnostic accuracy for liver cancer subtypes | [263] |
| AI for segmentation and classification | Medical imaging from MICCAI 2027 LITS database | High Dice similarity coefficients for accurate tumor identification and segmentation | [278] |
| AI in HCC treatment and prognosis | Various, including ANN and radiomics | Enhanced accuracy over traditional analysis for treatment and prognosis | [279] |
| AI for liver tumor segmentation | CT scans with synthetic tumors | Accurate segmentation without manual annotation, reducing time and resources for training | [280] |

The integration of AI in the management of HCC represents a significant change in how liver cancer is diagnosed and treated. AI improves the assessment of tissue samples, increases the effectiveness of diagnoses, and offers precise analysis of medical images, resulting in higher levels of specificity, sensitivity, and predictive capability. This leads to treatment regimens and prognoses that are tailored to each individual. The utilization of AI in HCC care represents a paradigm shift in the field of oncology, where technology plays a pivotal role in enhancing patient outcomes and streamlining operations.

## 6. Role of AI in Gastric Cancer

Cancer is a disorder that shows a correlation with age, with the typical age of cancer diagnosis being about 70 years in developed countries.

By 2030, the proportion of cancer diagnoses in individuals aged 65 and above is projected to exceed 70% [281]. The increasing geriatric population will therefore significantly augment the global burden of cancer. In 2020, there were 1.089 million new instances of gastric cancer (GC) diagnosed worldwide, making it the sixth most often diagnosed cancer. It also ranked as the third greatest cause of cancer-related deaths, with 0.769 million fatalities [282]. The median age at the time of diagnosis is 70 years. Glycogen storage disease (GC) impacts several physiological systems within the body. Helicobacter pylori infection, older age, smoking, high salt intake, and inadequate consumption of fruits and vegetables are risk factors that increase the probability of developing gastric cancer. Men have a significantly higher incidence of GC compared to women, with a rate that is more than twice as high. The rates of incidence and mortality vary significantly worldwide, with the highest rates observed in East Asia, Eastern Europe, Western Asia, and South America. [283].

Siegel et al. [142], report that although there is a general decrease in cancer mortality rates, including stomach cancer, the advancement in this area may face obstacles due to the rising incidence of specific types of cancer. This underscores the significance of progress in medical interventions and emphasizes the necessity for ongoing endeavour in timely identification and mitigation.

## 6.1 Traditional Detection Methods for the detection of Gastric Cancer

### 6.1.1 Second-generation narrow band imaging (2G-NBI)

Although the prognosis for gastric cancer is generally not good, if the disease is identified early, there is a significant increase in the probability of survival for five years (99.3% for mucosal cancer and 97.2% for submucosal cancer) [284]. Therefore, early detection should be considered as one of the optimal approaches to improve survival rates of stomach cancer. Nevertheless, there are currently no recognized and effective screening methods available for the early detection of stomach cancer, even in regions with a high incidence of the disease such as Asia, Russia, and South America. Currently, the standard method for identifying stomach cancer is endoscopy using white light imaging (WLI) [285]. The sensitivity of white light imaging (WLI) in detecting early gastric cancer (EGC) is inadequate [285]. Narrow band imaging (NBI) endoscopy is an advanced optical technique that enhances the visualisation of surface features and micro vasculature more effectively than conventional WLI. For example, first-generation NBI (1G-NBI) increased the detection rate of superficial head and neck and esophageal cancers over WLI [286].

### 6.1.2 Serum Screening System

The role of Helicobacter pylori (HP) infection in the onset and advancement of GC is extensively acknowledged. Correa reported the progression of GC from HP-induced atrophic gastritis to intestinal metaplasia, which is a recognized precursor of intestinal type GC [287] and subsequently by Uemura et.al [288]. The significant function of HP in GC was recently elucidated, highlighting its crucial involvement in the early diagnosis of GC by detecting HP-induced atrophic gastritis.

Serum pepsinogen (PG) levels have been demonstrated to accurately indicate the inflammatory condition of the gastric mucosa, namely atrophic gastritis. Recently, the diagnostic capability of PG to identify persons who are at risk for GC has been established [289, 290].

Subsequently, Miki et al. [291] employed these discoveries to create the ABC-classification system, a serum screening method that categorizes the likelihood of gastric cancer, referred to as the ABC classification system. This is achieved by assessing serum anti-HP antibody titers and PG levels.

The technique proved to be valuable in assessing the individual susceptibility to stomach cancer [292] Furthermore, it is anticipated that this technique will be valuable for widespread screening in order to identify cases of stomach cancer.

### 6.1.3 Liquid Biopsy Signature

In the field of oncology, liquid biopsy is becoming a game-changing instrument. Liquid biopsy offers a non-intrusive method for identifying many types of malignancies and assessing a patient's response to treatment. Early identification and dynamic monitoring of malignancies are crucial for improving patient outcomes. Research has mostly concentrated on three specific biomarkers: extracellular vehicles (EVs), circulating tumor cells (CTCs), and circulating tumor DNA (ctDNA).

The primary application of CTCs and ctDNA as biomarkers has been limited to tracking the advancement of disease because they are extremely uncommon and difficult to detect technically [293]. On the other hand, EVs have greater potential due to their stability and their release by cells throughout the body, which allows them to serve as a valuable source of information on the significant biological changes happening inside the tumor microenvironment [294, 295].

Recent studies emphasize the clinical importance of long noncoding RNAs (LRs) found in circulating extracellular vehicles (EVs). These researchers have identified multiple LR candidates originating from EVs that could serve as diagnostic biomarkers for cancer [296, 297]. These LRs provide a separate and distinctive biological perspective that is different from the information provided from tissue and circulating cell profiles. This highlights their potential to improve our understanding of cancer biology [298]. The carcinogenic function of lncRNA GClnc1 has been well-documented due to its capacity to enhance tumor progression. Moreover, the stability of LR expression in different body fluids makes it a prospective candidate for liquid biopsy biomarkers in the diagnosis of cancer [299].

### 6.1.4 Fecal Screening

H. pylori (Hp) infection is acknowledged as a significant risk factor for the development of stomach cancer according to Correa's model [300]. Additionally, sophisticated treatments for Hp eradication have been found to potentially have a partial protective impact in relation to GC [301]. Nevertheless, the likelihood of infected persons developing GCa is minimal, ranging from 2% to 3% [302]. This implies that additional elements, such as unidentified organisms [302] are necessary for the development of GCa.

Therefore, it is necessary to identify specific bacteria linked with GCa in order to create monitoring techniques that are based on risk categorization [303]. While multiple studies have identified different compositions of gastric microbiota in patients with gastric cancer (GCa) compared to those with chronic gastritis (ChG) [304], our understanding of the involvement of non-Helicobacter pylori (Hp) bacteria in GCa remains restricted. Several recent research, including our own, have found an increased presence of Streptococcus anginosus (Sa) in the tumor tissues of individuals with gastric cancer (GCa) [305, 306]**.**

## 6.2 Limitations of Current Techniques for the detection of Gastric Cancer

Table 13 presents a thorough summary of the advantages and disadvantages of different methods used to identify stomach cancer. The visualization of possible stomach problems is now more evident with the improved narrow band imaging (NBI), however its effectiveness in comparison to the conventional white light approach is still unknown.

The blood test system categorizes individuals into different risk groups based on particular markers, although it still produces some incorrect negative results.
Even though liquid biopsy is less intrusive, it is not always successful in identifying disease indicators such circulating tumor DNA.

Ultimately, the identification of certain pathogenic bacteria in stool samples seems to be a more favorable option compared to endoscopy or biopsy. However, its practical usefulness is still being assessed. Each of these methods offers useful insights into the benefits and limitations of various diagnostic procedures for early identification of stomach cancer.

**Table 13:** Emerging Diagnostic Technologies for Gastric Cancer: Functions and Limitation.

| **Diagnostic Type** | **Function** | **Limitations** | **Ref.** |
|---|---|---|---|
| Second-generation narrow band imaging (2G-NBI) | Enhances visualization of surface structures and blood vessels for better detection of EGC compared to WLI. | Limited by the sensitivity of WLI and requires further studies to confirm its efficacy in detecting EGC. | [284] |
| Serum Screening System | Uses serum anti-HP antibody titers and PG levels for a stratified risk assessment of gastric cancer. | Some gastric cancer patients fall into the HP (-)/PG (-) category, indicating a need for further validation. | [287] |
| Liquid Biopsy Signature | Utilizes circulating biomarkers like EVs for non-invasive diagnosis and real-time surveillance of malignancies. | Challenges include the rarity and unreliable detection of primary targets like CTCs and ctDNA. | [307] |
| Fecal Screening | Identifies specific bacteria (e.g., Streptococcus anginosus) in fecal samples for non-invasive GCa screening. | Specificity and sensitivity in differentiating GCa from other gastrointestinal diseases need to be fully determined. | [305] |

### 6.3 Role of Ai in the Detection of Gastric Cancer

AI is introducing a new era in the diagnosis of stomach cancer, facilitating the investigation and creation of inventive approaches for early identification and treatment. Recent research has shown that AI has greatly enhanced the precision of early gastric cancer (EGC) diagnosis when compared to traditional endoscopic pictures, pathology evaluations, and other diagnostic methods. DL algorithms, known for their effectiveness in diagnosing EGC, have demonstrated exceptional accuracy in detecting the condition, even when its visual aspects are difficult for human experts to perceive.

Timely detection is particularly vital during the initial phases of diagnosing stomach cancer, as it can greatly impact the course of the disease and enhance patient prognosis.

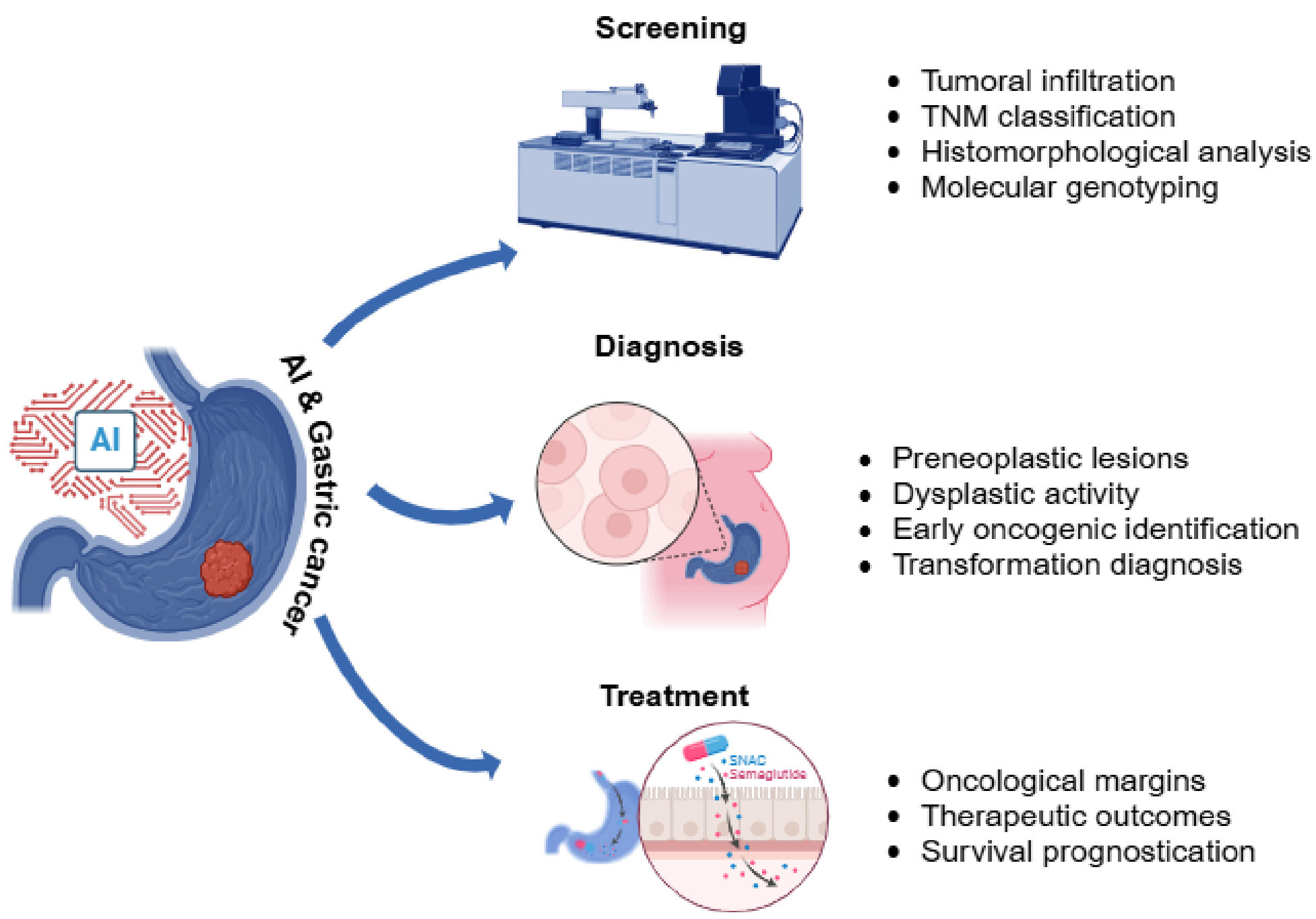

**Figure 9:** This image illustrates the role of AI in gastric cancer management, highlighting its applications in diagnostic imaging, personalized treatment, and targeted drug delivery. AI aids in analyzing tumor characteristics, optimizing treatment plans, and enhancing medication effectiveness for improved patient outcomes

The application of AI in stomach cancer imaging goes beyond just cancer diagnosis, encompassing a wide range of diagnostic purposes. AI has the capability to not only detect lymph node metastases, but also to simulate medication responses and provide prognostic forecasts. The implementation of an all-encompassing strategy for early identification and categorization is essential for the precise management of stomach cancer in the field of precision medicine, facilitating exceptionally precise disease control [308].

**Table 14:** Therapeutic use of deep learning technology trained on CT data.

| Lesion Type | Diagnostic Method | Dataset Capacity | AI Method | Efficacy | Ref. |
|---|---|---|---|---|---|
| Lymph node metastasis | CT | Patient: 730 | DL (Multivariable linear regression analysis) | C-index Outperformed External validation: 0.797 International validation: 0.822 | [308] |
| Lymph node metastasis | CT | Patient: 204 | DCNN | AUC: 0.82 Outperformed | [309] |
| Lymph node metastasis | CT | Patient: 1,699 | ResNet-18 | Extranational validation median AUC: 0.876 Accuracy: nearly 90% Sensitivity: 0.743 Specificity: 0.936 | [310] |
| Risk prediction of overall survival | CT | Patient: 640 | ResNet | C-index Outperformed than DL (internal) vs Clinical(external) vs clinical radiomics Radiomics | [311] |
| Predict prognosis | CT | Patient: 1,615 | S-Net | Extranational validation Useful than TNM C-index staging system | [312] |

Table 14 provides a concise overview of AI research conducted in the field of medical diagnostics, specifically focusing on the analysis of CT images to forecast the spread of cancer to lymph nodes and the overall survival rate. Several models, such as DL, DCNN, ResNet-18, and S-Net, routinely perform better than older approaches, offering superior prognostic evaluation in comparison to the TNM staging system.

Since determining the extent of invasion is crucial for determining the appropriate treatment, it is important to present this assessment to the patient during the diagnostic phase [313]. Nowadays, EUS is a commonly used technique for predicting the depth of invasion. However, it has drawbacks such as being time-consuming, expensive, reliant on the operator's skills, and occasionally resulting in a greater percentage of misdiagnosis [314]. Comparing with experienced endoscopists, a retrospective study by Zhu et al. [315] demonstrated a notable level of precision, sensitivity, and specificity in the diagnosis of invasion depth by CAD [316]. In a referral centre, a study was conducted by the same group using AI (ENDOANGEL). The results showed that AI had lower but still excellent accuracy, sensitivity, and specificity compared to previous image-based studies: 84.6%, 86.7%, 82.3% versus 84.6% to 94.8%, 78.3% to 85.3%, 80.8 to 90.6% . Additionally, the use of AI significantly reduced the time required for diagnosis [317].

**Table 15:** AI-Based Diagnostic Technologies for Early Gastric Cancer Detection: Efficacy and Outcomes.

| Lesion | Diagnostic Method | Dataset Capacity | AI Technology | Summarized Outcomes | Ref. |
|---|---|---|---|---|---|
| EGC | BLI-ME | Patients: 95 | SVM | Validated with SVM, notable cancer detection rate | [318] |
| Classification of lesion | WLE | Images: 5,017; Patients: 1,269; Images: 200; Patients: 200 | Inception-ResNet-v2 | High AUC values in both retrospective and prospective validations, good average accuracy | [319] |
| Gastric lesions | NBI | NBI images: 2,000 | ResNet | Excellent AUC, high accuracy, sensitivity, and good specificity | [320] |
| EGC | WLE, NBI | GC: 2,639; Images: 13,584 | SSD | Validated with CHN and expert comparisons, mixed sensitivity and high specificity | [321] |
| EGC | WLE | Neoplastic: 150; non-neoplastic: 165 | Tango | Validated with Tango/Expert, reasonable accuracy and sensitivity | [322] |

Tables 15 and 16 provide a concise overview of how AI is used in the diagnosis of stomach cancer lesions. SVM and Inception-ResNet-v2 exhibited higher AUC values in both prospective and retrospective trials, although ResNet for NBI pictures yielded the most favorable metrics in terms of accuracy, sensitivity, and specificity.

**Table 16:** Transformative AI Applications in Gastrointestinal Diagnostic Methods.

| Traditional Diagnostic Techniques | Integration of AI | Improvements | Ref. |
|---|---|---|---|
| Endoscopic ultrasound (EUS) | AI systems for depth prediction using endoscopic images | AI achieved similar accuracy in T staging and assessing serosal involvement as EUS | [313] |
| White-light imaging (WLI), No magnifying narrow-band imaging (NBI), Indigo-carmine dye contrast imaging | Development of AI systems for predicting the invasion depth of gastric cancer using WLI, NBI, and Indigo | High lesion-based accuracy and no significant difference between the WLI, NBI, and Indigo AI systems | [323] |
| Magnifying endoscopy with narrow-band imaging (ME-NBI) | AI-assisted convolutional neural network (CNN) computer-aided diagnosis (CAD) system based on ME-NBI images | Improved diagnostic accuracy of the AI-assisted CNN-CAD system for EGC diagnosis | [324] |
| Conventional histopathology | AI-driven histopathological diagnosis system for gastric cancer detection using deep learning | Near 100% sensitivity and an average specificity of 80.6% on a real-world test dataset | [325] |
| Fluorescence hyperspectral imaging | DL combined with spectral-spatial classification method for early diagnosis of gastric cancer | Overall accuracy of 96.5% for differentiating no precancerous, precancerous, and gastric cancer groups | [326] |

The implementation of AI in the field of GC screening and personalized treatment provides a greater likelihood of achieving positive results. Although these technologies hold potential in the management of gastric cancer, additional optimization is necessary to improve their efficacy in fighting the illness.

## 7. Esophageal squamous cell carcinoma (ESCC) and AI

One of the most prevalent forms of cancer that affects the digestive system is esophageal cancer (EC), which develops from the epithelial cells that line the mucosa of the esophagus. EC shows notable variations in occurrence and distribution throughout different regions, with East Asia having the highest incidence rates. In this region, the incidence rates are approximately double the global average of 12.2 cases per 100,000 people [327]. ESCC is the predominant histological subtype of EC in China, representing over 90 percent of all EC cases [328]. Conversely, adenocarcinoma is the most common subtype in regions such as the United States and Europe with lower incidence rates [329]. In 2015, China, which plays a significant role in the global burden of esophageal cancer, reported 246,000 newly diagnosed cases and 188,000 deaths, as stated in a 2019 report by the National Cancer Centre.

These statistics rank sixth and fourth, respectively, in terms of their prevalence among all types of malignancies worldwide [330, 331]. However, at the time of diagnosis, 70% of EC patients are deemed unsuitable for surgical intervention [330]. Early-stage ESCC and precancerous diseases can often be effectively treated with endoscopic methods, which have a 5-year survival rate of up to 90% [332, 333]. The early identification rate of EC is still far too low [330]. The majority of EC patients are diagnosed in an advanced stage, underscoring the significance of early detection [334, 335].

### 7.1 Conventional Diagnostic Approaches and Limitations in ESCC

The issue of rising demands and diagnostics issues may be addressed by AI-enhanced diagnostics [336, 337]. The limitations of the widely employed current diagnostic method for ESCC are presented in Table 17.

**Table 17:** ESCC Detection Strategies: Advancements and Challenges.

| Diagnostic Technique | Function | Limitation | Ref. |
|---|---|---|---|
| Upper Gastrointestinal Endoscopy | The gold standard for diagnosis of ESCC. | Limited by macroscopic judgment and expertise. | [338] |
| White Light Imaging (WLI) Endoscopy | The routine examination method aids in biopsy. | This may lead to missed diagnoses of precancerous lesions. | [339] |
| Lugol Chromo endoscopy (LCE) | Enhances detection of early cancer. | Some patients experience discomfort after iodine staining. | [338] |
| Narrow-Band Imaging (NBI) Endoscopy | Improves sensitivity in early EC diagnosis. | Specificity may need improvement. | [340, 341] |
| Magnifying Endoscopy | Defines lesion range, and aids observation. | Efficient but may require additional expertise. | [341] |
| Electronic Staining Imaging | Enhances mucosal surface display. | No adverse reactions but may not be widely available. | [341] |
| AI-Assisted Endoscopic Diagnosis | Addresses high workload and low efficiency. | Emerging technology, limited clinical application. | [336, 337, 342] |

## 7.2 AI-Based Solutions Transforming ESCC Diagnosis

AI is becoming a crucial tool for the early detection of ESCC, enhancing the accuracy of diagnoses by analyzing both photographs and video Data.

Image-based data can be influenced by bias when selecting the proper sections of imaging for research. On the contrary, AI diagnostics that utilize video enable immediate assessment, making them more applicable in clinical environments and offering vital insights into the system's performance under realistic conditions. The figure labelled as Figure 10 illustrates the mechanism by which the Ai model functions to delete ESCC.

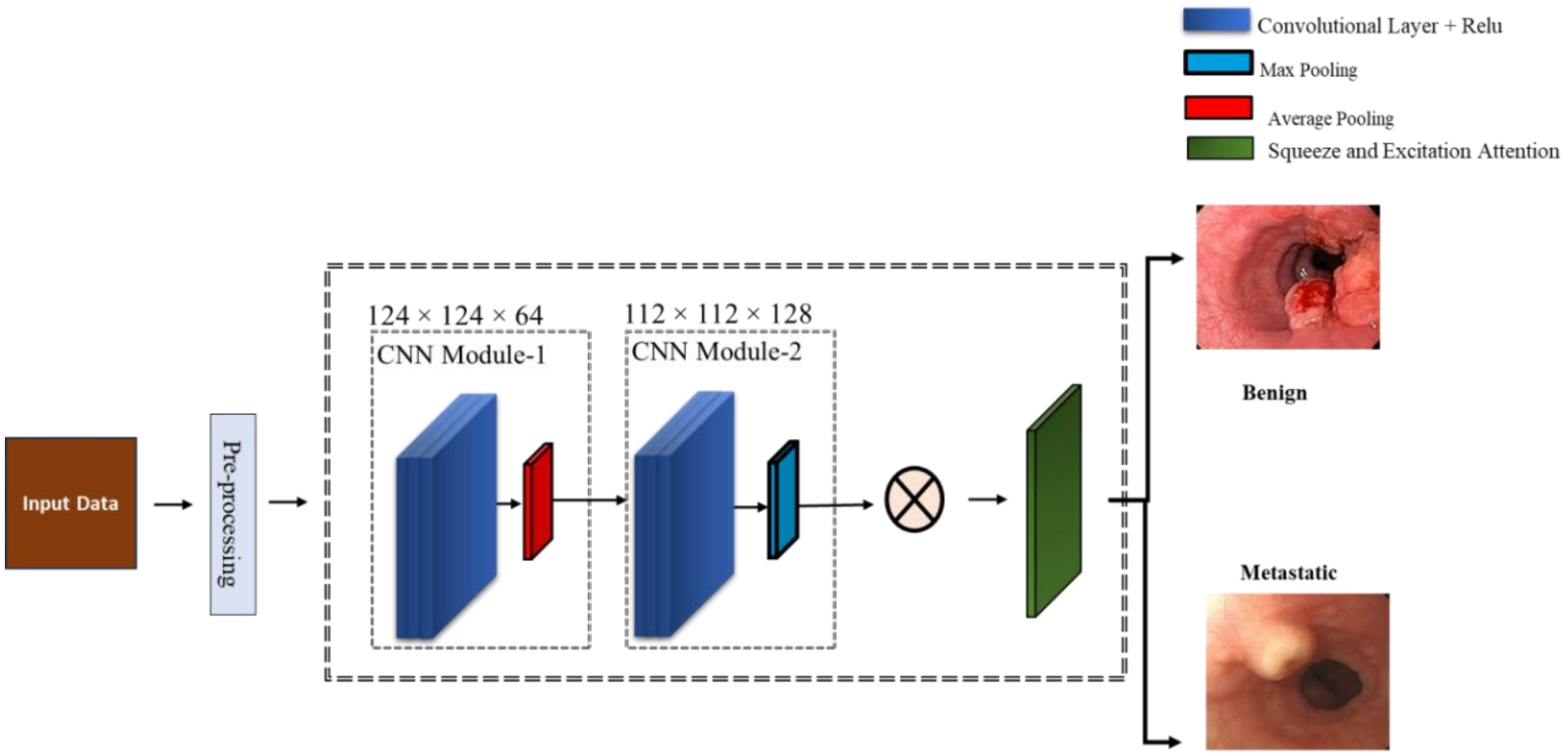

**Figure 10:** The diagram shows how to use CNNs, to identify ESCC from endoscopic images. In order to get set up for examination, the incoming data first undergoes preprocessing, which includes normalization. After that, the data is processed through two different CNN modules, each of whose is intended to improve and refine the extraction of ESCC characteristics over time. The presented images highlight how AI technology can greatly increase diagnosis accuracy in medical imaging by illustrating certain ESCC properties.

A multitude of researchers are currently trying to integrate the latest technology into the diagnosis of ESCC and devise novel techniques to address the limitations, as depicted in Table 18.
Guo et al. developed a CAD system capable of rapidly identifying precancerous lesions and early-stage ESCCs. In order to train the system, a total of 3,703 NBI photos were used from 358 cases that were noncancerous. Additionally, 2,770 Narrow-Band Imaging (NBI) images were included from 191 cases of early ESCC or precancerous lesions.
The test dataset comprised 80 video segments obtained from 33 cases of normal oesophagus and 27 cases of precancerous or early ESCC.

The dataset exhibited a sensitivity of 99.9% per frame and a specificity of 91.5% per frame. The diagnostic sensitivity for non-magnified videos was 60.8%, while for magnified videos it was 96.1% [343]. Fukuda et al. have developed an AI system that can detect ESCC in real-time. The system was trained using a dataset of 23,746 images taken from 1,544 cases of superficial ESCC and 4,587 photos of noncancerous tissues. The technique was validated using Blue Laser Imaging (BLI) and NBI video clips acquired from 144 patients. The findings demonstrated that the AI had higher sensitivity (91% for non-magnified NBI footage) compared to expert endoscopists, but had worse specificity [344].

Tajiri et al. used 25,048 photos from 1,433 superficial ESCC patients to construct an AI system focused on real-time early ESCC diagnosis; these photos were verified against NBI recordings of probable superficial ESCC. The AI system exhibited a specificity of 75.0%, a sensitivity of 85.5%, and an accuracy of 80.9% in diagnosing ESCC, surpassing the diagnostic abilities of non-ESCC specialised endoscopic professionals [345].

Shimamoto et al. developed an AI system that can diagnose the depth of invasion in ESCC in real-time. The system was trained using over 20,000 photos of ESCC that were pathologically confirmed. The system was validated using 102 ESCC video images. The study showed that it had higher accuracy, sensitivity, and specificity compared to trained endoscopists in both non-magnified and magnified videos [346].

Although many studies rely on historical data from a single center, AI-based diagnostic algorithms have demonstrated performance that corresponds to, or even superior to, that of endoscopic professionals. The effectiveness of these AI systems in practical medical settings may be limited by the accuracy and consistency of the training datasets in relation to real-life endoscopic conditions. Yuan et al. presented an AI system capable of identifying and outlining flat abnormalities in real-time during endoscopic surgeries. This system functions as a model for possible use in clinical settings to diagnose ESCC [347].

**Table 18:** AI-Based Model in the Detection of ESCC

| Study Objective | Image Type | Model | Test Set | Reference |
|---|---|---|---|---|
| Differentiation of abnormal IPCL from normal | ME-NBI | CNN | 17 (10 ESCCs), 7046 images | [348] |
| Detection of ESCC | ME-NBI | CNN-SVM | 219 (165 ESCCs), 1383 images | [349] |
| Determining the invasion depth of ESCC | WLI, NME, ME-NBI/BLI, LCE | DCNN | 155 ESCCs, 914 (509 ME) images | [350] |
| Localize and identify ESCC | WLI | DNN | 52, 187 images | [351] |
| Real-time diagnosis of ESCCs | NBI | CNN-SegNet | 60 (27 early ESCCs), 80 videos | [343] |
| Detect and differentiate ESCC | WLI, NME, ME-NBI/BLI, LCE | CNN-SSMD | 135 (52 ESCCs), 727 (204 ME) images | [352] |
| Measure ESCC invasion depth | WLI, NBI | CNN-Google Net | 291, 291 images | [353] |
| Diagnosing ESCC with videos | NBI or BLI | CNN-SSMD | 144, 144 videos | [344] |
| Calculate cancer invasion depth | WLI, NME, ME-NBI/BLI, LCE | CNN-PyTorch | 102, 102 ESCC videos | [354] |
| Recognition of IPCL patterns and predict invasion depth | ME-NBI | CNN-ResNet-18 | 114 (69 dysplastic), 67,740 (39,662 dysplastic) images | [355] |
| Real-time to diagnose ESCC | WLI | DCNN | 162 (58 ESCCs), 700 (207 ESCCs) images | [356] |
| Predict multiple LVLs | WLI, NBI | CNN-Google Net and Caffe | 72 (32 with multiple LVLs), 667 (342 with multiple LVLs) images | [357] |
| Automatic diagnosis of early ESCC | WLI, NME, ME-NBI/BLI, LCE | DCNN - Yolo V3, ResNet V2 | 1055 cases, 2309 images, 104 videos | [358] |
| Classify the micro vessels of ESCCs | ME-NBI | CNN-ResNeXt-101 | 131 lesions, 747 (419 B1, 292 B2, 36 B3) images | [359] |
| Detect ESCC with Videos without focusing on the lesion | WLI, NBI, and BLI | DL-BiSeNet | 100 (50 ESCCs), 100 videos (50 ESCCs) | [360] |
| Detect ESCC from endoscopic videos | WLI and NBI | DCNN-SSMD | 72 patients, 144 Videos | [346] |
| Detect and differentiate histological grade of ESCC | WLI, NBI | CNN-SSD | Unknown, 264 (54 normal) images | [361] |
| Identify ESCC under NBI imaging and compare it with WLI | NBI, WLI | CNN-VGG | 112 cases (42 ESCCs), 316 pairs of images (133 abnormal) | [362] |
| Detecting ESCC | WLI, NME, ME-NBI, LCE | DCNN - YOLO v3 | 119, 2088 (1245 ESCCs) images, 142 videos (76 ESCCs) | [363] |
| Predict IPCLs subtypes of ESCC | ME-NBI | DCNN - HRNet+OCR | 176 patients, 1323 images | [364] |
| Detect and delineate margins of ESCC | WLI | DCNN - YOLACT | 312 (96 external validation), 3506 (890 external validation) images | [365] |
| Detect ESCC in simulated clinical situations | WLI, ME, non-ME, NBI/BLI | BiT-M (ResNet-101×1) | 147 lesions (83 ESCCs), 147 videos (83 ESCC videos) | [345] |
| Diagnosis of ESCC | WLI, NBI, and LCE | DLYOLOv5l | 101, 1462 images | [366] |
| Detect and delineate the extent of ESCC | NBI | DCNN - YOLACT | 414 cases (311 ESCCs), 2517 (1488 ESCCs) images and 140 videos (70 ESCCs) | [367] |

The integration of AI into early ESCC detection shows promise, although facing hurdles. It has the ability to decrease the burden of endoscopists and enhance diagnostic efficiency.

The invasiveness of endoscopic diagnosis, along with the diversity of pathological diagnosis, has generated a growing interest in the field of early gene diagnosis for EC. Xing et al. [368] utilized RNA transcriptome data from ESCC patients to identify distinctive secretary proteins, with the aim of constructing a diagnostic model. While the model showed a significant level of sensitivity, its ability to accurately diagnose early ESCC was considered insufficient.

In addition, Shen et al. [369] examined the application of developing diagnostic models employing microRNAs and protein markers that are particularly present in the plasma of patients with ESCC. Based on their analysis, these models effectively differentiate between esophageal squamous dysplasia (ESD) and healthy controls. However, there has been relatively little attention given to the use of AI-assisted gene identification for identifying dysplasia associated to Barrett's esophagus (BE) and early Esophageal Adenocarcinoma (EAC).

Slaby et al. [370] showed that certain microRNAs in BE tissues can distinguish between BE with dysplasia and BE without dysplasia. The early identification of ESCC poses a challenge due to the absence of initial symptoms and the necessity of a diagnosis from a specialist. AI systems that use advanced algorithms to identify ESCC using image and dynamic video data have demonstrated superior sensitivity and specificity when compared to experienced endoscopists.

The utilizations of real-time AI diagnostics can enhance the early identification of ESCC. Despite the existing constraints, endoscopists can benefit from a decrease in their workload and an improvement in diagnostic efficiency by integrating AI into the diagnosis of ESCC [347]. The main objectives of future endeavours should focus on incorporating AI into healthcare workflows, validating AI models across diverse populations, and enhancing specificity. AI has the ability to drastically revolutionize the prognosis and diagnosis of ESCC, offering hope for improved outcomes in the battle against this deadly cancer. Long-term surveillance and education for healthcare professionals are additional areas where AI can be applied.

## 8. Role of AI in Cervical Cancer

Cervical cancer (CC) is a significant worldwide public health issue, particularly prevalent in numerous low-income and middle-income countries (LMICs) [371]. CC arises from the tissue of the uterine cervix and is characterized by uncontrolled proliferation and death of cervical cells [371]. Cancer ranks fourth in terms of both the number of new cases and the number of deaths it causes [372]. Young ladies have the highest occurrence rate [373]. In 2018, there were around 570,000 reported instances of cervical cancer, resulting in 311,000 fatalities. These figures represent 6.6% of all newly diagnosed cases and 7.5% of all female cancer-related deaths [374]. The estimated greatest prevalence of CC was found in Eswatini, while the average age of diagnosis and death varies greatly worldwide, demonstrating the disease's extensive impact. Despite lower incidence and death rates in some areas as a result of widespread prevention efforts, it is anticipated that the global incidence of CC will increase by about 50% by 2030 [375].

### 8.1 Causes, Prevention, and Diagnosis

The primary histological classifications of CC consist of squamous cell carcinoma (SCC) and adenocarcinoma (ADC), which account for the majority of CC cases [376]. The prognosis of various types of individuals varies, with certain studies suggesting a less favourable prognosis for people with ADC [377]. Chronic infections with high-risk human papillomavirus (HPV) are a major contributing factor to the development of CC, responsible for the majority of cases [375]. Persistent infection with carcinogenic HPV variants has been linked to the development of cancer, presenting new opportunities for primary and secondary prevention [378]. Various risk factors, including cigarette smoking, early marriage, engaging in sexual activity with multiple partners, having unprotected intercourse, giving birth multiple times, and using oral contraceptives, have been associated with an elevated risk of CC [379, 380]. Although there are difficulties, CC may mostly be prevented by implementing efficient measures such as HPV vaccination and screening, especially with methods that are based on HPV.

Nevertheless, the global efforts to decrease the occurrence and death rates of CC have not been consistent, particularly in low- and middle-income countries (LMICs) where progress has been slower and, in certain instances, there have been rises in the rates of occurrence or death during the previous decade [381].

### 8.2 Traditional Methods for Detection Cervical Cancer

The methods for cervical screening consist of the HPV DNA test, the liquid-based Pap smear test, the traditional Pap smear test, and visual inspection with acetic acid (VIA) or Lugol's iodine (VILI) [382].

#### 8.2.1 Papanicolaou (PAP) smear Screening Test

The Papanicolaou (PAP) smear is the most prevalent method used for cervical cancer screening. In order to identify any complications, the procedure entails extracting cells from the cervix and examining them using a microscope.

In addition to detecting cervical cancer, the PAP smear test can also detect precancerous lesions, which can be treated before they progress into cancer. Research has shown that the PAP smear test can significantly reduce the likelihood of dying from cervical cancer by 70%.In underdeveloped nations, the average percentage of cervical cancer screening coverage is 19%, while in industrialised countries, it is 63% [383].

### 8.2.2 HPV testing

The World Health Organisation (WHO) recommends HPV testing as the initial screening method, followed by the treatment of precancerous lesions or referral of patients for further evaluation and therapy for invasive malignancy.

A Pap test and an HPV test are similar in terms of their diagnostic capabilities. When a Pap test detects abnormal changes in the cervix, the practitioner may opt to conduct an HPV test in conjunction with the Pap test using a sample of cervix cells.

The WHO has officially supported a worldwide plan to expedite the elimination of cervical cancer by the year 2030. The objectives are to immunise 90% of females against HPV, conduct high-performance screening on 70% of women, and provide treatment to 90% of women with cervical disease [384].

### 8.2.3 Colposcopy Method

Colposcopy has gained widespread usage globally as an essential diagnostic tool and for the management of cervical cancer [385]. Colposcopy, in addition to its emphasis on cervical imaging, also includes the application of chemical staining to cervical epithelial cells. The quality of the solution, the technique used during the procedure, and the techniques of observation can all impact the staining process. Consequently, there are substantial discrepancies in the agreement among and within operators when it comes to colposcopy impressions and pathology. The results range from 52% to 66% [386].

### 8.2.4 Liquid based cytology technique

Liquid-based cytology (LBC) is a modern technique that improves the quality of samples and the interpretation of results, when compared to standard smears [387]. These slides are more straightforward to evaluate compared to standard smears, and the remaining sample can be utilized for extra testing, namely for HPV. Furthermore, it has been recognized that LBC is beneficial in the field of automation. Implementing automation in the slide preparation process reduces the likelihood of human mistakes and ensures consistent quality in the final outcomes.

Moreover, LBC can be employed in conjunction with automated image processing systems, hence facilitating screening and initial diagnosis [388]. By reducing the number of inadequate samples, LBC can enhance the sensitivity of endometrial cancer detection [389]. In relation to ovarian cancer, the precision of cytologic diagnosis can be enhanced by utilizing Liquid-Based Cytology (LBC) to assess pelvic washings or samples of ascitic fluid [390].

The Food and Drug Administration has approved the initial LBC technologies, namely ThinPrep and SurePath. Regrettably, the utilization of these LBC techniques, which necessitate a costly automated apparatus, is limited in developing countries due to the high expenses associated with testing. LiquiPrep, a more advanced version of LBC technology, reduces the requirement for the majority of devices used in first-generation LBC procedures [391].

### 8.2.5 Cone Biopsy

Cone biopsy screening is employed to promptly detect abnormal cells in the cervix. The Pap smear, or Papanicolaou test, is the most widely used screening test that involves the collection of a small sample of tissue from the lining of the cervix using a brush. This sample is then examined under a microscope to detect any abnormalities in the cells. This type of test can be conducted to ascertain the presence of HPV in the cervix. The images obtained from the procedure are often known as Pap smear images, and they play a vital role in the early identification and categorization of cervical cancer [392].

## 8.3 Limitations in the current Detection methods of Cervical Cancer

Table 19 provides a concise overview of conventional techniques used for screening cervical cancer. It outlines their uses in identifying the disease at an early stage and managing it, as well as any restrictions they may have. This essay compares commonly employed tests, such as the Pap smear and HPV testing, with more specialist techniques like colposcopy and liquid-based cytology. Each of these methods has its own importance and restrictions when considering global health issues.

**Table 19:** Comparative Overview of Traditional Cervical Cancer Screening Methods and Their Limitations

| Traditional Method | Application | Limitation Summary | Ref. |
|---|---|---|---|
| Papanicolaou (PAP) Smear Test | Used for screening to detect cervical cancer and precancerous lesions; significantly reduces cervical cancer mortality. | Time-consuming and requires expert review; subject to manual errors and interpretation inconsistencies; may miss some lesions. | [383] |
| HPV Testing | WHO's primary screening recommendation; part of a strategy to eliminate cervical cancer by testing for HPV presence or absence in cervical cells. | High costs, requires laboratory processing, and results may take time. | [384] |
| Colposcopy Method | Used for diagnosis and management of cervical cancer through imaging and staining; focuses on detailed cervical examination. | Variability in diagnostic impressions due to operator differences; accuracy depends on technique and observation quality. | [386] |
| Liquid Based Cytology (LBC) | Enhances sample quality and accuracy over conventional smears; suitable for automation and compatible with image analysis systems for better diagnosis. | Higher procedure costs than conventional cytology; although it reduces unsatisfactory samples, it doesn't eliminate them completely. | [387] |
| Cone Biopsy | Utilized for early detection when abnormal cells are suspected, through tissue removal for microscopic examination; aids in early cervical cancer detection. | Specific limitations are not detailed, but the procedure is typically reserved for cases where abnormal cells are suspected and requires surgical intervention. | [392] |

### 8.4 Integration of AI in the prognosis of Cervical Cancer

AI is transforming the field of cervical cancer detection and treatment by improving diagnostic methods like PAP smear analysis, disease segmentation, and non-invasive staging.

Table 20 demonstrates the incorporation of AI into imaging technologies such as CT and MRI, which enhances the precision of diagnosis and has the potential to be utilized for tailored patient care.

**Table 20:** From Traditional Screening to AI: The Changing Landscape of Cervical Cancer Diagnostics

| Advancement | Summary | Ref. |
|---|---|---|
| AI in PAP Smear Diagnosis | AI techniques have been utilized to enhance the accuracy and efficiency of PAP smear diagnoses, offering quantitative, consistent, and reproducible results, and potentially reducing the need for invasive procedures. | [393] |
| AI for Disease Diagnosis | AI has been increasingly applied to various disease diagnoses, improving the segmentation of cells and identification of dysplasia. Its performance in population-based screening is still being evaluated. | [394] |
| CAD Systems | CAD systems automate the detection of cervical cancer, improving screening efficacy and reducing human error and subjectivity. | [395] |
| Cervical Cytology Limitations | Traditional cytology screening is limited in identifying tumor path and scope, leading to inaccurate cervical cancer assessment. | [396] |
| Imaging Technologies | CT and MRI are widely used for their high-resolution imaging capabilities in female malignant tumors. MRI, in particular, is valuable for pre-operative staging of cervical cancer and response evaluation in treatment. | [397] |
| Imaging Limitations | MRI and CT have limitations, such as false-positive findings post-treatment. MRI is generally more accurate for early stages of cervical cancer. Other modalities like TRUS and TVU are used for specific invasion detections. | [398] |
| Machine Learning (ML) | ML improves cancer detection and treatment, offering a new perspective and broader prognosis capabilities. The trend towards personalized and predictive medicine in oncology is reflected in the use of ML. | [399] |
| Deep Learning (DL) | DL technologies improve with larger datasets, recognizing complex features in medical imaging for disease diagnosis. CNN and RNN are notable DL algorithms used in this field. | [400] |
| Treatment Guidelines | The NCCN guidelines recommend surgery or radiotherapy for early-stage cervical cancer and CCRT for advanced stages without metastasis. | [401] |

Figure 11 demonstrates the crucial importance of AI in the management of cervical cancer, emphasizing the substantial influence of DL in the process of diagnosing and treating the disease. This integration facilitates accurate diagnosis, individualized prognostics, and enhanced treatment effectiveness, highlighting the significant role of AI in advancing oncological healthcare.

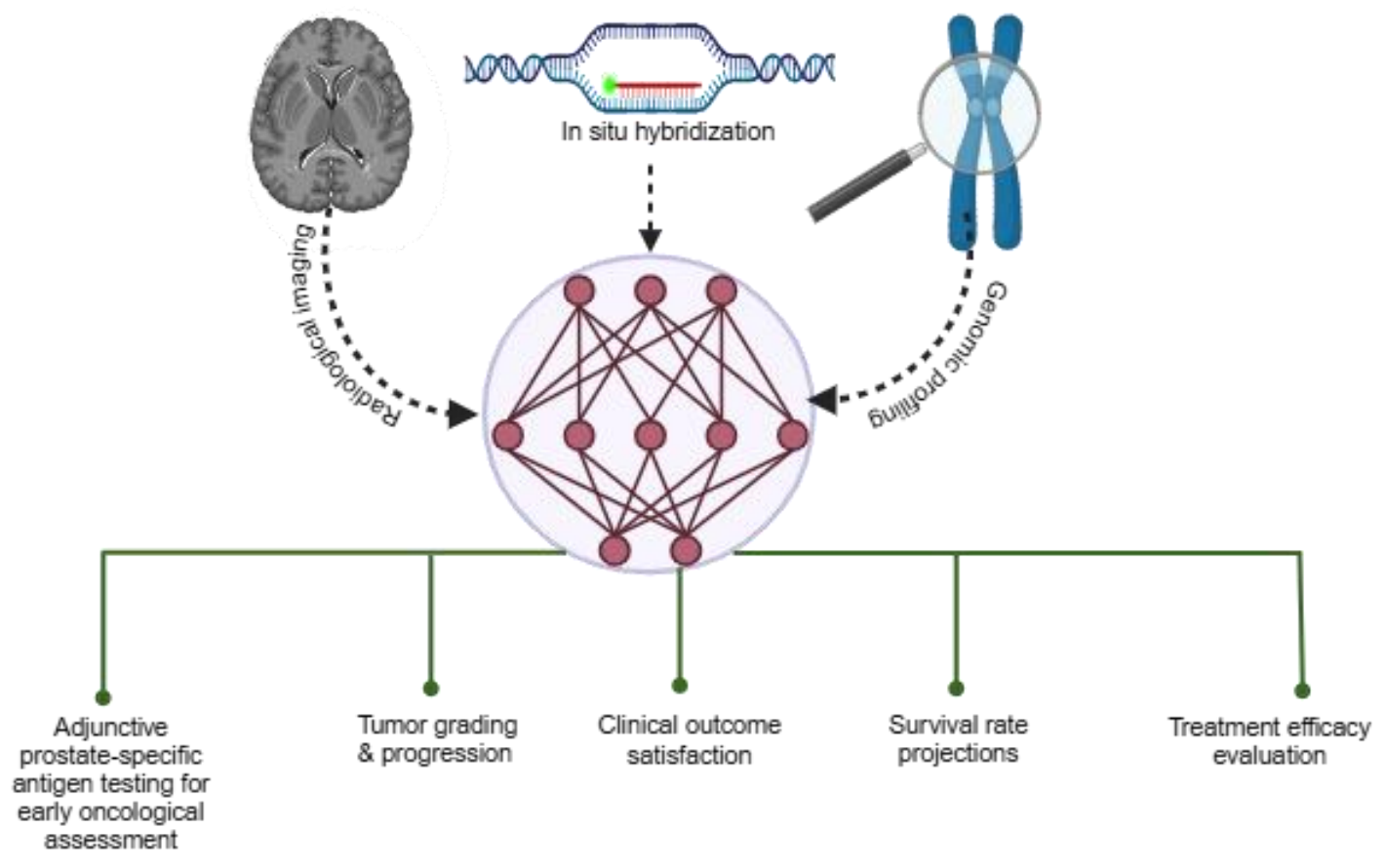

**Figure 11:** Artificial Intelligence Workflow in Prostate Cancer Diagnosis and Management

The image highlights the significant importance of technology, specifically AI, in the treatment of prostate cancer. The initial diagnostic inputs consist of computer analytics, radiological imaging, in situ hybridization, and genomic profiling. These inputs are then processed using either guided or autonomous AI training.

These functions aid in the process of diagnosing and classifying diseases, predicting outcomes, and determining characteristics, ultimately improving clinical results such as assessing the severity of tumors, evaluating the effectiveness of treatments, and predicting survival rates.
Significant progress has been made in the incorporation of AI in the field of cervical cancer. The table below provides a summary of many multidimensional applications of AI in the field of Cervical Cancer.

Table 21: Artificial Intelligence in Cervical Cancer: Detection to Treatment

| **AI Application** | **Application** | **Advantage** | **Ref.** |
|---|---|---|---|
| Detection and Diagnosis | AI algorithms identify HSIL and cervical cancer considering the impact of HPV on protein expression. | Improved diagnostic accuracy with DL techniques. | [372] |
| Treatment Individualization | AI enables understanding of cervical cancer's pathogenesis, allowing for personalized treatments. | Aids in evaluating the effectiveness of surgical and drug treatments. | [371] |
| Screening Improvement | AI models, using techniques like SVM and deep learning, analyze screening images. | High diagnostic accuracy in screening processes. | [373] |
| Clinical Decision Support | ML and DL categorize cervical cytology and colposcopy images for clinicians. | Automates segmentation and aids in clinical decision-making. | [374] |
| Prevention Strategies | AI detects precancerous lesions and assists in treatment decisions. | Enhances early detection and prevention efforts. | [375] |
| Reproducible Screening | AI pipelines diagnose treatable precancerous lesions. | Provides robust and clinically translatable screening tools. | [402] |
| MRI in Treatment Stratification | MRI and AI work together for accurate staging and treatment triaging. | AI interprets MRI results for appropriate treatment selection. | [376] |
| Real-Time Diagnostic Tools | AI/ML models with fluorescence and polarized spectroscopy for diagnostics. | Development of real-time, minimally invasive tools for early detection. | [377] |

Conventional methods for detecting cervical cancer, such as PAP smears and HPV testing, are time-consuming, expensive, and necessitate specialized analysis. AI-enabled techniques have the potential to greatly enhance cervical cancer care by improving detection and diagnosis, ultimately leading to personalized treatment.

These methods involve improving the precision of screening, aiding in clinical decision-making, and creating diagnostic instruments that provide real-time results. By utilizing deep learning, AI has the potential to transform cervical cancer care by analyzing intricate data to enhance diagnosis accuracy, therefore tailoring patient care to individual needs. The combination of AI and its inherent capacity for tailored treatment in the management of cervical cancer presents the rising possibility of predictive, personalized therapy at the point-of-care, skipping the need for laboratory testing. This offers a promising approach to reducing the worldwide impact of this cancer.

## 9. Thyroid Cancer and AI

The thyroid gland, situated in the lower neck, is the largest endocrine gland in an adult and has a distinctive butterfly shape. It controls cellular metabolism by producing hormones and regulating calcium balance in the human body. An effectively functioning thyroid gland is crucial for maintaining the precise hormonal equilibrium required to regulate metabolism at an ideal rate [403]. The connection of this gland to the most common endocrine tumors is making it more and more well-known worldwide [404].

Thyroid tumors are frequently encountered in adults. Thyroid nodules are found in up to 50% of people [405]. Thyroid cancer has shown a continuous increase in occurrence since the beginning of the 20th century, and it currently has the highest rate of occurrence among all types of cancer [406].

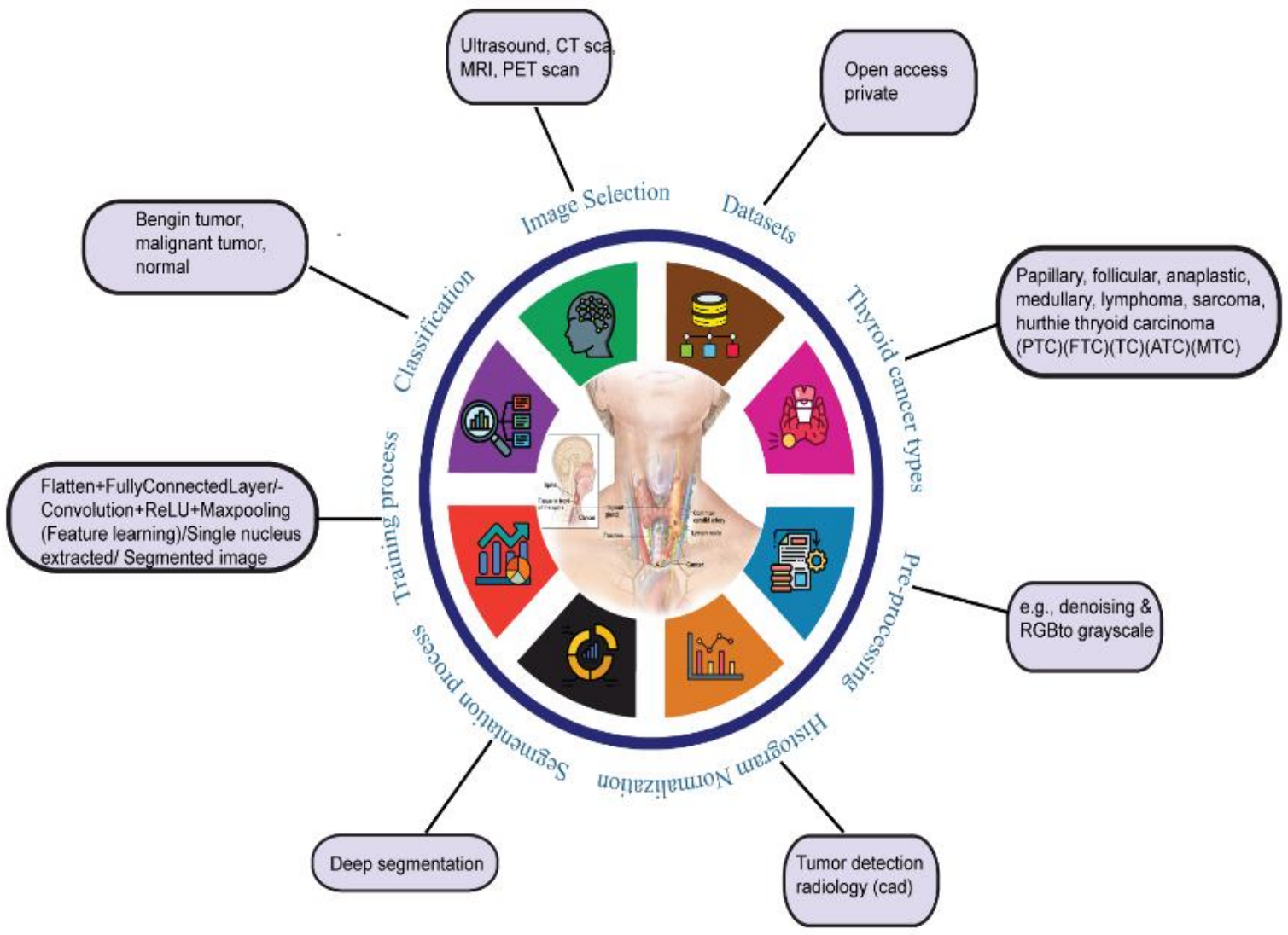

**Figure 12:** The image delivers a detailed and complete structure for the detection and classification of thyroid cancer employing AI, explaining each step from image acquisition to diagnosis. It emphasizes the use of radiological imaging modalities, pre-processing, and deep learning methods for improving the diagnostic accuracy.

Recent research shows that roughly 43,800 new cases of thyroid cancer will be reported in the United States in 2022 [407]. Thyroid cancer ranked as the fourth most commonly diagnosed ailment among females in China [408]. Identifying the cause of thyroid cancer, improving diagnostic methods, and providing personalized treatments to reduce the likelihood of the disease returning are difficult efforts because to the clear impact of greater rates of illness and death associated with this condition.

Therefore, it is crucial to perform a methodical examination of relevant established procedures. Analyses are anticipated to be performed on research that employ statistical and ML methods to investigate the development of thyroid cancer, create personalized treatment recommendation systems, and enhance diagnostic accuracy and efficiency. The workflow of the AL based system utilized for the identification of Thyroid cancer is illustrated in Figure 12.

## 9.1 Thyroid Cancer Pathogenesis

Clinicians encounter difficulties in identifying the etiology of thyroid cancer due to the complex nature of its pathogenesis, which is influenced by various factors including dietary patterns, medical history, genetic predisposition, and recurrent health conditions. It is a common decision made by clinicians to assess one component for its potential link to thyroid cancer development. After taking this first step, a number of possible co-morbidities and risk factors for thyroid cancer were identified: For example, a deficiency in vitamin D[409], exposure to radiation [410], obesity [411], diabetes [412], iodine consumption [413], smoking status [414], and familial history [415], among others. Assessing individual risk factors for thyroid cancer based on previous research is inefficient and fails to account for their interrelationships, leading to misconceptions about the disease's origins. The existing approaches employed to evaluate the complex interaction among several risk factors in thyroid cancer exhibit shortcomings, hence amplifying the knowledge gaps about its etiology. Thus, there is a substantial need for an automated approach that can effectively integrate complex medical data to uncover the fundamental, multifactorial causes of thyroid cancer.

### 9.2 Limitation in the current Prognosis of Thyroid Cancer

Thyroid cancer is a multifaceted and contentious topic within the medical field. Well-established factors, such as exposure to radiation and genetic inheritance, are recognized to play a role in the development of thyroid cancer [416]. Table 22 highlighted the limitation of current diagnostic techniques in Thyroid cancer.

**Table 22:** Current diagnostics techniques and limitations.

| Diagnostic Technique | Use of Technique | Limitations | Ref. |
|---|---|---|---|
| Thyroid Function Examination | Measures hormones produced by the thyroid gland (TSH, T3, T4, FT3, FT4). Ultrasonography suggested for abnormal results. TIRADS scoring used for risk assessment based on ultrasound features. | TIRADS scores alone cannot provide precise decisions for thyroid cancer diagnosis. Further FNAC needed for nodules with TIRADS scores 2-5. | [417, 418] |
| Fine-Needle Aspiration Cytology | Biopsy from thyroid nodules using needles guided by ultrasonography. Biopsy assessed by pathologists for malignancy risk stratification. | Results easily affected by the expertise of pathologists, leading to potential unnecessary surgeries for nodules later determined as non-malignant. | [419, 420] |
| Thyroid Surgery | Required for thyroid cancer patients, includes partial thyroidectomy, total thyroidectomy, | Over-treatments and post-therapy hypothyroidism may occur. Customized treatment plans based on individual factors lacking. | [421, 422] |
| Radiation Therapy (Iodine-131) | Mitigates recurrence risks by performing Iodine-131 treatment post-thyroidectomy. | Adverse effects, such as salivary gland dysfunction, can occur if iodine dosage is not carefully determined. Lack of studies proposing customized treatment recommendation systems. | [423] |

### 9.3 Role of AI Based techniques in Thyroid cancer detection

Artificial neural networks (ANN), which are ML techniques, are commonly used in the field of thyroid cancer to predict mortality, survival, and recurrence [424]. Prediction of survival rates is accomplished by utilizing ML algorithms such as Multilayer Perceptron (MLP), logistic regression, and decision trees. These algorithms consider variables such as age, tumor size, and health status [425]. Among the methods used to assess recurrence rates are ensemble algorithms, random forests, decision trees, and XGBoost [426]. When trying to forecast the likelihood of recurrence, factors such as lymph node ratio are taken into account [427].

The Al data set and Al model utilised for detecting Thyroid cancer are presented in Tables 23 and 24, respectively. The determination of treatment procedures for thyroid cancer is based on risk stratification, which differs according to the subtype of the cancer (such as papillary and anaplastic), with anaplastic being the most severe [428]. Integrating patient-specific factors such as weight, age, and medical history is an essential element of personalized treatment decision support systems [429]. The fuzzy cognitive map technique is supplemented by the utilisation of the Apriori and DeepSurv models in the construction of personalized treatment procedures [430, 431]. The extraction methods employed DL and the performance of different thyroid cancer types are evaluated in Table 25 and 26. The research emphasizes the increasing importance of Deep Neural Networks (DNNs) in detecting thyroid cancer due to their outstanding accuracy.

The research highlights the persistent challenges, chief among which is the lack of clean datasets, even if it also recognizes their crucial significance in real-world applications. The existing AI-based systems possess specific constraints that necessitate attention, as outlined in Table 26.

**Table 23:** Public and Private Thyroid Cancer Database used in thyroid cancer detection.

| TCD | Image Type | Data Type | Instance | Year/Ref |
|---|---|---|---|---|
| BMU | Sonographic | PNG | 1077 | [432] |
| TCCC | US | PNG | 370 | [433] |
| Clinical | US | JPEG | 117 | [434] |
| Hospital | US | JPEG | 62 | [435] |
| TIRADS | US | JPEG | 5278 | [436] |
| Peking Union | US | JPEG | 4309 | [437] |
| Medical Center | US | PNG | 1425 | [438] |
| PubMed | CT scans | JPEG | 2108 | [439] |
| ACR | DICOM | DICOM | 1629 | [440] |

**Table 24:** The advantages and disadvantages of artificial intelligence-based methods for the detection of thyroid cancer.

| AI Method | Acc. (%) | Advantages | Drawbacks | Ref. |
|---|---|---|---|---|
| DAE | 92.9 | No need for labeled data | Insufficient training data, requires relevant data | [441] |
| CNN | 85.0 | High detection rate | Insufficient labeled data, weak interpretability | [442] |
| RNN | 98.2 | No need for labeled data | Slow computation, complex training | [443] |
| MLP | 95.0 | Adaptive learning | Limited results | [444] |
| KNN | 93.0 | High sensitivity | Insufficient labeled data | [445] |
| SVM | 97.0 | High sensitivity | Weak interpretability, long training time | [446] |
| DT | 73.10 | No. of scaling normalization needed | Unstable | [447] |
| LR | - | Low-cost training, easy implementation | Difficulty labeling data | [448] |
| B | 94.88 | High detection rate | Loss of interpretability, high computational cost | [449] |

**Table 25:** Selection and Extraction methods based on DL conducted in the diagnosis of thyroid cancer.

| Classifier | Feature | Feature Engineering | Contributions | Ref. |
|---|---|---|---|---|
| KNN | FC/IG | ANFIS | Reduced data redundancy, handling missing data | [450] |
| SVM | FC/CFS | Kernel-based feature extraction | Geometric and moment feature extraction | [451] |
| CNN | FC/R | Combined ML and feature selection algorithms | Analyzed SEER database using various algorithms | [452] |
| CNN | FE/PCA- | Oversampling and PCA for dimension reduction | Improved accuracy with unbalanced data | [453] |
| O | FE/TD | CAD, DWT, and texture feature extraction | Combined CAD, wavelet transform, and texture features | [454] |
| CNN | FE/AC- | Image enhancement, segmentation, multi-feature extraction | Enhanced images, extracted multi-feature information | [455] |
| SVM | FE/LBP- | CNN deep features fused with HOG and SIFT | Combined handcrafted and deep features for improved accuracy | [436] |
| SVM | FE/GLCM | Median filter, contour delineation, GLCM texture features | Noise reduction, feature extraction, and multiple classifiers | [456] |
| SVM | FE/ICA | Multi-kernel-based SVM | Improved classification accuracy with ICA features | [457] |

**Table 26:** Percentage-based performance evaluations of several frameworks for thyroid cancer.

| AI Model | Dataset | ACC (%) | SPE (%) | SEN (%) | Ref. |
|---|---|---|---|---|---|
| CNN | PD | 88.00 | 79.10 | 98.10 | [458] |
| ELM | PD | 87.72 | 94.55 | 78.89 | [459] |
| MLP | PD | 87.16 | 87.05 | 91.18 | [452] |
| SVM | PD | 63.27 | 71.85 | 38.46 | [460] |
| RF | PD | 86.80 | 87.90 | 85.20 | [461] |
| LR | PD | 77.80 | 79.80 | 70.60 | [462] |
| B | PD | 84.69 | 86.96 | 82.69 | [456] |
| Ensemble DL | Cytological images | 99.71 | - | - | [463] |
| VGG-16 | Cytological images | 97.66 | - | - | [464] |
| VGG-16 | - | 99.00 | 86.00 | 94.00 | [465] |
| RF | Ultrasound | - | - | 94.00 | [466] |
| k-SVM | Ultrasound | - | - | 95.00 | [434] |
| ANN | Ultrasound | - | - | 69.00 | [467] |
| SVM RF | Ultrasound | - | - | 95.10 | [468] |
| ANN SVM | Ultrasound | 96.00 | - | - | [469] |
| RF | Ultrasound | - | - | 75.00 | [470] |
| CNN | DICOM | 83.00 | 85.00 | 82.40 | [471] |
| CNN | DICOM | - | 91.50 | - | [472] |
| Fine-tuned DCNN | PD | 99.10 | - | - | [473] |
| ResNet18-based network | PD | 93.80 | - | - | [474] |
| Multiple-scale CNN | PD | 82.20 | - | - | [475] |
| ThyNet | PD | - | - | 92.10 | [476] |
| AlexNet CNN | PD | 86.00 | - | - | [477] |
| DNN | ACR TIRADS | 87.20 | - | - | [478] |
| CNN (BETNET) | Ultrasound | 98.30 | - | - | [479] |
| ResNet | TIRADS | 75.00 | - | - | [480] |
| Xception | CT images | 89.00 | 92.00 | 86.00 | [481] |
| DCNN | Sonographic images | 89.00 | 86.00 | 93.00 | [438] |
| Google inception v3 | Histopathology | 95.00 | - | - | [482] |
| Cascade MaskR-CNN | Ultrasound | 94.00 | 95.00 | 93.00 | [483] |
| VGG16 | Ultrasound | - | 92.00 | 70.00 | [484] |
| VGG19 | Ultrasound | 77.60 | 81.40 | 72.50 | [485] |
| VGG16 | Ultrasound | 74.00 | 80.00 | 63.00 | [486] |

Additional research should address these shortcomings to enhance the efficacy of cancer detection. Enhanced diagnostic accuracy and novel techniques are necessary to differentiate between malignant nodules of different sizes. The study acknowledges the significance of state-of-the-art technologies, such as explainable AI, edge computing, remote sensing systems, and predictive modelling, in making diagnostic procedures more efficient. In order to facilitate a progressive transformation in cancer detection, there is a focused endeavour to develop advanced technologies that priorities privacy. These technologies aim to identify individuals with thyroid cancer and are also being utilized in broader applications such as telemedicine.

## 10. Role of AI in the Prostate Cancer

When it comes to cancer, prostate cancer (PC) is the most common type in men (apart from skin cancer) and the third leading cause of cancer-related death overall. Roughly twenty percent of American men will at some point in their lives encounter the same medical problem [487]. Another clinical issue resulting from the condition is the increased need for prostate biopsies and the shortage of urological pathologists, both of which provide challenges for treating the disease [488]. The process of clinical decision-making is further complicated by the inconsistency in assessing pathology indicators, leading to the potential for both overtreating and undertreating this condition [489]. By the year 2024, almost 50% of cancer cases in males will be prostate cancer (PC), with colorectal and lung cancers emerging as the three most common types [490].

### 10.1 Traditional Methods Used for the Prognosis of Prostate Cancer

#### 10.1.1 Digital Rectal Examination (DRE)

The digital rectal examination (DRE) is a widely employed screening method, however its diagnostic precision for prostate cancer remains uncertain due to inconsistent study findings on its sensitivity and specificity rates. Especially in primary care settings, there is not enough data to support its use as a test on its alone or in combination with other technologies [491]. In addition, Ying and his co-authors assess the precision of DRE in a group of patients suspected to have prostate cancer. Ultimately, it is recognized that DRE plays a crucial function in delivering information regarding prostate cancer [492].

### 10.1.2 The Prostate-Specific Antigen (PSA) Test

Benign prostate cells produce a protein known as the prostate specific antigen, or PSA. There has been debate on the efficacy of blood testing as a screening method for prostate cancer. PSA blood levels that are elevated can be a sign of prostate cancer, benign prostatic hyperplasia (BPH), or inflammation. This could start conversations regarding the validity of PSA testing and raise concerns about the potential for overdiagnosis. Merriel along with others. [493] a group of researchers who conducted a study to assess the efficacy of PSA in symptomatic patients. Ultimately, it is discovered that while PSA exhibits a significant level of sensitivity, what renders this marker ineffective is its lack of specificity, which therefore renders this test unable of differentiating between benign and malignant diseases.

### 10.1.3 Prostate Cancer Antigen 3

The gene responsible for prostate cancer antigen 3 (PCA3) produces a non-coding RNA molecule that is significantly more expressed in malignant prostate tissue compared to normal prostate tissue in humans. PCA3 is specifically and dramatically upregulated in patients of prostate cancer. PCA3 RNA is a good tumor marker due to its suppressed expression profile. A meta-analysis of thirteen trials assessed the diagnostic efficacy of urine PCA3 in prostate cancer.

The research revealed sensitivities and specificities of 75% and 62%, respectively [494]. Furthermore, the effectiveness of this method in diagnosing conditions has been confirmed in multiple countries and across various ethnicities [495]. PCA3 demonstrates effectiveness in predicting a positive biopsy result, particularly in cases when serum levels of prostate-specific antigen (Sr. PSA) range from 4 to 10 ng/ml. When PSAD is paired with it, the predictive accuracy is further increased [496].

### 10.1.4 OPKO 4K Test

The OPKO 4K test quantifies four kallikreins present in the serum: tPSA, fPSA, iPSA, and hk2 (total PSA, free PSA, intact PSA, and human kallikrein 2). These kallikreins are used in combination with age, previous biopsies, digital rectal examination (DRE), and prior biopsies to calculate a likelihood score for prostate cancer.

The presence of this substance was identified in the blood serum of individuals with ERSCP who underwent prostatectomies, and it seemed to serve as a predictive marker for the presence of a severe form of the disease.

The OPKO 4K test accurately forecasts the presence of aggressive prostatic cancer that has spread outside the prostate, has a tumor volume larger than 0.5 cm3, or has a Gleason grade of four or higher. The test has a high level of accuracy, with an area under the curve (AUC) value of 0.84 [497].

### 10.1.5 Multiparametric MRI

Prostatic cancer multiparametric MRI is used to both detect and stage the disease. The technology combines high-resolution T2-weighted images (T2WI) to assess anatomy, diffusion-weighted imaging (DWI), and MR spectroscopic imaging (MRSI) to accurately characterize lesions [498]. DCE-MRI is highly effective in detecting cancer. It utilizes magnetic resonance imaging (MRI) with the addition of contrast dye to visualize the movement of blood within tissues. Malignant tumors frequently exhibit distinct flow patterns compared to healthy tissues [499].

### 10.1.6 PIRADS

Prostate Imaging Reporting and Data Systems (PIRADS), which was established by the European Society of Urogenital Radiology (ESUR) in 2012, is a tool used to classify the likelihood of prostate cancer [500] employs multiparametric MRI images to develop a grading system for lesions. Each measure, including T2WI (differential descriptions of the peripheral zone (PZ) and transition zone (TZ)), DWI, DCE-MRI, and MRSI, is given a score of five on this scoring system. Subsequently, a medically significant abnormality can be anticipated by assigning a comprehensive score to each abnormality. A score is awarded to each variable [501].

### 10.1.7 Histo-Scanning

The PHS technology utilizes characterization algorithms to analyses distinct patterns found in both benign and malignant tissues. It mathematically processes the raw ultrasonic data using Histo-Scanning, an ultrasound-based technique. A volumetric equipment is used to get a precise measurement of the total volume of the prostate and any tumor lesions. Preliminary studies yielded promising results [502].

### 10.1.8 Elastography

The differentiation between healthy and malignant prostate tissue is mostly determined on tissue stiffness, making this approach effective. Currently, two techniques are used: strain elastography and shear-wave elastography. Research has shown that using an elastography-focused method is more effective than sextant biopsy in detecting high-risk prostate cancer. Furthermore, it increases overall sensitivity when performing a combination biopsy, requires fewer cores to diagnose prostate cancer, and lowers imaging time [503].

### 10.1.9 Biomarkers Screening for Prostatic Cancer

Biomarkers facilitate the advancement of novel diagnostic and therapeutic methodologies by offering valuable information about the molecular processes involved**.**

MicroRNAs (miRNAs) have a significant impact on controlling gene expression in cancer cells [504]. Detection of miRNAs is feasible in both blood and urine samples. Aberrant miRNAs produced by prostatic cancer can potentially be used as biomarkers to identify risk groups and determine the need for prostatic biopsy. Additionally, they can assist in the creation of treatments based on molecules [505]. Application of miRNAs 141 is promising [504].

### 10.2 Limitations of Tradition Diagnostic Method in prostate Cancer

Prostate cancer affects a significant proportion of men worldwide, and timely detection is crucial in combating this prevalent health problem. Medical practitioners have developed standardized techniques for identifying the disease however they are not completely optimal. Table 27 outlines the constraints of conventional approaches.

**Table 27:** Comparative Limitations of Prostate Cancer Diagnostic Methods

| Detection Method | Limitation | Ref. |
|---|---|---|
| Digital Rectal Examination (DRE) | Inaccuracy in prostate cancer detection tests can lead to either undiagnosed aggressive cancer or unnecessary surgeries for non-cancerous enlarged prostates. | [506] |
| The Prostate-Specific Antigen (PSA) Test | Elevated Prostate Specific Antigen (PSA) levels alone do not definitively diagnose prostate cancer. This can lead to unnecessary testing and treatments, especially in cases where prostatitis or benign prostatic hyperplasia (BPH) are present. | [507] |
| Prostate Cancer Antigen 3 (PCA3) | Variable sensitivity and high specificity may result in the omission of therapeutically relevant prostate cancer detection due to the heterogeneous tumor manifestations. | [508] |
| OPKO 4K Test | Despite its notable predictive accuracy, this method may not substantially diminish the necessity for prostate biopsies due to its inability to differentiate between dormant and aggressive malignancies with specificity. | [509] |
| Multiparametric MRI | Its application is restricted by accessibility, expense, and interpretation variability; as tumor size and Gleason score increase, so does sensitivity, which may result in the omission of small or exceptionally distinct lesions. | [510] |
| PIRADS | A greater incidence of false negatives for clinically significant prostate cancer, particularly in the transition zone, is attributable to interpretation variability and limited accuracy in specific instances. | [511] |
| HistoScanning | The diagnostic value and reproducibility of its results in detecting and defining prostatic lesions have been raised by inconsistent results throughout future investigations. | [512] |
| Elastography | Differentiating prostatitis from prostate cancer still requires further validation procedures, although such diagnostic approach can be applied just in limited conditions, once shown to have the potential to increase the accuracy of this procedure. | [513] |
| Biomarkers Screening for Prostatic Cancer | The new biomarkers present a bright functionality, triable they are standardized and proved their effectiveness. In spite of this, it remains challenging to identify the highly aggressive prostate tumors from their less damaging counterparts. | [514] |

### 10.3 Role of AI in Prostate Cancer

ML and AI are innovative technologies that can greatly improve the early identification of prostate cancer and reduce unneeded therapies. This has the potential to completely transform current medical procedures. Although the existing system for identifying PC has some shortcomings, AI shows great potential as a convincing alternative. By integrating imaging, physical findings, and genetic markers, AI will enhance the accuracy and thoroughness of prostate cancer detection. This is because AI is essential in the realm of big data, especially when considering its use in medical diagnostics.

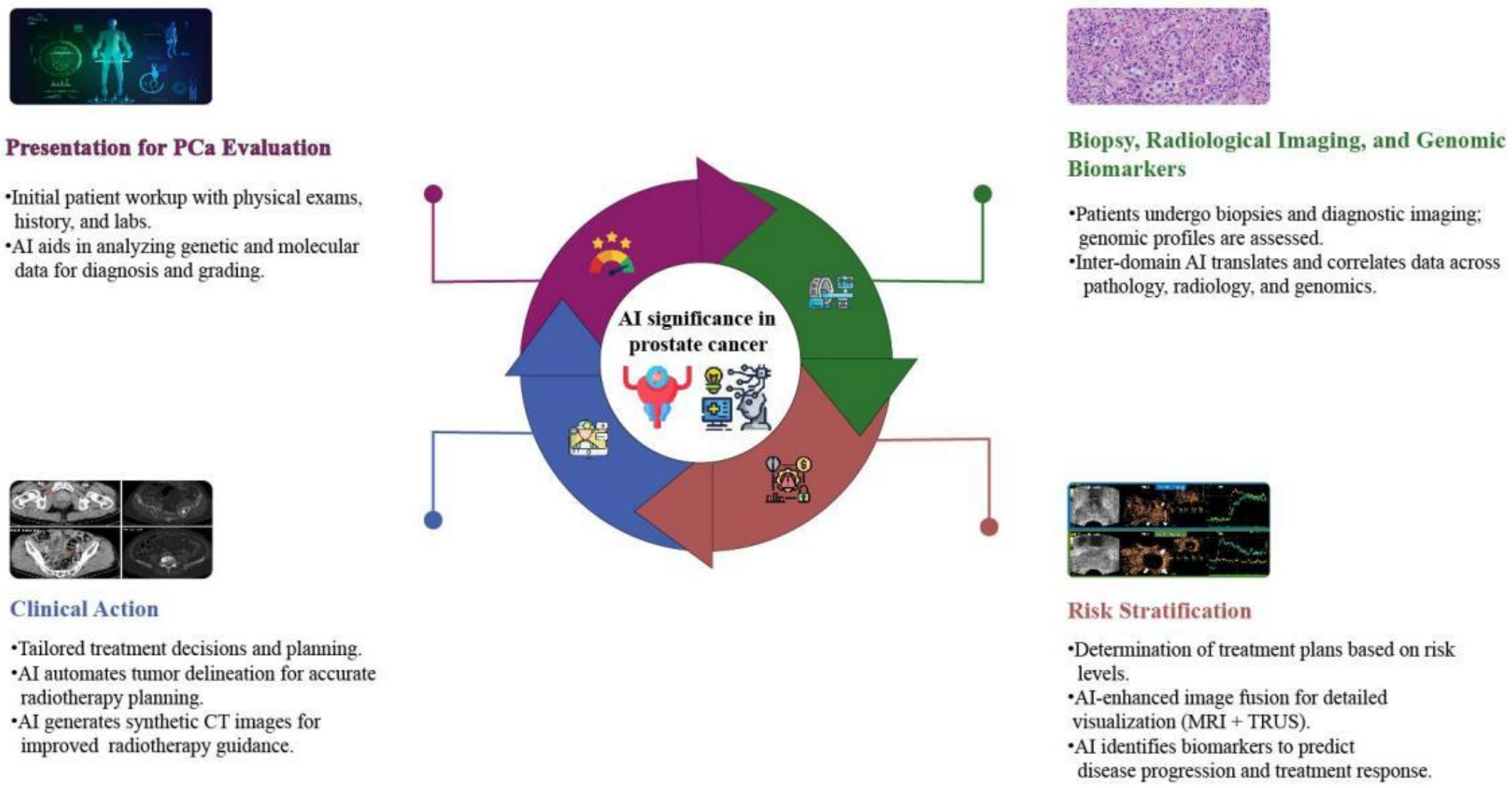

.

Figure 13 shows that AI has become a potent tool in the treatment of prostate cancer. AI carefully examines genetic and cellular data to improve our knowledge of the condition and interprets intricate details that a human analyst could ignore. AI actively contributes to the development of customized treatment plans, guaranteeing a one-of-a-kind approach for every patient, and enhances radiation accuracy by means of virtual imaging and ideal placement AI will influence several phases of diagnosis, such as risk and prognosis evaluation, by developing computer-based imaging and identifying markers that indicate the possibility of cancer development and medication response.

.

The field of prostate cancer research is undergoing a significant transformation with the use of AI in the domain of medical imaging. AI accurately assesses radiological pictures with a high degree of precision. This technology advancement enhances the diagnostic process by optimizing assistance and reducing the likelihood of errors through improved accuracy, reduced variability, and strong clinical decision-making [515].

A programmer capable of using AI to detect prostate cancer was developed by scanning core biopsy needles. An investigation analyzed the more extensive matter. An AI system that was put into action demonstrated a degree of accuracy in identifying cancer that was comparable to that of a highly skilled human pathologist [516].

The forthcoming effort will concentrate on developing and evaluating a DL system optimization to enhance the accuracy of classifying prostate MRI abnormalities. This study used PI-RADS for the purpose of resolving the issue of lesion categorization. The dataset utilised was sufficiently extensive to allow for several iterations of a ML algorithm, thereby identifying the constraints that would result in AI surpassing human doctors in real-world scenarios. This work is a positive step in exploring the potential of AI to enhance healthcare design, namely in its ability to assist clinicians in analyzing MRI data [517].

Several academics have explored the possibilities of AI technology in analyzing prostate cancer histopathology. Their goal is to improve clinical procedures by increasing accuracy, streamlining workflow, and reducing patient wait times. The deployment of Paige Prostate, an AI-driven automated system for prostate cancer detection, involves the utilizations of distinct datasets. The study showcases the potential of AI in enhancing patient care by its capacity to accurately identify situations requiring further examination and minimizing diagnostic inconsistencies.

AI exhibits a high level of sensitivity and specificity. The integration of AI in histopathology analysis is a promising advancement for the future of healthcare as it signifies a significant progress in introducing intelligent technology into the diagnostic process [518]. The major objective of the research was to detect prostate tumors using PET-CT scans. Thus, the potential of a completely automated AI approach was also taken into consideration. Translate that AI exhibits comparable, if not superior, proficiency to that of trained radiologists.

The efforts to incorporate AI capabilities in the analysis of PET-CT images demonstrate the inclination towards complexity and unpredictability of technology, which in turn has the potential to result in substantial advancements and novel methods to diagnosis [519].
A recent study examined the effects of two distinct artery input functions on distinguishing between intra-prostatic peripheral zone tumors and non-euplastic prostate tissue in DCE-MRI analysis.

The study showcases the use of high precision pharmacy-kinetic modelling to demonstrate how AI may be applied to analyze data from DCE-MRI [520]. The work highlights the need of accurate predictive modelling in identifying prostate cancer using MRI, which improves the performance and effectiveness of AI in diagnostic imaging.

AI intelligence has made significant advancements in the field of prostate biopsies, assisting doctors in more precise and efficient diagnosis, grading, and prognosis of prostate malignancies compared to traditional procedures. DL algorithms integrated into AI models are utilized to extract prostate cancer data from a substantial quantity of biopsy samples, so introducing a novel field in the realm of prostate cancer diagnosis.

This AI software provides pathologists with a comprehensive list of indicators that suggest the presence or absence of cancer in a patient, including the Gleason score (GS) and perineural invasion (PNI). These factors are intended to assist pathologists in making accurate decisions and, ultimately, to enhance patient outcomes [516].

A significant study focused on the utilisation of AI to develop a model that can differentiate prostate cancer from a visually identical condition using transrectal ultrasonography (TRUS) images of biopsy needle tract tissue. The algorithm is trained using a large dataset of thousands of TRUS pictures and has demonstrated superior accuracy compared to expert radiologists [521]. Furthermore, the attainment of this notable accomplishment demonstrates the capacity of AI to enhance the analysis of prostate cancer using imaging data. This, in turn, represents a new and innovative approach in utilizing AI for the diagnosis and evaluation of a prostate biopsy. These investigations clearly showcase the advanced level of AI in its application to prostate biopsy tests. Our accuracy, evaluation, scoring, and treatment planning capabilities are more than adequate. The advancements made in the field of AI in prostate cancer treatment can be recognized by studying its complexity.

Table 28 serves as a sparse collection of data points, but more importantly, it captures a comprehensive representation of competence. Within the artwork, AI models serve as the brushstrokes that enhance our comprehension and approach in combatting prostate cancer. These models, which incorporate various forms of diversification by utilizing inputs from different data sources such as histopathology slides, magnetic resonance imaging, and integrated clinical databases, play a crucial role in driving innovations in cancer care. They collectively contribute to the ongoing narrative of how cancer is currently being treated. These AI possess remarkable capabilities that are noteworthy.

The advent of cancer technologies has transformed the landscape of cancer care. These technologies have the power to unveil the evolution of the disease, enhance the grading of different stages, and enable more personalized treatment approaches. The following paragraph exemplifies AI's dedication to revolutionizing our response to a significant healthcare concern of our era, in conjunction with the strength and intensity of AI.

**Table 28:** Efficacy of AI Models in Prostate Cancer Diagnosis.

| AI-based Model | Data Type | Efficacy | Ref. |
|---|---|---|---|
| Model DEEP NET using deep learning for biopsies DEEP NET | Histopathological Data | High accuracy in detecting and scoring, prognostic for detrimental organizing and resurgence. | [522] |
| ResNet50 model based on TRUS images | MR Image | Enhanced diagnostic accuracy, increased sensitivity, and specificity | [523] |
| A motel YOLO3 propelled by AI for the early prediction of recurrence | Histopathological Data | Forecasts early recurrence following prostatectomy and pinpoints driving regions. | [524] |
| AI for prostate treatment planning automation | MR Image | Prostate and OARs outlines have a high level of clinical acceptance. | [525] |
| "Dr. Answer" AI for clinical outcome prediction | Clinical Data | Precise forecasting of clinical results after radical prostatectomy | [526] |
| A review of the use of AI to detect prostate cancer | MRI, Histopathological, Video | Describes the pros and cons of clinical workflow aid | [515] |
| AI to help doctors diagnose and treat patients | MR Image, Clinical Data | Potentially extensive applicability in terms of treatment and diagnosis efficacy | [527] |
| An AI algorithm for the diagnosis of prostate cancer in CNBs | Histopathological Data | Cancer detection, grading, and clinically significant findings with exceptional precision | [516] |
| Utilization of AI in imaging prostate cancer | MR Image | Possibility to direct treatment, reduce variability, and enhance diagnosis | [528] |

The unparalleled ability of AI to analyses various data sources, such as tiny tissue images, MRI scans, and patient files, has enabled it to outperform the current diagnostic tools, which were originally created to assist doctors and pathologists [529]. The research demonstrates the long-term potential of AI in enhancing the diagnosis of prostate cancer by implementing extremely precise techniques and advancing personalized therapy alternatives.

Nevertheless, there are still several unresolved aspects related to the discussed topic, such as the need for comprehensive validation efforts, the integration of widely accepted models into medical processes, and ensuring consistency in interpretations. Despite the hurdles, the advancement of diagnostic techniques utilizing AI is the most promising approach to enhance the effectiveness of prostate cancer therapy and other diseases in a more personalized and effective manner. This development holds the key to success for both patients and the medical community**.**

## 11. Skin Cancer and AI

Skin cancer is a widespread form of cancer that constitutes over one-third of all documented cancer cases globally. The occurrence of this phenomenon is steadily rising annually [530]. Over 9500 people are thought to be detected with skin cancer every day in the United States only [531]. Timely identification and proper diagnosis of skin cancer are crucial for achieving excellent treatment outcomes and enhancing patient survival rates, despite the disease being often curable [532-534]. Skin cancer is the predominant form of cancer in the United States, with around 5.4 million instances identified each year [535]. The treatment of skin cancer is possible, but its effectiveness relies on the timely identification of the disease. Postponing the identification and medical intervention of an infection can result in unfavorable outcomes, including higher rates of metastasis, morbidity, and death. Traditional approaches to detecting skin cancer, such as visually examining the skin and performing a biopsy, are subjective, lengthy, and prone to considerable variations in interpretation [536]. There has been a growing interest in using ML technologies, particularly DL approaches, to improve the effectiveness and accuracy of skin cancer diagnosis [537].

### 11.1 Evaluating Tradition Detection Method and AI Innovation in Skin Cancer

There is an increasing interest in utilizing ML technologies, DL methods, to enhance the efficiency and precision of skin cancer diagnostics. Skin cancer is characterized by the excessive growth of abnormal cells in the skin. Melanoma, squamous cell carcinoma, basal cell carcinoma, and Merkel cell carcinoma are the most common types of skin cancer [531]. The primary cause of DNA damage resulting in non-melanoma skin cancer is exposure to ultraviolet light [538]. Timely identification and proper diagnosis of skin cancer are crucial for achieving excellent treatment outcomes and enhancing patient survival rates, despite the disease being often curable [532-534]. In the past, skin cancer detection has relied on a combination of histological investigation and visual inspection. However, both methods have had notable limits in terms of accuracy and scalability. Nevertheless, the traditional approaches for detecting skin cancer, such as visually examining and studying tissue samples, are time-consuming, subjective, and exhibit significant variability among different observers. Over the past few years, numerous non-invasive imaging methods [539] have been developed to aid in the detection and monitoring of skin cancer.
One well-known example is the use of multi-spectral sensors to detect changes in the refraction index in millimeter-wave to terahertz photonic near-field imaging [540]. The application of novel AI-based methodologies for skin cancer detection is presented in Table 29. Advancements in several aspects of AI have had significant and wide-ranging impacts across multiple fields. AI has the capacity to. Significantly improve cancer diagnoses.

Recent advancements in DL algorithms have greatly contributed to the field of medical diagnostics. Cutting-edge models like Xception, ResNet50V2, and InceptionResNetV2 are being utilised to enhance the accuracy and efficiency of medical diagnoses [541, 542]. The achievement of consistent and accurate diagnoses in skin cancer detection is challenging due to the limitations of current diagnostic technologies and the extensive range of abnormalities. Furthermore, transfer learning has resulted in enhancements in the precision of classification [543, 544]. Several techniques involve utilizing pre-trained CNN models to extract characteristics from medical images. The application of DL techniques for the identification and classification of skin cancer has emerged as a rapidly growing area of research in recent years.

Multiple studies have demonstrated that CNN model has the ability to accurately identify and classify skin cancer from medical images. Additionally, the utilization of transfer learning can further improve the accuracy of classification [545, 546].

By employing AI algorithms for image analysis and skin cancer diagnosis, it is possible to mitigate healthcare inequities by eliminating inadvertent doctor bias and improving accessibility and cost-effectiveness [532]. An AI technique is employed for the identification of skin cancer, as depicted in Figure 13. AI systems have demonstrated similar levels of performance [547, 548] as well as, in some cases, superior performance compared to dermatologists in accurately categorizing skin lesions [549]. The potential for collaboration between humans and computers to greatly boost diagnostic accuracy is substantial [550].

**Table 29:** Recently researched various skin cancer datasets and AI Models

| Model | Datasets | Performance | Reference |
|---|---|---|---|
| AlexNet, VGG16, VGG19 | ISIC 2016, 2017, PH2, HAM10000 | Accuracy = 98.33%, F1 score = 96% | [551] |
| Deep Residual Network | ISIC 2017, HAM10000 | Accuracy = 96.97% | [552] |
| CNN | Xiangya-Derm | AUC = 0.87 | [553] |
| CNN, ResNet50 | HAM10000 | F1-score = 0.859 (CNN), 0.852 (ResNet50) | [554] |
| ResNet-101 and SVM | ISIC 2019, 2020 | Accuracy = 96.15% (ISIC19), 97.15% (ISIC20) | [555] |
| NASNet | ISIC 2020 | Accuracy = 97.7%, F1-score = 0.97% | [556] |
| ResNet15V2, MobileNetV2 | ISIC 2019 | Accuracy = 89%, F1-score = 0.91 | [541] |
| Xception, InceptionV3, VGG19, ResNet50, MobileNet | HAM10000 | Accuracy = 90.48% (Xception) | [557] |
| Xception, DenseNet201, ResNet50V2, MobileNetV2, VGG16, VGG19, Google Net | ISIC 2019 | Accuracy = 76.09% | [542] |
| CNN, VGG16, Xception, ResNet50 | HAM10000 | Accuracy = 88% (VGG16) | [558] |
| ResNet101, InceptionV3 | ISIC archive | F1-score = 84.09% (ResNet101), 87.42% (InceptionV3) | [559] |
| MobileNetV2 | ISIC 2020 | Accuracy = 98.20% | [560] |
| InSiNet, U-Net | HAM10000, ISIC 2019, 2020 | Accuracy= 94.59% (HAM10000), 91.89% (ISIC2019), 90.54% (ISIC2020) | [561] |
| ResNet101, DenseNet201 | ISBI 16, 17, 18, PH2, HAM10000 | Accuracy = 98.70% (PH2), 98.70% (HAM10000) | [562] |
| A hybrid model | ISBI 2018 | Accuracy = 92.70% | [563] |
| Top 10 model Average | ISIC 2018 | Accuracy = 86.7% | [564] |
| CNN | ISIC 2018 | Accuracy = 83.2% | [565] |

Most AI developments have primarily used homogeneous datasets[566, 567] collected from countries with predominantly European ancestry [568]. The omission of the skin of color from training datasets increases the likelihood of inaccurate diagnoses or the complete omission of skin malignancies [547]. Moreover, it increases the already existing racial disparities in the field of dermatology [569].

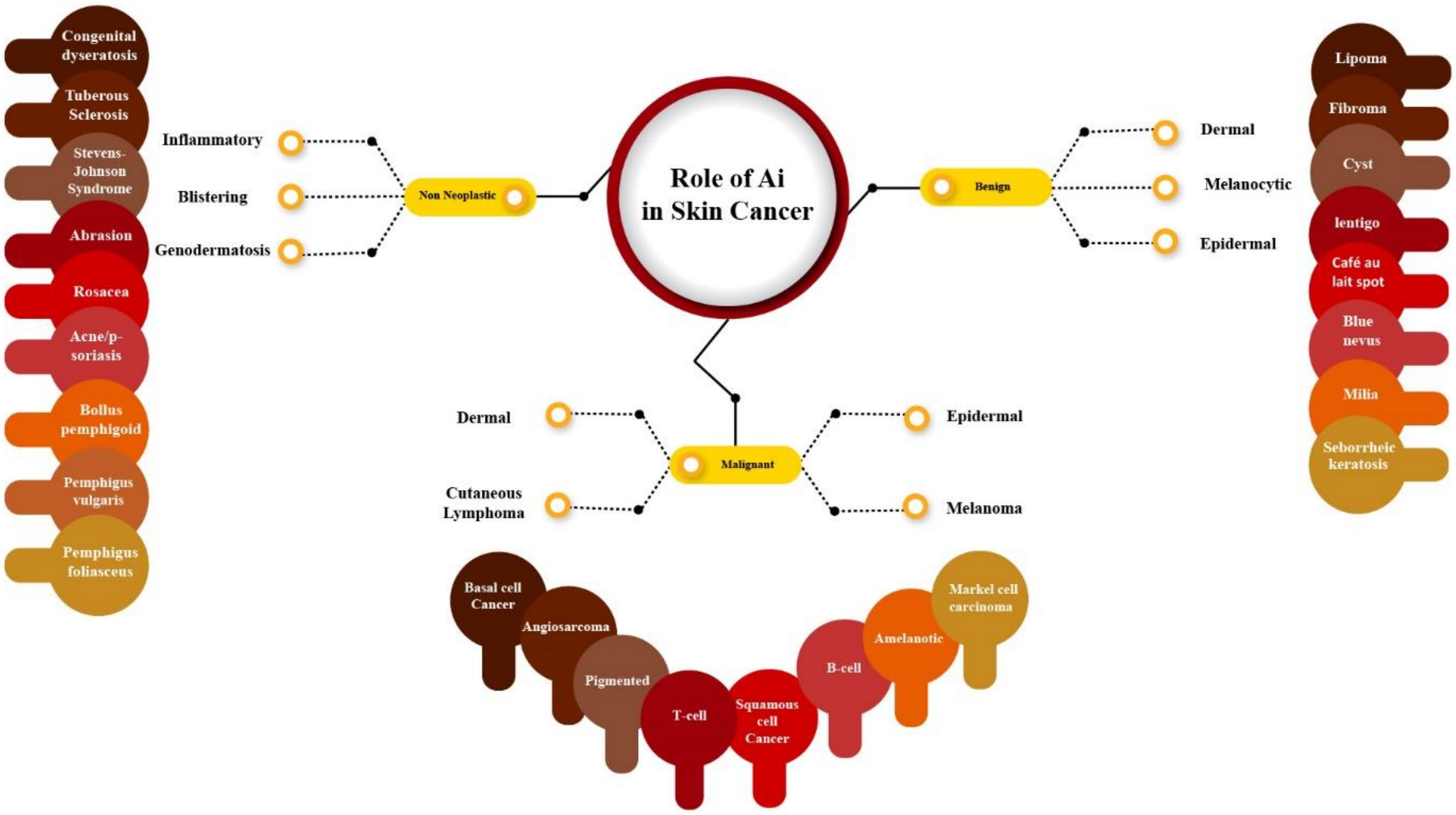

**Figure 14:** This figure, emphasizing on the use of AI in the detection and treatment of skin cancer, divides different skin disorders into three categories: benign, malignant, and non-neoplastic. Inflammatory and blistering diseases are examples of non-neoplastic disorders; benign conditions are further classified into dermal, melanocytic, and epidermal

types, each of which has particular examples such as nevi and keratoses. On the other hand, malignant conditions focus on several forms of skin cancer, including cutaneous lymphoma, melanoma, squamous cell carcinoma, and basal cell carcinoma. The main focus is on the possible application of AI in the diagnosis and differentiation of various skin conditions, which can range from benign to malignant. This is compatible with the growing contribution of AI to dermatology's prognosis evaluations, treatment planning, and diagnosis precision.

While numerous studies have assessed the effectiveness of AI-driven models in identifying skin cancer [570, 571], there is currently no research investigating the utilization of AI in populations with darker skin tones. AI algorithms are capable of categorizing photos of pigmented skin lesions in populations with darker skin tones. Integrating AI algorithms into the diagnosis of pigmented lesions and screening for skin cancer has the potential to greatly improve patient outcomes and streamline operations in the area of dermatology. The efficacy and reliability of an AI algorithm that analyses photos are significantly impacted by the caliber of the data utilized for training and evaluation. If AI models utilized in clinical practice do not incorporate photos from diverse racial groups, they may fail to detect or experience delays in identifying tumors in individuals of color. This would further complicate health consequences [572, 573].

## 12. Discussion

AI applications in oncology are an important milestone towards customized cancer care. The development will involve a variety of aspects, including improved accuracy in detecting early-stage cancer and the administration of personalized treatments to specific patients. AI is used to treat various types of cancer, including lung and skin cancer, among others. This research revealed the detailed and intricate role that AI can play in cancer diagnosis, treatment planning, and outcome prediction.

AI has the potential to improve patient outcomes and reduce the overall cost of cancer care by analyzing vast datasets, identifying complex patterns, and even predicting clinical outcomes more accurately than humans. DL algorithms supported by AI have a higher reputation for identifying lung and breast cancer in the early stages with greater precision. It has been shown that using AI to look at imaging and genomic data can help find disease biomarkers and create personalized treatment plans for colorectal and liver cancers. AI technology improves endoscopic imaging and histological evaluations, making it critical in the diagnosis of stomach and esophageal malignancies. This allows for early detection and precise staging of these tumors. Including cervical, thyroid, and prostate cancers in our study shows that AI improves the accuracy of screenings and biopsies, which leads to finding cancers earlier and figuring out how bad they are. AI algorithms have demonstrated outstanding performance in classifying various types of skin lesions in patients with skin cancer, which could reduce the burden on dermatologists and enhance diagnostic precision. Nevertheless, the integration of AI into everyday healthcare environments brings some challenges. The data requirements include addressing any discrepancy that may arise from non-representative data training, guaranteeing the transparency and understanding of the algorithms, and making significant use of labeled training datasets to strengthen the models.

Furthermore, more academic research and regulatory development are required to address ethical concerns regarding patient confidentiality in the context of using AI as a tool in the medical sector.

## 13. Conclusion

The integration of AI into the field of oncology represents a groundbreaking change in the approach to combating cancer. The novel approach to treating cancer, which involves early identification, diagnosis, personalized therapy, and prognosis prediction, has emerged as a result of this significant transformation. This study on the utilizations of AI in tackling 10 cancers has showcased its capacity to expedite the clinical decision-making process, enhance the precision of diagnoses, and adapt treatment strategies to cater to the unique needs of individual patients. The enhanced powers of AI allow it to manipulate complicated data and recognize patterns beyond the capacity of the human brain. This has the potential to drive advancements in identifying new biomarkers, improving risk analysis, and enabling early detection of cancer. However, there are constraints to fully utilizing the potential of AI in cancer. These limitations consist of the necessity to bridge the technological gap, offer an extensive variety of data, be transparent in algorithms, and maintain ethical integrity to guarantee that AI breakthroughs are available to a large audience. The effective implementation of AI in cancer treatment needs a collaborative effort that brings together individuals from the fields of computation and healthcare, including patients and legislators. AI is not the only possible remedy, but it is a powerful tool that offers the potential to enable clinicians to deliver superior, more precise, and individualized cancer treatment. In order to ensure that AI improves rather than replaces the essential human qualities of empathy and competence in helping others, cooperation is necessary to tackle the ethical, technological, and therapeutic issues associated with the combination of AI in the future. Our common goal as we move through this transitional time is still to effectively employ AI to revolutionize cancer treatment, improve patient outcomes globally, and lead the way in precision-based oncology.

### 13.1 Future Directions

The incorporation of AI in the treatment of cancer is now in a promising stage, with significant advancements on the horizon that might potentially revolutionize the area of oncology. The incorporation of AI with genetics, imaging, and clinical data is anticipated to become more widespread, providing a comprehensive viewpoint on patient therapy. The progress of sophisticated algorithms capable of combining multiple types of data will enable the creation of more accurate and tailored treatments. Furthermore, advancements in the comprehensibility and clarity of AI will enhance trust and approval among professionals. The availability of AI tools through open-source platforms has the potential to accelerate research and improve accessibility, enabling resource-limited situations to reap the advantages of emerging technologies. In addition, there will probably be a shift in focus towards preventive oncology, utilizing AI to identify risk factors and intervene before the onset of cancer.